\PassOptionsToPackage{table,dvipsnames}{xcolor}
\documentclass[]{fairmeta}

\title{BigO(Bench) - Can LLMs Generate Code with Controlled Time and Space Complexity?}

\author[1,2]{Pierre Chambon}
\author[1,*]{Baptiste Roziere}
\author[2]{Benoit Sagot}
\author[1]{Gabriel Synnaeve}

\affiliation[1]{FAIR at Meta}
\affiliation[2]{Inria}
\affiliation[*]{Work done at Meta, now working at Mistral AI}

\correspondence{Pierre Chambon at \email{pchambon@meta.com}}
\metadata[Code]{\url{https://github.com/facebookresearch/bigobench}}
\metadata[Data]{\url{https://huggingface.co/datasets/facebook/BigOBench}}
\metadata[Homepage]{\url{https://facebookresearch.github.io/BigOBench}}
\metadata[Leaderboard]{\url{https://facebookresearch.github.io/BigOBench/leaderboard.html}}

\usepackage{amsmath,amsfonts,bm}

\def\eqref#1{equation~\ref{#1}}

\def\1{\bm{1}}

\DeclareMathAlphabet{\mathsfit}{\encodingdefault}{\sfdefault}{m}{sl}
\SetMathAlphabet{\mathsfit}{bold}{\encodingdefault}{\sfdefault}{bx}{n}

\usepackage{url}
\usepackage{rotating}
\usepackage{xspace}
\usepackage{forest}
\usepackage{algorithm}
\usepackage{algorithmic}
\usepackage{listings}
\usepackage{wrapfig}
\usepackage{adjustbox}
\usepackage{array}
\usepackage{newunicodechar}
\crefname{algorithm}{Algorithm}{Algorithms}
\Crefname{algorithm}{Algorithm}{Algorithms}

\newunicodechar{’}{'}
\newunicodechar{‘}{`}
\newunicodechar{“}{``}
\newunicodechar{”}{''}
\newunicodechar{…}{\ldots{}}
\newunicodechar{–}{--}
\newunicodechar{—}{---}
\newunicodechar{²}{\ensuremath{^2}}
\newunicodechar{ă}{\u{a}}
\newunicodechar{Ł}{\L{}}
\newunicodechar{ń}{\'n}

\definecolor{keywords}{RGB}{255,0,90}
\definecolor{comments}{RGB}{0,0,113}
\definecolor{red}{RGB}{160,0,0}
\definecolor{green}{RGB}{0,128,128}
\definecolor{codegreen}{rgb}{0,0.6,0}
\definecolor{codegray}{rgb}{0.5,0.5,0.5}
\definecolor{codepurple}{rgb}{0.58,0,0.82}

\lstdefinestyle{mystyle}{
    backgroundcolor=\color{white},
    commentstyle=\color{codegreen},
    keywordstyle=\color{magenta},
    numberstyle=\tiny\color{codegray},
    stringstyle=\color{codepurple},
    basicstyle=\footnotesize,
    breakatwhitespace=false,
    breaklines=true,
    captionpos=b,
    keepspaces=true,
    numbersep=5pt,
    showspaces=false,
    showstringspaces=false,
    showtabs=false,
    tabsize=2
}
\newcommand{\benchmark}{\mbox{\textsc{BigO(Bench)}}\xspace}
\newcommand{\vendor}{\mbox{\textsc{Code Contests}}\xspace}
\newcommand{\codeforces}{\mbox{\textsc{Codeforces}}\xspace}
\newcolumntype{R}[2]{%
    >{\adjustbox{angle=#1,lap=\width-(#2)}\bgroup}%
    l%
    <{\egroup}%
}

\newif\ifshowtext
\showtexttrue

\abstract{%
We introduce \benchmark, a novel coding benchmark designed to evaluate the capabilities of generative language models in understanding and generating code with specified time and space complexities.
This benchmark addresses the gap in current evaluations that often overlook the ability of models to comprehend and produce code constrained by computational complexity.
\benchmark includes tooling to infer the algorithmic complexity of any Python function from profiling measurements, including human- or LLM-generated solutions.
\benchmark also includes of set of 3,105 coding problems and 1,190,250 solutions from \vendor annotated with inferred (synthetic) time and space complexity labels from the complexity framework, as well as corresponding runtime and memory footprint values for a large set of input sizes.
We present results from evaluating multiple state-of-the-art language models on this benchmark, highlighting their strengths and weaknesses in handling complexity requirements.
In particular, token-space reasoning models are unrivaled in code generation but not in complexity understanding, hinting that they may not generalize well to tasks for which no reward was given at training time.
}

\begin{document}

\maketitle

\begin{center}
\includegraphics[width=0.95\textwidth]{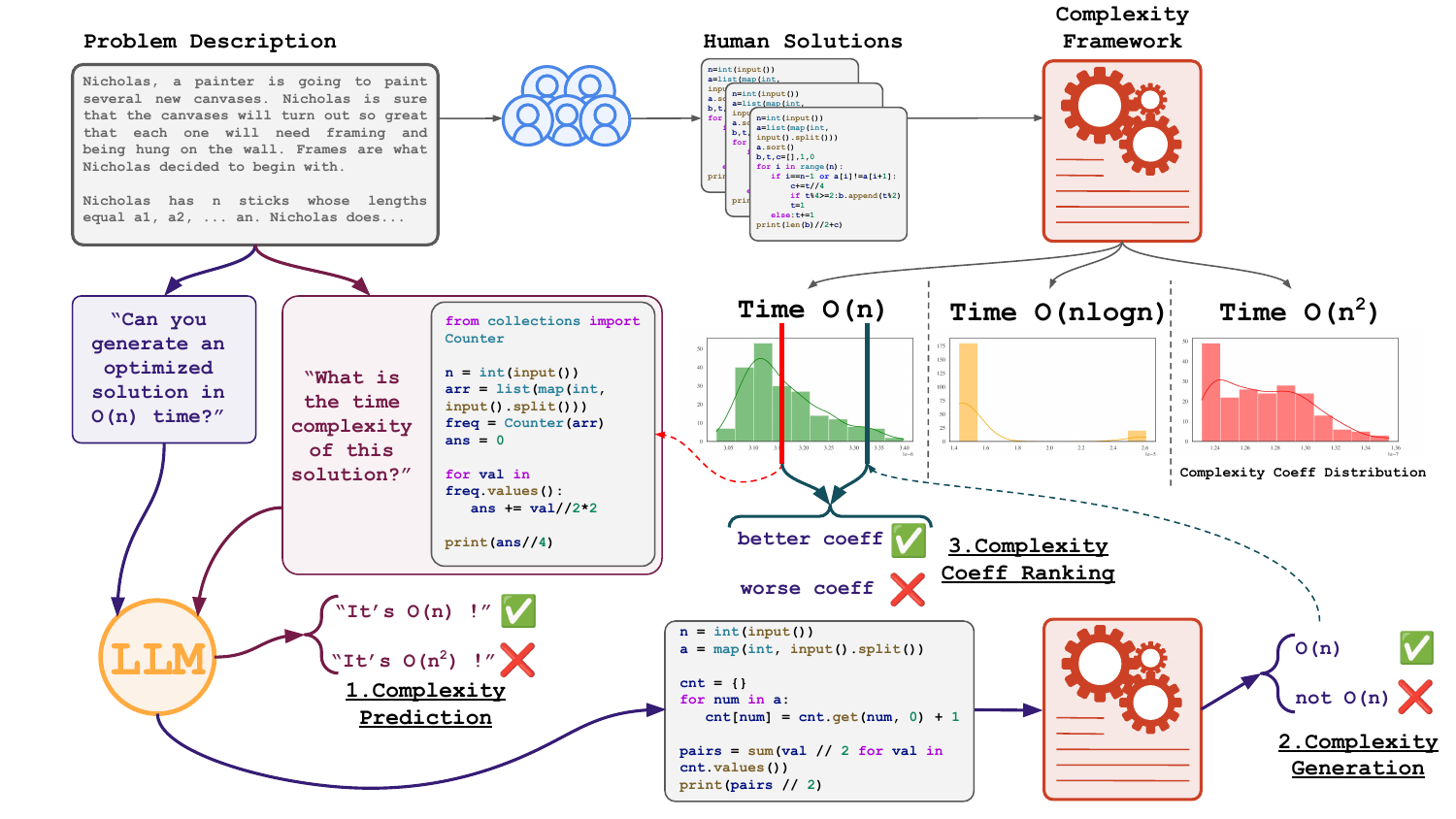}
\vspace{-0.7em}
\captionof{figure}{
\benchmark framework overview: Given a coding problem and human solutions, the framework evaluates LLMs on three key tasks: (1) predicting time-space complexities of existing solutions, (2) generating new code that meets specified complexity requirements, and (3) ranking solutions against human-written code with similar complexity profiles. The framework validates model outputs by computing runtime distributions and curve coefficients.
\vspace{-1em}
}
\label{benchmark_overview}
\end{center}

\section{Introduction}
\label{introduction}
\vspace{-0.5em}

A junior developer writes an elegant solution to a coding challenge that passes all test cases, yet fails catastrophically in production. The issue isn't a bug -- it's an $O(n^2)$ algorithm processing millions of records, when an $O(n\cdot\mathrm{log}(n))$ solution could have handled the load effortlessly. As large language models (LLMs) increasingly assist in code generation, their ability to understand and control computational complexity becomes critical. While modern LLMs can generate syntactically correct and functional code with impressive accuracy, our new benchmark \benchmark shows that they often struggle with the higher-level reasoning required to optimize time and space complexity -- a skill that separates novice programmers from experienced engineers.

Our comprehensive evaluation of state-of-the-art code generation models reveals a concerning gap: while reasoning models like DeepSeek R1 \citep{deepseekai2025deepseekr1incentivizingreasoningcapability} achieve above 70\% accuracy (pass@1) on programming contests (\textsc{CodeForces}\footnote{\href{https://codeforces.com/}{https://codeforces.com/}}), they show significantly weaker performance (4.8\%) when tasked with generating solutions under specific complexity constraints (a detailed failure example is in Appendix \ref{appendix:complexity_generation_example}). They fare barely better than non-reasoning models at analyzing a function for its complexity, e.g. 6.8\% better in complexity prediction vs. Llama 4 Maverick \citep{meta2025llama}. This limitation becomes particularly acute in real-world applications, where scalability and controllable, understandable, performance constraints are often as critical as functional correctness.

Our main contributions are threefold:
\vspace{-0.7em}
\begin{itemize}
    \item Firstly, to address the challenge mentioned above, we introduce \benchmark, a novel benchmark for code generation that evaluates a model's understanding of time and space complexities, including runtime and memory profiling measurements for a set of 3,105 coding problems and 1,190,250 solutions from \vendor \citep{Li_2022}. As shown in \cref{benchmark_overview}, for a given coding challenge and human solution, the model can be queried to a. predict time-space complexities, b. generate code that solves the challenge while adhering to a specified feasible complexity, and c. on top of it ranks better than human solutions of the same challenge and complexity.
    \item Secondly, we develop a complexity inference framework that takes a Python function and returns time and space complexities. It is a rule-based algorithm based on fuzzing, profiling, and regression over major complexity classes (including multi-dimensional ones). This is what we used to produce ground truth labels for \benchmark, which are statistically significant ground truth performance profiles and not theoretical complexities. This complexity evaluation framework achieves 84\% and 82\% match (with human annotated theoretical complexity) respectively on time/space complexity test sets.
    \item Thirdly, we evaluate 14 popular models on our benchmark along fine-tuned ones and compare performance in detail: using our $All@1$ metric, \textsc{DeepSeek-R1 Llama70B} \citep{deepseekai2025deepseekr1incentivizingreasoningcapability} scores best on time complexity prediction (41.4\%), while \textsc{Qwen3 32B}\citep{qwen3technicalreport} leads on time complexity generation (6.5\%) and \textsc{Llama 3.1 Nemotron-Ultra}\citep{bercovich2025llamanemotronefficientreasoningmodels} on space complexity generation (5.6\%).
\end{itemize}

\section{Related Work}
\label{related_work}

\begingroup
\titleformat*{\paragraph}{\normalfont\sffamily\bfseries}

\paragraph{Benchmarks for Code Generation}
\label{benchmarks_code_generation}

As the coding skills of LLMs were still limited, benchmarks for code generation originally focused on simple functions and coding challenges, as illustrated by HumanEval \citep{chen2021evaluating} and MBPP \citep{DBLP:journals/corr/abs-2108-07732} , probably the most famous coding benchmarks.
Today, these benchmarks are considered saturated, as top reported $pass@1$ scores lean towards 90-95\% success rate. A first area of work has focused on extending, improving quality and correcting these benchmarks, be it with HumanEval+, which adds more tests to pass \citep{liu2023codegeneratedchatgptreally}, or HumanEvalPack, which adds more tasks and programming languages \citep{muennighoff2024octopackinstructiontuningcode}.

A different area of research pursues scale and reasoning around code as a way to formulate benchmarks that can challenge LLMs over a longer term. SWE-Bench \citep{jimenez2024swebenchlanguagemodelsresolve} and AssistantBench \citep{yoran2024assistantbenchwebagentssolve} leverage Github as a source of large software development tasks, that do not resemble nor include obvious patterns of reproduction, therefore not solvable with simple fine-tuning on similar data. Nevertheless, the evaluation cost in time and compute is a non-negligible limitation that restrains certain teams from using these benchmarks.

\benchmark is a tentative benchmark to integrate challenging notions of reasoning around code into a simple formulation, providing a practical evaluation metric for code LLMs that can easily be used to discriminate and iterate improvements of their coding capabilities.

\paragraph{Time-Space Complexity Task}
\label{time_space_complexity}

Coding interviews for software engineers are centered around small coding challenges to be solved and explained, typically by discussing the time-space complexity of the proposed solutions.
Only a few previous works attempted to frame the task of time-space complexity explanation for LLMs. Nevertheless, they all fall short of providing sufficient elements to build a solid benchmark: CoRCoD \citep{sikka2019learningbasedmethodscode} contains 932 Java code pieces labeled for five time complexity classes without using any LLMs;
TASTY \citep{moudgalya2023tastytransformerbasedapproach} uses 3000 C++/Python problems across five complexity classes, limited to classification only and just benchmarking small BERT models \citep{devlin2019bertpretrainingdeepbidirectional};
CodeComplex \citep{baik2024codecomplextimecomplexitydatasetbilingual} contains 10k Python/Java programs annotated for five time complexity classes as classification only; finally RACE \citep{zheng2024correctnessbenchmarkingmultidimensionalcode} contains limited test cases ($\sim$100) and only measure runtime similarity between proposed and human solutions.

Most of the work around code efficiency focuses on absolute runtime, memory, or resource measurement, scoring code as faster or lighter under a fixed execution setting \citep{qiu2025efficientllmgeneratedcoderigorous,peng2025coffecodeefficiencybenchmark,huang2024effilearnerenhancingefficiencygenerated,liu2024evaluatinglanguagemodelsefficient,peng2024perfcodegenimprovingperformancellm,du2024mercurycodeefficiencybenchmark}. More recently, this line has expanded to multi-language efficiency evaluation \citep{qing2025effibenchxmultilanguagebenchmarkmeasuring}, synthesized performance constraints \citep{yang2025synthesizingperformanceconstraintsevaluating}, repository-level optimization for software-engineering agents \citep{shetty2025gsochallengingsoftwareoptimization,he2025sweperflanguagemodelsoptimize,ma2025swefficiencylanguagemodelsoptimize,ren2025peaceefficientprojectlevelefficiency}, system-level high-performance code \citep{jing2026perfcodebenchbenchmarkingllmssystemlevel}, and optimization methods based on execution feedback, relative preferences, reusable skills, or iterative rewriting \citep{du2025afterburnerreinforcementlearningfacilitates,hajizadeh2026effipairimprovingefficiencyllmgenerated,wang2026effiskillagentskillbased,yang2026perfcoderlargelanguagemodels,wu2025fasterpyllmbasedcodeexecution}. Whereas absolute runtime remains context-dependent and conditioned on hardware and specific test cases being executed, code complexity reveals the intrinsic comprehension the model has of the underlying algorithmic structure of the code, measuring its efficiency at scale and its asymptotic performance.

Several recent papers have since studied closely related questions around complexity-aware and execution-grounded code evaluation. DynaCode, released in the same period, studies dynamic code-generation tasks with controlled structural and call-graph complexity, providing a complementary view of task difficulty rather than explicit time-space complexity control \citep{hu2025dynacodedynamiccomplexityawarecode}. Other papers examine runtime behavior among correct generations, showing that solutions passing the same tests can still diverge in memory traces, opcode-level execution patterns, or stability under sampling \citep{rajput2026correctnessisntefficiencyruntime,rajput2025dynamicstabilityllmgeneratedcode}. SwiftSolve is a direct follow-up to the benchmark setting: it uses the work introduced in the present paper as an evaluation bed to study whether a multi-agent self-repair loop can improve generated solutions under complexity constraints \citep{singh2025swiftsolveselfiterativecomplexityawaremultiagent}. These papers reinforce the same conclusion that pass/fail correctness is not sufficient, while \benchmark specifically evaluates whether models can predict, generate, and rank solutions under explicit asymptotic time and space complexity constraints.

In order to improve on the previous attempts, \benchmark explores not only time but also space complexity, out of an unconstrained set of classes to capture more various solutions and problems.
The benchmark not only studies the classification task but also the more challenging open-framed generation task, so to mimic the real-world thought process of designing a solution for a target complexity. Altogether, this turns out to create a challenging task that wide-used LLMs are benchmarked upon, and hopefully it provides a new perspective on the limitations of current models and their reasoning capabilities around code.

\endgroup

\section{Dynamic Complexity Inference Framework}
\label{complexity_prediction_framework}

Throughout this study, complexity refers to worst-case complexity, finding how input growth maximally affects runtime and memory. Python is considered as the only language studied, and therefore complexity can account for python-specific optimizations (e.g. CPython \citep{cpython} or the compiler) that get reflected in the empirical time and space measures. In the quest of finding the worst case scenario of a snippet of code, natural language constraints on the inputs as detailed in the problem description can be ignored, as long as the program runs and does not fail; \cref{app:contest_constraints} explains why we scope the measurement this way rather than to the input bounds stated by the problem. The framework therefore returns an empirical estimate of these complexities rather than a proof about the source, and \cref{time_space_complexity_definitions} states the conventions it follows.

\begin{figure}[t!]
\centering
\includegraphics[width=\textwidth]{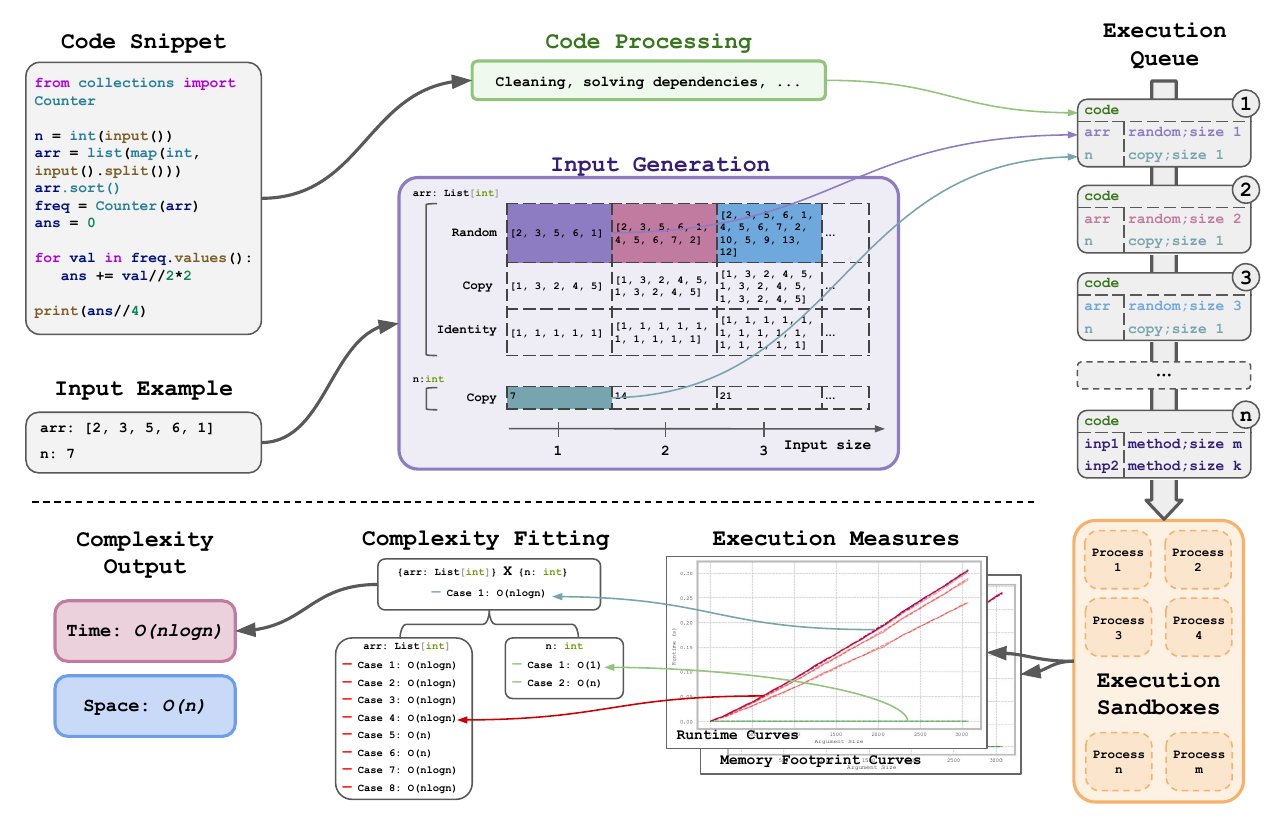}
\caption{
Outline of the dynamic complexity inference framework. The framework takes a code snippet and a single example of inputs to this code snippet. Then, it processes the code snippet and proceeds with extensive inputs generation, based on the provided example of inputs: inputs are independently or interdependently increased in size, using several expansion methods that can be the identity or random, among else. This forms a queue of synthetic inputs on which to execute the provided code snippet. These executions happen independently in sandboxes, where runtime and memory footprint measures are taken. Once all the measures are collected, the framework can model the code snippet time and space dependencies to the different inputs. Using curve fitting, the time and space complexity of the code is computed on each input separately and then altogether. The global time and space complexity over all inputs is what is being returned.
\vspace{-1.2em}
}
\label{framework_complexity}
\end{figure}

\paragraph{Implementation}
\label{implementation}

The time-space complexity framework is a rule-based algorithm that processes Python functions to infer time and space complexities dynamically. As depicted in \cref{framework_complexity}, it takes a Python function with example inputs and corresponding dataclass (\cref{dataclass_generation}), processes them, then measures runtime and memory during several executions.
From a high-level perspective, the framework increases input sizes using various strategies to assess size impact on execution metrics (runtime, memory).
For multi-argument functions, arguments can be expanded independently or together to determine overall complexity, considering interdependencies.
Prepared code and expanded inputs run in independent \textit{Bubblewrap} sandboxes \citep{bubblewrap} to prevent harmful side effects. While running, \textit{Cprofiler} is used for runtime measures and \textit{tracemalloc} for memory footprint.
Using non-negative least squares curve fitting \mbox{\citep{Lawson1976SolvingLS}} on each set of measures, the coefficients and residuals of each complexity class are computed. The gold complexity class output for a given set of measures is chosen as the minimizer of the residuals, taking into account a simplicity bias (the more simple the complexity class is, the smaller the simplicity bias).
This curve fitting is applied on each set of measures, each corresponding to a different subset of arguments being expanded with a different expansion method.
Using ensemble methods, the global complexity of the Python function is computed by aggregating the individual complexity outputs along the different set of measures.
Finally, the complexity framework also returns the coefficients of the curve of each elected complexity. These coefficients can be leveraged to rank and classify different optimized Python solutions within the same complexity class. More details are provided in \cref{implementation_details}.

\paragraph{Parametrization}
\label{parametrization}

The framework involves three main parametrized steps. The first step, \textit{Process Allocation}, handles the multiple $(Code, Expanded\ inputs)$ pairs to be run and measured for time and space, trying to maximize the execution throughput while minimizing its variability and instability. Second, \textit{Execution Measures} consists in leveraging various measuring tools for various ranges of input size values with more or less granularity.
Third, \textit{Complexity Fitting} relies on the measures obtained to apply various curve fitting methods, aggregation methods and ensemble methods to form the global complexity formula across all inputs.

Parameters from each of these three groups were optimized towards three metrics of interest: pure accuracy of detecting the correct complexity class, coverage by handling as many code snippets and problems as possible, and self-consistency of outputting stable results over multiple runs and across different compute instances. How much the released labels depend on each of these parameters, and how far the compute budget can be reduced before the labels degrade, are measured in \cref{app:parameter_sensitivity,app:cost_quality}.

\section{Benchmark Data Release}
\label{benchmark_data_release}

\subsection{Composition}
\label{composition}

\begin{wrapfigure}[24]{r}{0.4\textwidth}
    \vspace{-2em}
    \includegraphics[width=\linewidth]{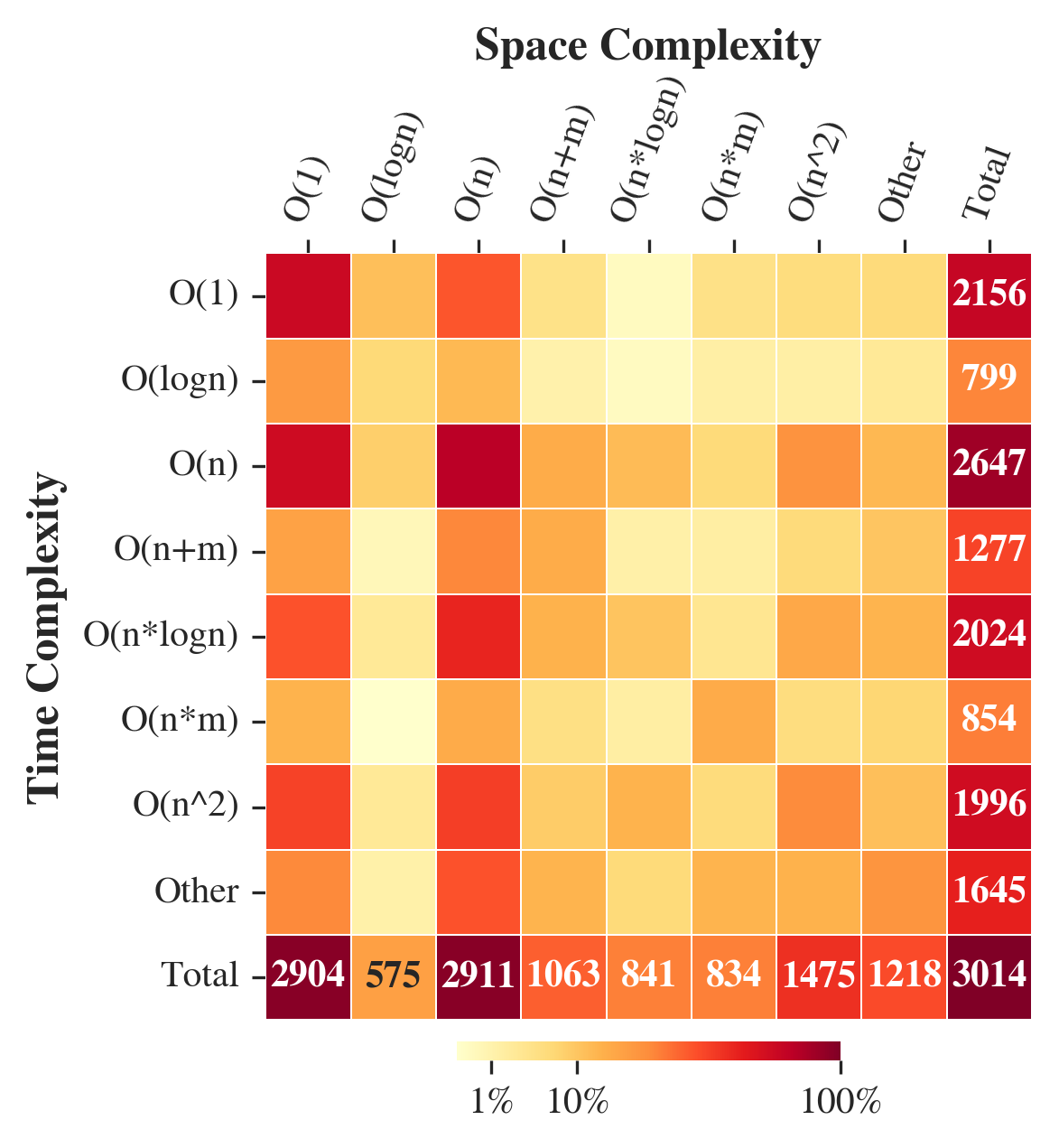}
    \caption{\looseness=-1 Distribution of time-space complexity classes across \benchmark dataset of 3,105 coding problems. Each problem is included when at least one solution exists with that specific time-space complexity pair.
    The chart orders classes by computational efficiency, with less common classes grouped under ``other’’.
    }
    \label{fig:data_composition}
\end{wrapfigure}

\codeforces is an online competitive coding platform that gathers challenging problems to be solved in various programming languages. Humans can submit candidate solutions that are rewarded more the faster and more memory-efficient they are. Using coding problems and solutions from \codeforces mostly (and in minority from a few other coding platforms), \vendor \citep{Li_2022} is a dataset that provides the problem descriptions along with correct and incorrect human solutions.

We annotated data from \vendor for both time and space complexity to create \benchmark using the complexity framework described in \cref{complexity_prediction_framework}.
\vendor data was limited to correct solutions (according to public and private tests) written in Python code only, which sum up to 8,139 problems and 1,485,888 solutions in total.
In addition, problems that have no working dataclass (see \cref{dataclass_generation}), too few solutions (fewer than 50) or unusual data types are also filtered out.
This finally leads to our general \benchmark dataset, consisting of 3,105 coding problems and 1,190,250 annotated solutions. Problems are characterized by their difficulty level (A: 942; B: 682; C: 427; D+: 321; Unknown: 733), their algorithmic notions (37 different notions in total) and their inputs (1 to 11 distinct arguments across 32 different data types).

Each solution is annotated by the complexity framework and associated with time and space complexity classes, the corresponding coefficients of the complexity curves and the runtime/memory-footprint measures that were used to infer these attributes. Therefore, each problem gets associated with one or several time-space complexity classes consisting of solutions that have various complexity coefficients, corresponding to different optimization tricks within the same class of complexity. Dataclasses generated for each problem are also included (see \cref{dataclass_generation}).

\begin{wrapfigure}[45]{r}{0.4\textwidth}

    \begin{minipage}{0.4\textwidth}
    \captionof{table}{Comparison of models for generating problem-specific dataclasses that can parse the incoming input streams into each problem's variables, on \vendor. All models but CodeLlama 70B Instruct (16k only) \citep{rozière2024codellamaopenfoundation} use a context window of 32k tokens.}
    \vspace{-0.5em}
    \label{tab:dataclass_generation}
    \begin{center}
    \begin{small}
    \begin{sc}
    \resizebox{\linewidth}{!}{
    \begin{tabular}{lcc}
    \toprule
    \textbf{Model} & \textbf{Corr$@10$} & \textbf{BckTr$@10$} \\
    \midrule
    Codestral 22B & 63.6 & 54.0 \\
    CodeLlama 34B Instruct & 22.1 & 17.8 \\
    CodeLlama 70B Instruct & 10.3 & 7.9 \\
    Llama 3.1 8B Instruct & 31.9 & 21.4 \\
    Llama 3.1 405B Instruct & \textbf{70.2} & \textbf{58.1} \\
    \bottomrule
    \end{tabular}
    }
    \end{sc}
    \end{small}
    \end{center}
    \end{minipage}
    \begin{minipage}{0.4\textwidth}
    \begin{algorithm}[H]
        \captionsetup{type=algorithm}  %
        \caption{Dataclass Template}
        \label{alg:dataclass_template}
        \begin{lstlisting}[
            language=Python,
            basicstyle=\tiny\ttfamily,
            aboveskip=0pt,
            belowskip=0pt
        ]
@dataclass
class Input:
    var_1: type_1
    var_2: type_2
    ...
    @classmethod
    def from_str(cls, input_: str):
        ...
        return cls(var_1, var_2, ...)
    def __repr__(self):
        ...
        return input_
        \end{lstlisting}
    \end{algorithm}
    \end{minipage}

    \begin{minipage}{0.4\textwidth}
    \centering

    \vspace{0.2em}
     \resizebox{0.99\linewidth}{!}{
     \centering
     \includegraphics[width=\linewidth]{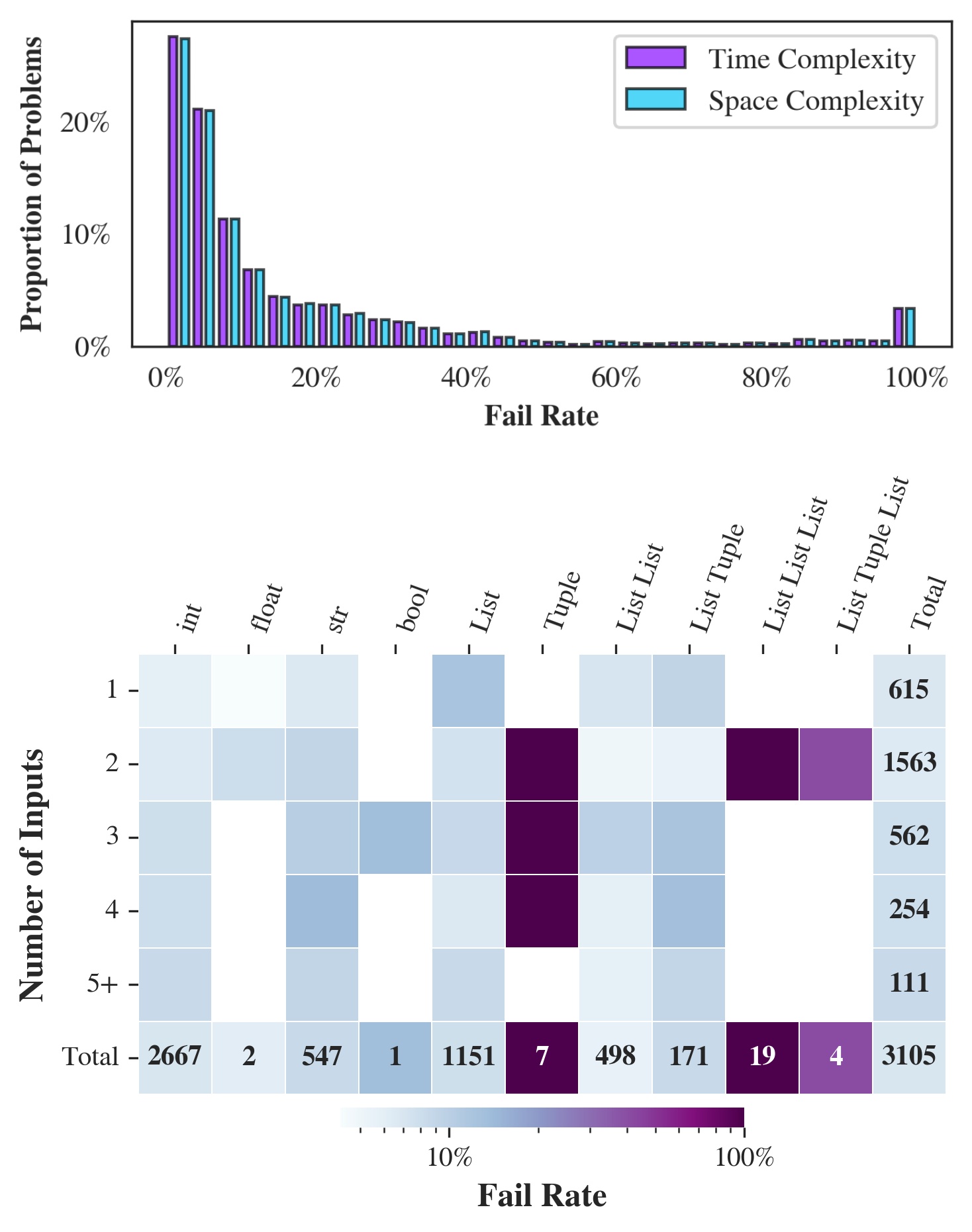}
    }
    \caption{
    Failure rate analysis of the complexity inference framework. The top plot shows the overall distribution of framework failures across all problems. The bottom heatmap breaks down failure rates by input type and number of distinct inputs.
    \vspace{-0.12in}
    }
    \label{fig:error_analysis}
    \vskip -0.05in
    \end{minipage}

\end{wrapfigure}

Time/space complexity test sets are selected among this global pool of problems/solutions by executing a range of post-processing and filtering steps, to provide meaningful metrics by enforcing diversity of classes as well as performance and stability of the framework on the problems. For instance, only problems with several complexity classes are kept; absolute and relative thresholds (to the most popular class of the problem) filter out outliers; complexities with unlikely variable counts or high failure rates are removed; finally solutions with unlikely abstract trees (relative to their complexity) or unstable predictions are withdrawn.
The resulting test sets have distinct supports of problems (though there is an overlap of 63 problems), since few problems have diverse classes in both time and space.
Moreover they ignore official \vendor splits, as the \vendor test set lacks sufficient discriminative power, and lacks problems with multiple complexity classes.

The time complexity test set is made out of 311 problems and 640 corresponding solutions covering 11 different classes (the most represented ones being $O(n)$, $O(n \cdot \mathrm{log}(n))$, $O(n^2)$, $O(1)$, $O(n\times m)$ and the least represented $O((n+m)\mathrm{log}(n+m))$). The space complexity test set consists in 308 problems and 636 solutions covering 5 different classes (by order of popularity $O(n)$, $O(1)$, $O(n^2)$, $O(n+m)$, $O(n\times m)$). A training split for fine-tuning purposes is also included.
\cref{fig:data_composition} shares more details about the distribution of complexity classes in the dataset. It is imbalanced and heavily tailed: linear time complexity represents 38\% of all solutions, constant time complexity 20\%; for space complexity, distribution is even more skewed with respectively 47\% and 25\%.

\subsection{Dataclass Generation}
\label{dataclass_generation}

To infer labels with the complexity framework on \vendor code snippets, code inputs must be parsable into a dataclass matching \cref{alg:dataclass_template}. We define the task of dataclass generation as querying a LLM for such dataclass given the problem description and an example solution. For each generated dataclass \texttt{Input}, we introduce two metrics that measure the quality of the dataclass methods \texttt{Input.from\_str} (converting string stream inputs to argument dictionaries) and \texttt{Input.\_\_repr\_\_} (the reverse conversion). These dataclasses let the framework expand selected arguments and rerun the original solution on new input scales, instead of being limited to raw \vendor input strings. They also make input parsing a separately measured bottleneck: when no correct dataclass can be produced, the problem cannot be used for dynamic complexity inference, even if many accepted Python solutions are available.

Performance is measured by \textsc{\textbf{Corr}}, that accounts for the correction (executability) of the methods \texttt{Input.from\_str} and \texttt{Input.\_\_repr\_\_}, and \textsc{\textbf{BckTr}}, which measures accuracy of the backtranslation \citep{edunov2018understandingbacktranslationscale}:

~~~~\texttt{\small input\_ == Input.from\_str(input\_).\_\_repr\_\_()}

\textsc{\textbf{Corr@k}} and \textsc{\textbf{BckTr@k}} are unbiased estimators of performance of \textsc{\textbf{Corr}} and \textsc{\textbf{BckTr}} among k samples, following the definition of \citet{chen2021evaluating}. Mathematical definitions of these metrics are provided in \cref{dataclass_metrics}.
\cref{tab:dataclass_generation} sums up the benchmark results of the dataclass generation task. \textsc{Llama 3.1 405B Instruct} \citep{dubey2024llama3herdmodels} reaches best performance, capable of 58.1\% correct backtranslation for one dataclass out of ten attempts per problem. To further boost performance, several passes corresponding to different solutions per problem are performed, thus generating a correct dataclass for 82\% of \vendor problems.

\subsection{Complexity Framework Performance}
\label{complexity_framework_performance}

\textbf{Accuracy}~~A human review measured the accuracy of the labels as output by the complexity framework (after post- processing and filtering) compared to the labels assigned by a human. The framework achieves 84\% and 82\% accuracy on time and space complexity test sets respectively (125 sample split each). \Cref{app:label_error} measures how much an error rate of this size can affect the comparison between models.

\textbf{Coverage}~~\cref{fig:error_analysis} measures the ratio of solutions per problem for which the framework fails to predict a label. Whatever the type of complexity, approximately 84\% of problems have a fail rate  below 30\%, and only 4.5\% of problems have a fail rate above 0.9, for reasons ranging from incorrect generated dataclass to an edge case not covered by the range of tests the framework performs. Most input types are correctly covered by the framework. Exceptions (e.g. tuple and triple-nested lists) are infrequent.

\textbf{Self-consistency}~~Relying on empirical runtime and memory measures exposes the framework to stochastic noise affecting prediction reliability. Running the framework 20 times on 10 solutions of every problem and complexity class of the candidate test set, before any filtering based precisely on stability, 91.9\% (resp.~89.1\%) self-consistency is achieved for time (resp.~space) complexity, for a total of 10,130 (resp.~10,520) different code solutions. \Cref{app:label_error_rate} shares more details about the different error rates, including the human-based review, this self-consistency number and the reproducibility of the framework verdicts on model-generated code.

\section{Evaluation}
\label{evaluation_section}

\begin{table*}[t!]
\centering\small
\caption{
\benchmark benchmark results for popular LLMs.  \textbf{Program Synthesis} checks the correctness of model-generated solutions to given programming problems, not taking into account any complexity requirement. \textbf{Complexity Prediction} measures whether a model can find the time-space complexity of an existing code snippet. \textbf{Complexity Generation} evaluates whether a model can output a working code snippet to a given problem that meets a time-space complexity requirement. $Pass@k$ considers each complexity class of all problems independently and calculates a macro-average between them. $Best@k$ is a refinement of $Pass@k$, focusing only on the most optimized complexity class for each problem. $All@k$ checks if all complexity classes for each problem are correctly predicted or generated simultaneously, then macro-averages across all problems.
}
\label{table:complexityeval}
\begin{sc}
\resizebox{\linewidth}{!}{
\begin{tabular}{llcclllllllll}
\toprule
\multirow{3}{*}{\textbf{Model}} & ~ & \multicolumn{2}{c}{\textbf{Prog. Synthesis}} & ~ & \multicolumn{3}{c}{\textbf{Complexity Pred.}} & ~ & \multicolumn{4}{c}{\textbf{Complexity Gen.}} \\
 & & \multicolumn{1}{c}{Pass} & \multicolumn{1}{c}{Pass} & & \multicolumn{1}{c}{Pass} & \multicolumn{1}{c}{Best} & \multicolumn{1}{c}{All} & & \multicolumn{1}{c}{Pass} & \multicolumn{1}{c}{Pass} & \multicolumn{1}{c}{Best}& \multicolumn{1}{c}{All} \\
 & & \multicolumn{1}{c}{\bf $@1$} & \multicolumn{1}{c}{\bf $@10$} & & \multicolumn{1}{c}{\bf $@1$}  &\multicolumn{1}{c}{\bf $@1$}&\multicolumn{1}{c}{\bf $@1$}& & \multicolumn{1}{c}{\bf $@1$} & \multicolumn{1}{c}{\bf $@10$} &\multicolumn{1}{c}{\bf $@1$} & \multicolumn{1}{c}{\bf $@1$} \\
\hline\vspace{-0.1in} \\
\textbf{Time} & & \\
~~Baselines & & 30.3 & 55.4 & & 39.5 & 68.5 & 0.0 & & 12.1 & 29.7 & 19.0 & 0.9 \\
~~Llama 3.3 70B & & 43.4 & 66.0  & & 58.2 & 72.6 & 33.7 & & 17.7 & 40.0 & 25.7 & 3.3 \\
~~Llama 4 Scout 17Bx16E & & 61.7 & 80.1 & & 48.7 & 66.3 & 23.0 & & 22.8 & 48.0 & 31.8 & 3.5 \\
~~Llama 4 Maverick 17Bx128E & & 59.3 & 78.9 & &  57.4 & 70.8 & 32.8 & & 19.9 & 44.6 & 27.0 & 5.3 \\
~~Gemma 3 27B & & 37.7 & 45.6 & & 60.8 & 69.2 & 37.6 & & 15.1 & 20.9 & 17.8 & 1.8 \\
~~Codestral 22B & & 23.7 & 47.5  & & 56.0 & 67.8 & 33.5 & & 10.6 & 26.6 & 14.9 & 1.3 \\
~~Qwen2.5-Coder 32B & & 30.5 & 50.8 & & 58.5 & 68.2 & 34.9 & & 12.2 & 26.5 & 15.2 & 3.1\\
~~GPT-4o & & 51.0 & 78.3  & & 57.7 & 69.7 & 33.1 & &  20.6 & 44.7 & 30.2 & 4.3 \\
~~O1-mini & & 62.5 & 76.8 & & 58.3 & 65.2 & 35.6 & & 19.8 & \textbf{65.2} & 27.6 & 4.5 \\
~~DeepSeekCoderV2 236B & & 44.1 & 65.5 & & 54.9 & 68.9 & 29.6 & & 19.5 & 38.0 & 27.6 & 3.3 \\
~~DeepSeekV3 671B & & 41.4 & 63.6 & & 54.4 & 72.4 & 27.1 & & 17.7 & 37.7 & 23.0 & 3.4 \\
~~DeepSeekR1 Qwen 32B & & 70.1 & 83.7 & & 62.2 & 72.7 & 41.1 & & 29.0 & 49.9 & 46.1 & 4.8 \\
~~DeepSeekR1 Llama 70B & & 70.1 & 83.8 & & \textbf{64.2} & \textbf{75.4} & \textbf{41.4} & & 29.2 & 51.6 & 46.5 & 4.8 \\
~~Llama 3.1 Nemotron-Ultra 253B & & \textbf{80.0} & \textbf{89.4} & & 63.1 & 74.5 & 41.1 & & \textbf{33.5} & 54.9 & \textbf{51.8} & 6.1 \\
~~Qwen3 32B & & 70.0 & 81.3 & & 61.3 & 70.5 & 39.0 & & 29.1 & 53.8 & 43.5 & \textbf{6.5} \\
\hline\vspace{-0.1in} \\
\textbf{Space} & & \\
~~Baselines & & 30.1 & 52.6 & & 45.4 & 50.3 & 0.0 & & 12.2 & 32.4 & 17.8 & 1.3 \\
~~Llama 3.3 70B & & 42.6 & 62.5 & & 41.1 & 55.2 & 10.9 & & 15.0 & 37.7 & 21.9 & 1.8 \\
~~Llama 4 Scout 17Bx16E & & 56.7 & 73.5 & & 37.4 & 58.8 & 5.2 & & 20.0 & 40.1 & 31.1 & 2.2 \\
~~Llama 4 Maverick 17Bx128E & & 58.4 & 75.7 & & 44.6 & 54.5 & 8.9 & & 16.8 & 28.2 & 30.4 & 0.8 \\
~~Gemma 3 27B & & 40.3 & 49.0 & & 44.8 & 62.9 & 13.2 & & 16.2 & 24.3 & 22.5 & 1.4 \\
~~Codestral 22B & & 25.7 & 47.6 & & 44.3 & 62.5 & 10.6 & & 11.0 & 29.4 & 16.7 & 1.3 \\
~~Qwen2.5-Coder 32B & & 31.1 & 49.2 & & 45.6 & \textbf{63.4} & 12.6 & & 10.1 & 23.3 & 15.3 & 1.2 \\
~~GPT-4o & & 51.6 & 74.4 & & 43.4 & 61.4 & 11.0 & & 18.1 & 39.9 & 28.0 & 1.4 \\
~~O1-mini & &  58.0 & 72.9 & & 42.7 & 45.6 & 8.1  & &  16.6 & \textbf{61.3} & 25.7 & 2.5 \\
~~DeepSeekCoderV2 236B & & 43.1 & 63.8 & & 44.1 & 59.6 & 8.2 & & 16.7 & 34.5 & 25.6 & 1.0 \\
~~DeepSeekV3 671B & & 41.8 & 62.5 & & 43.5 & 62.6 & 11.2 & & 15.0 & 35.4 & 22.6 & 1.6 \\
~~DeepSeekR1 Qwen 32B & & 68.0 & 80.6 & & 43.2 & 55.0 & 8.1 & & 24.8 & 48.6 & 38.6 & 3.1 \\
~~DeepSeekR1 Llama 70B & & 68.8 & 81.2 & & 44.4 & 56.1 & 10.4 & & 25.6 & 50.0 & 38.7 & 3.4 \\
~~Llama 3.1 Nemotron-Ultra 253B & & \textbf{77.7} & \textbf{86.4} & & 45.2 & 54.7 & 10.3 & & \textbf{30.4} & 55.5 & \textbf{45.3} & \textbf{5.6} \\
~~Qwen3 32B & & 65.9 & 77.7 & & \textbf{47.7} & 58.3 & \textbf{15.1} & & 25.5 & 47.8 & 39.6 & 5.1 \\
\bottomrule
\end{tabular}
}
\vspace{-1em}

\end{sc}\end{table*}
We use \benchmark to evaluate several LLMs commonly used for coding and reasoning tasks: \textsc{Llama 3.3 70B} \citep{dubey2024llama3herdmodels}, \textsc{Llama 4} models \citep{meta2025llama}, \textsc{Gemma 3 27B} \citep{gemmateam2025gemma3technicalreport}, \textsc{Codestral 22B} \citep{mistralai2024codestral}, \textsc{GPT-4o} \citep{openai2024gpt4technicalreport}, \textsc{o1-mini} \citep{openai2024openaio1card}, \textsc{Qwen 2.5-Coder 32B} \citep{hui2024qwen25codertechnicalreport}, \textsc{DeepSeek-Coder-V2 236B} \citep{deepseekai2024deepseekcoderv2breakingbarrierclosedsource}, \textsc{DeepSeek-V3 671B} \citep{deepseekai2024deepseekv3technicalreport}, \textsc{DeepSeek-R1 Qwen} and \textsc{Llama} distilled \citep{deepseekai2025deepseekr1incentivizingreasoningcapability}, \textsc{Llama 3.1 Nemotron-Ultra} \citep{bercovich2025llamanemotronefficientreasoningmodels} and \textsc{Qwen3 32B} \citep{qwen3technicalreport}. All models are evaluated using their \textsc{instruct} variant, when available, in a zero-shot fashion (unless otherwise stated). \textsc{GPT4-o} and \textsc{o1-mini} do not share any estimate on inference compute. Also, \textsc{o1-mini} returned many empty answers, potentially due to reasoning collapse: we discarded these answers and used only non-empty answers to compute metrics. As a result, its performance can be regarded as an upper-bound optimistic estimate. \textsc{DeepSeek-R1} distilled models used substantially more compute than \textsc{Llama 4} ($\times2$ compute nodes, $\times5$ compute time and $\times16$ generation tokens).

Pure program synthesis performance is also displayed on the same test splits as the rest of the metrics. It is evaluated for $pass@k$ using all public, private and generated tests.  For each metric, best values or any values not significantly lower than the best are displayed in boldface. Metrics are macro-averaged first by complexity classes within each problem and then across problems. More details and metric definitions are provided in \cref{format_metrics}. \footnote{One-tailed paired t-tests on 1000 bootstraps samples of the model results evaluate the significance of the superiority of the best model. Any $@k$ metric uses an unbiased estimator based on 20 samples.}

\subsection{Time-Space Complexity Prediction}
\label{results:first_task}

The first evaluation task of \benchmark, \textbf{Complexity Prediction}, consists in predicting the time and space complexity given a problem description and a human solution. Our baseline for this task is the naive model that always returns $O(n)$, the most frequent class. $Pass@k$ measures the accuracy of finding the correct complexity, using a parsing script that compares the output of the LLM with the ground-truth complexity inferred by the framework; $Best@k$ measures accuracy only across the most optimized complexity class of each problem; $All@k$ requires correct complexity output across all complexity classes at once per problem: the LLM has to correctly output a working solution that meets the complexity requirement for all classes of complexity of the problem. For each metric, $@k$ is the unbiased estimator among k samples, following the definition of \citet{chen2021evaluating}. Metrics are macro-averaged across complexity classes ($Pass@k$), and then across problems ($Pass@k$, $Best@k$ and $All@k$).

Results are displayed in \cref{table:complexityeval}. A query example, along with an output example of \textsc{DeepSeek-R1 Llama 70B}, is provided in \cref{appendix:complexity_prediction_example}. More metric definitions are detailed in \cref{complexity_metrics}.

\subsection{Time-Space Complexity Code Generation}
\label{results:second_task}

The second task \textbf{Complexity Generation} requires the LLM to generate a correct solution to a given problem description that has to respect a feasible time or space complexity requirement. Our baseline for this task is a \textsc{Llama 3.1 70B} model that is queried for the same prompts without the complexity requirement. $Pass@k$ measures the accuracy of finding a correct solution, according to public, private and generated tests, that has the correct complexity, as measured by the complexity framework; $Best@k$ and $All@k$ are similarly defined as their counterparts of \cref{results:first_task}. Results are displayed in \cref{table:complexityeval}. An example with \textsc{DeepSeek-R1 Llama 70B} is provided in \cref{appendix:complexity_generation_example}.

\subsection{Time-Space Complexity Coefficient Percentile Ranking}
\label{results:third_task}

The third task, \textbf{Complexity Coefficient Percentile Ranking}, measures how a generated solution to a given problem, respecting a complexity requirement, ranks among human solutions of the same complexity class and problem. The ranking is performed based on the coefficient of the complexity curve, as measured by the framework: the lower the coefficient, the more flat the complexity curve and the more optimized the solution. Ranking results are given in percentile of the distribution, where a solution of the $nth$ percentile is more optimized than $n\%$ of human solutions. The querying is similar to \cref{results:second_task} with the addition of the requirement "Try to optimize the runtime of your code as much as you can, while respecting the time complexity requirement". See \cref{table:coefficient_ranking}.

\begin{table}[t]
\caption{
Using the complexity framework, the best measured coefficient of the complexity curve, out of 20 attempts, is used to rank LLM-generated code among human solutions from the same problem and time-space complexity class. Ranking is percentile based, n\% ranking score amounts for n\% human solutions having worse complexity coefficient. If no LLM solution passes correctness tests, ranking score is set to 0. \textsc{Intersec} is the subset where all starred (*) models have at least one successful solution.}
\label{table:coefficient_ranking}
\centering\small
\begin{sc}
\resizebox{0.55\textwidth}{!}{
\begin{tabular}{lccc}
\toprule
\multirow{3}{*}{\textbf{Model}} & \multicolumn{2}{c}{\textbf{Coefficient}} & ~ \\
 & \multicolumn{2}{c}{\textbf{Ranking}} & All \\
& \multicolumn{1}{c}{Full} & \multicolumn{1}{c}{Intersec} & $@1$ \\
\hline  \vspace{-0.1in} \\
\textbf{Time} & & \\
~~Llama 3.3 70B & 33.8 & 65.0 & 2.8 \\
~~Llama 4 Scout 17Bx16E & 39.0 & 69.9 & 3.4 \\
~~Llama 4 Maverick 17Bx128E & 43.2 & 72.1 & 3.6 \\
~~Gemma 3 27B & 10.4 & 25.7 & 1.6 \\
~~Codestral 22B & 21.6 & 50.7 & 1.5 \\
~~Qwen2.5-Coder 32B & 19.7 & 50.4 & 2.2 \\
~~GPT-4o\text{*} & 36.6 & 70.9 & 4.2 \\
~~O1-mini\text{*} & 26.3 & 78.8 & 3.1 \\
~~DeepSeekCoderV2 236B & 27.7 & 54.1 & 2.8 \\
~~DeepSeekV3 671B & 28.7 & 58.6 & 3.4 \\
~~DeepSeekR1 Qwen 32B\text{*} & 38.6 & 79.0 & 4.2 \\
~~DeepSeekR1 Llama 70B\text{*} & 38.3 & 79.0 & 4.0 \\
~~Llama 3.1 Nemotron-Ultra 253B\text{*} & 41.6 & 75.8 & 5.1 \\
~~Qwen3 32B\text{*} & \textbf{44.0} & \textbf{79.6} & \textbf{6.1} \\
\hline\vspace{-0.1in} \\
\textbf{Space}  & & \\
~~Llama 3.3 70B & 32.8 & 73.0 & 1.6 \\
~~Llama 4 Scout 17Bx16E & 34.5 & 78.8 & 2.1 \\
~~Llama 4 Maverick 17Bx128E & 21.8 & 60.0 & 0.8 \\
~~Gemma 3 27B & 17.5 & 49.3 & 1.4 \\
~~Codestral 22B & 25.2 & 64.6 & 1.2 \\
~~Qwen2.5-Coder 32B & 20.5 & 68.7 & 0.6 \\
~~GPT-4o\text{*} & 31.6 & 86.3 & 1.3 \\
~~O1-mini\text{*} & 21.1 & 82.8 & 1.5 \\
~~DeepSeekCoderV2 236B & 26.8 & 68.5 & 1.2 \\
~~DeepSeekV3 671B & 27.2 & 72.3 & 1.3 \\
~~DeepSeekR1 Qwen 32B\text{*} & 40.1 & 88.5 & 3.0 \\
~~DeepSeekR1 Llama 70B\text{*} & 41.6 & \textbf{89.5} & 3.3 \\
~~Llama 3.1 Nemotron-Ultra 253B\text{*} & \textbf{45.4} & 87.7 & \textbf{4.6} \\
~~Qwen3 32B\text{*} & 40.5 & 88.2 & 4.3 \\
\bottomrule
\end{tabular}
}
\end{sc}
\end{table}

\subsection{Prediction and Generation Fine-Tuning}

Using training sets of ~2000 problems and ~20k code solutions, \textsc{Llama 3.1 70B} is being fine-tuned for the first and second task. Time and space generation training sets sums up to ~22M tokens, prediction training sets to ~18-19M. Each complexity class of each problem includes 10 human examples, filtered following the same steps as the test set creation (see \cref{composition}). Models are fine-tuned for 10 epochs in instruct format. See \cref{table:finetuning_experiments}.

\section{Quantitative Analysis}
\label{quantitative_analysis}

\begin{wrapfigure}[43]{r}{0.5\textwidth}
    \begin{minipage}{0.5\textwidth}
    \vspace{-1em}
    \captionof{table}{\benchmark benchmark results for fine-tuned \textsc{Llama 3.1 70B} on time-space prediction and generation tasks. Same metrics as in \cref{table:complexityeval}.}
    \label{table:finetuning_experiments}
    \centering\small
    \begin{sc}
    \resizebox{\linewidth}{!}{
    \begin{tabular}{lcc@{}cc@{}c}
    \toprule
    \multirow{3}{*}{\textbf{Method}} & \multicolumn{1}{c}{\textbf{Prog.}} & \multicolumn{2}{c}{\textbf{Prediction}} & \multicolumn{2}{c}{\textbf{Generation}} \\
    & \multicolumn{1}{c}{\textbf{\textbf{Synth.}}} & \multicolumn{1}{c}{\textbf{Time}} & \multicolumn{1}{c}{\textbf{Space}} & \multicolumn{1}{c}{\textbf{Time}} & \multicolumn{1}{c}{\textbf{Space}} \\
     & Pass$@1$ & All$@1$ & ~All$@1$ & All$@1$ &  ~All$@1$
    \\ \hline \vspace{-0.1in}\\
    Zero-shot & \textbf{29.6} & 33.8 & 11.9 & \textbf{3.1} & \textbf{1.8} \\
    Few-shot & 28.9 & 33.6 & 12.1 & 2.4 & 1.4 \\ \vspace{-0.1in} \\
    \multicolumn{6}{l}{\textbf{Prediction Fine-tuning}} \\
    ~~~~Time & 27.4 & \textbf{36.5} & 6.6 & 2.9 & 1.3 \\
    ~~~~Space & 26.6 & 9.0 & \textbf{14.0} & 2.4 & 1.4 \\ \vspace{-0.1in} \\
    \multicolumn{6}{l}{\textbf{Generation Fine-tuning}} \\
    ~~~~Time & 23.2 & 34.7 & 12.7 & 1.2 & 1.3 \\
    ~~~~Space & 23.4 & 34.6 & 13.0 & 1.5 & 1.4 \\
    \bottomrule
    \end{tabular}
    }
    \end{sc}

\end{minipage}
\begin{minipage}{0.5\textwidth}
     \vspace{1em}
    \includegraphics[width=\linewidth]{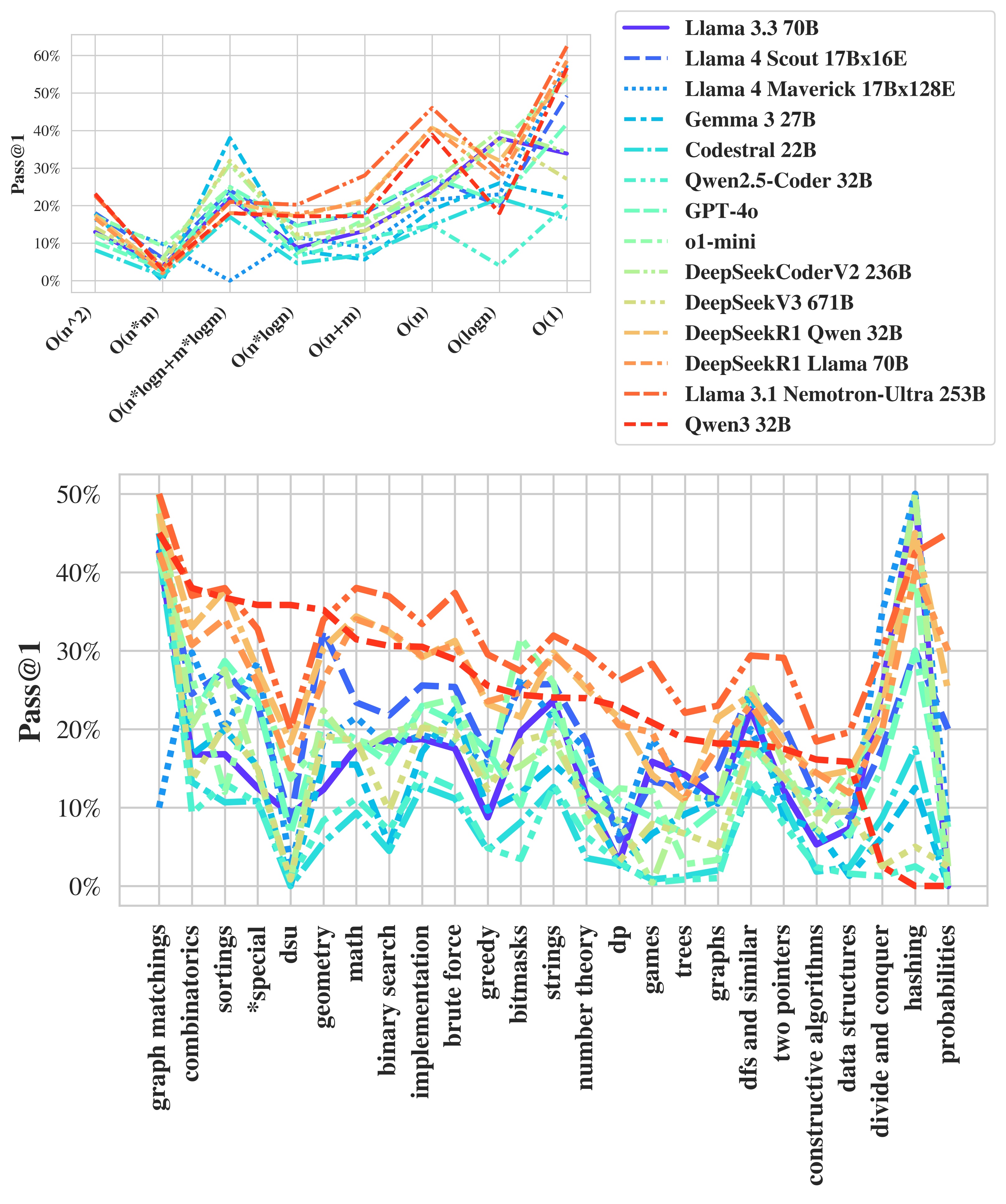}
    \vskip -0.1in
    \caption{LLM results aggregated by time complexity and algorithmic notions. Scores are $Pass@1$ on Time Complexity Generation.}
    \label{fig:success_per_metadata}
    \vskip -0.1in
\end{minipage}
\end{wrapfigure}

\textbf{Understanding time-space complexity}~~~Per \cref{table:complexityeval}, all LLMs show performance drops on the combined task \textbf{Complexity Generation} versus individual tasks \textbf{Program Synthesis} and \textbf{Complexity Prediction}. Across all tasks, the top performing models remain the reasoning models \textsc{DeepSeek-R1 Llama 70B}, \textsc{Qwen3 32B} and \textsc{Llama Nemotron-Ultra}, achieving 64.2 and 33.5 $Pass@1$ for time prediction and generation, except space prediction, where performance patterns are less clear with smaller differences, as these models tend to overthink extra space complexity despite explicit prompts. Models tend to be even more misled when asked to \textit{"Optimize the solution while respecting the complexity requirement"}, causing 12\% average loss in time generation $All@1$ (\cref{table:coefficient_ranking}), reaching ~30\% for \textsc{GPT-4o} and \textsc{o1-mini}. \Cref{app:generation_failures} decomposes these generation failures into their coding and complexity components.

At a more granular level, models tend to under-perform on non-optimal complexity classes, compared to the most optimized class of every problem, as underlined in \cref{fig:success_per_metadata}. This contradicts human programming patterns, where non-optimized solutions are typically easier than optimal ones, especially for competition problems. In addition, LLMs do not understand that adding dummy pieces of code (e.g., list sorting) could transform a, working linear solution into a less-optimized linearithmic one. In the end, $All@1$ metrics better capture true understanding by uniformly evaluating across optimized and non-optimized solutions, verifying that LLMs do not just stumble upon a solution of the right complexity because they learned by heart the widespread optimized code snippets. Top model \textsc{Qwen3 32B} achieves only 6.5 $All@1$ on time generation. These complexity metrics can also be used as a proxy of the understanding of other notions, such as combinatorics, where models explicitly optimized for math reach score higher (\cref{fig:success_per_metadata}).
When analyzing these results, one has to keep in mind that the splits were done on the training data of \vendor that includes all solutions of the different complexity classes to all problems, already seen by models.

\textbf{Token-space reasoning models}~~~Though they largely outperform other LLMs on pure program synthesis, reasoning models such as \textsc{o1-mini}, \textsc{DeepSeek-R1} and \textsc{Llama Nemotron-Ultra} are much closer in terms of performance on complexity-related tasks, and these are even outperformed by \textsc{Qwen2.5-Coder} on space complexity prediction specifically: while the latter obtains 12.6 $All@1$, the former respectively only reach 8.1, 10.4 and 10.3. For this specific case, reasoning models seem to exhibit patterns of overthinking, misunderstanding the notion of extra space complexity, even if clearly described in the prompt. When explicitly prompted in order to understand their failure modes, it turns out that such models are able to recognize all classes of complexity of a problem from a prediction point of view, but when asked to generate them, they fail to return the less optimized classes of complexity, favoring the optimal solution. Not to mention that these models could just easily `cheat' by tweaking the optimal solution with a dummy sort or nested for-loop, so to transform it, as any programmer could figure it out, as a less-than-optimal solution.

In general, these models struggle with the ambiguity of higher-level reasoning tasks, especially when there is no explicit verifier that they may have been confronted with during their reinforcement, such as programming tests to pass. This triggers the question of whether they really understand how to `think' about notions they `know', or if they only learn by heart patterns of `thoughts' exhibited by human annotation efforts and training rewards. As all these complexity solutions were, in fact, in their training data, it suggests that these highly efficient search models can still fail to recover the correct data points when they were not reinforced for the specific search criteria.

\begin{wrapfigure}[15]{r}{0.5\textwidth}
\vspace{-1.5em}
\begin{center}
\includegraphics[width=1\linewidth]{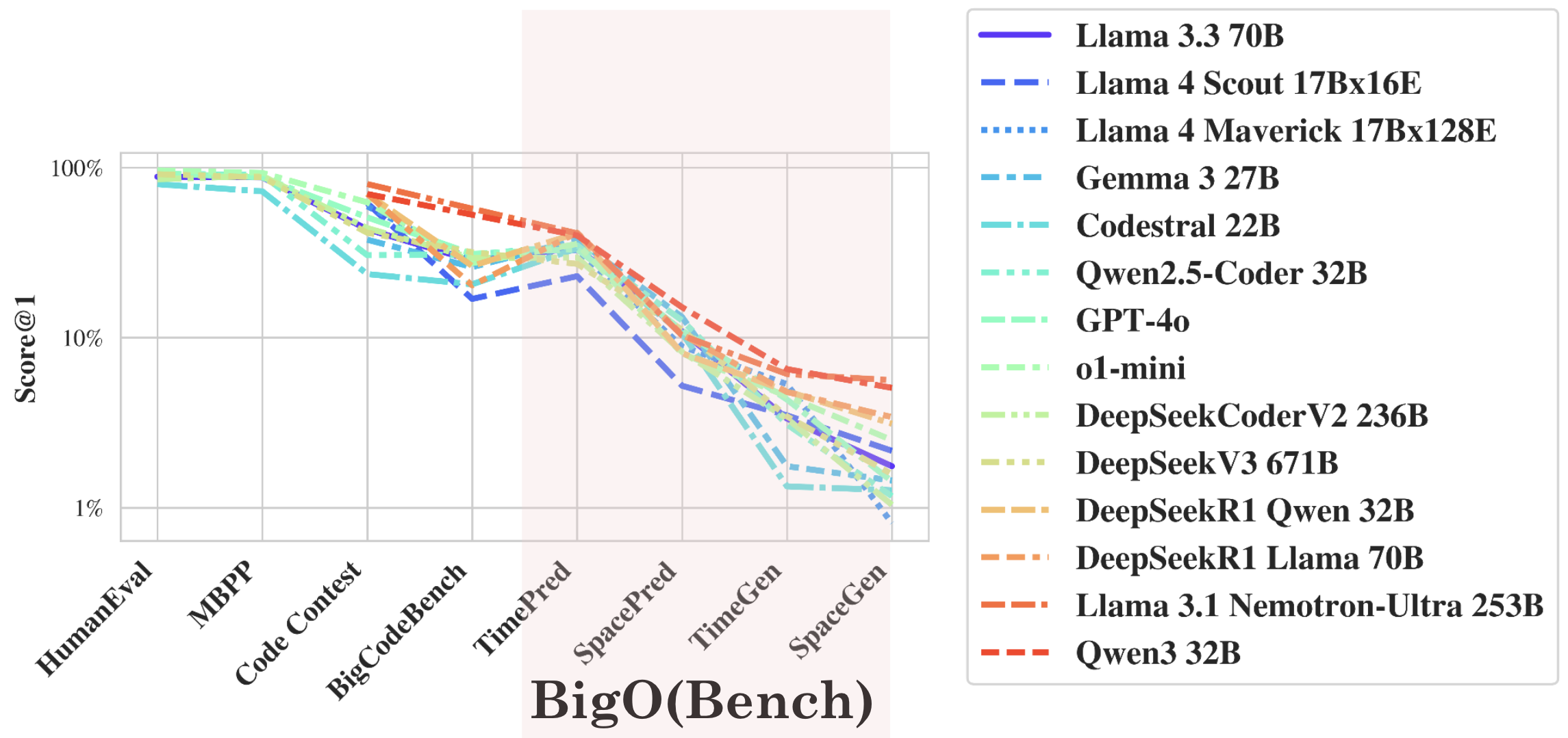}
\end{center}
\vspace{0.2em}
\caption{Model scores per coding benchmark: \textsc{HumanEval}, \textsc{MBPP} and \textsc{BigCodeBench} main metrics are all $Pass@1$; for \benchmark, we display $All@1$ results.}
\label{fig:benchmarking_benchmarks}
\vspace{0.4em}
\end{wrapfigure}

\textbf{Developing challenging reasoning benchmarks}\quad As newly released benchmarks usually quickly saturate, \benchmark aims at evaluating high-level reasoning skills that stay out-of-scope of current LLMs, bringing their performance down as displayed by \cref{fig:benchmarking_benchmarks}. \cref{table:finetuning_experiments} measures that the benchmark remains robust to fine-tuning. In particular, complexity prediction fine-tuning barely improves performance on the same task, and complexity generation fine-tuning even slightly hurt the performance of LLMs. This suggests that learning such high-level reasoning tasks is not effectively captured by standard fine-tuning to learn logical deduction and more convoluted patterns of thoughts.
With reasoning models topping benchmarks, new challenging benchmarks are perhaps more about finding out-of-distribution patterns of thinking rather than new data points of existing reasoning scenarios. It was especially difficult to design \benchmark without any human labels available, as the qualification level required from annotators was not reachable. But this is also probably why the models had been underexposed to this reasoning task, therefore making it challenging for them.

\textbf{Limitations}~~~The complexity framework itself is prone to errors, as for specific problems it can potentially fall upon worst-complexity edge cases. In addition, the measures on which the complexity prediction is based remain noisy, still relying on real CPU runtimes and using statistical measuring tools; they could potentially become more reliable with the help of virtual CPU cores. \Cref{app:label_error} quantifies how much this noise can affect the comparison between models, and \cref{app:parameter_sensitivity} how much the released labels depend on the framework parameters.

Although we fine-tune \textsc{Llama} models, we did not use advanced multiturn prompting \citep{zheng2024makeslargelanguagemodels} nor further reinforcement \citep{gehring2025rlefgroundingcodellms}. Proximal Policy Optimization \citep{schulman2017proximalpolicyoptimizationalgorithms} could help refine LLMs for these tasks.
Human annotations could also help models reason better for these tasks. Finally, the coding problems and the framework remain limited to Python. Mixing other languages such as C++ and Java could measure cross-languages optimization strategies.

\section{Conclusion}

In this work, we introduced \benchmark, a novel benchmark of LLMs on code generation, focusing on their understanding of time and space complexities when producing code. It consists in three tasks: first, given a coding challenge, predicting the time-space complexity of a given corresponding solution; second, for a given challenge and time or space complexity, generating a solution that solves the challenge while fulfilling the complexity requirements; third, optimizing the coefficient of the complexity cure compared to the human distribution. This benchmark is supported by time-space complexity labels corresponding to  3,105 coding problems and 1,190,250 corresponding solutions from \vendor. In addition, we developed a complexity framework capable of dynamically inferring the time-space complexity of a given snippet of code, used to automatically evaluate any synthetic snippet of code and therefore the generation of LLMs conditioned on a given complexity trade-off. Finally, we benchmark 14 LLMs considered as the top coding assistants and analyse their performance.

\benchmark is a challenging benchmark with current top scores belonging to the reasoning models \textsc{DeepSeek-R1 Llama 70B}, \textsc{Qwen3 32B} and \textsc{Llama Nemotron-Ultra}, achieving up to 6.5\% and 5.6\% $All@1$ on the time-space complexity generation tasks. Even when fine-tuning a \textsc{Llama 3.1 70B} model, performance increases only marginally, only on prediction tasks. We hope this benchmark can keep challenging upcoming models and help guide the development of new models towards better understanding of coding abstract notions beyond pure code generation.

\subsubsection*{Broader Impact Statement}

\benchmark does not include any model release, and therefore there is no risk for model misuse. The data used for \benchmark comes from an already existing public dataset \citep{Li_2022}. Any code execution is to be done within sandboxes, as documented in our implementation details and release materials, using the \textit{Bubblewrap} library \citep{bubblewrap}, to avoid any harmful side effects of the code being run.

Furthermore, we establish guidelines for users to follow when utilizing the benchmark, including the requirement to report any potential misuse or harmful applications, and to provide transparency in their use of the benchmark data and results. More details are provided in the Code of Conduct included in the release materials. We also establish a mechanism for users to report any concerns or issues related to the benchmark, and to provide feedback on how to improve the benchmark's safety and responsibility, on top of any potential security issues: this is detailed further in the security policy accompanying the released codebase as well.

The experiments reported in this paper used approximately 12,000 GPU hours and 180,000 CPU hours, as detailed in \cref{benchmark_experimental_setup}. This carries an environmental cost, and it puts a full reproduction out of reach of groups without comparable resources. This is one reason why we measure the trade-off between compute and label quality in \cref{app:cost_quality}: running the framework with a single measurement replica over the full input range costs a tenth of the compute and still reproduces 96.6\% of the released time complexity labels, which we consider a reasonable operating point for users who cannot afford the full setting. The GPU cost, on the contrary, is driven by the inference of the models being benchmarked, and scales with how many models a user chooses to evaluate.

The problems and human solutions come from \vendor \citep{Li_2022}. The licenses of the models we evaluate are listed in \cref{licence_section}. Because \benchmark is built on a public dataset that predates most of the models we evaluate, contamination is a real possibility, which we discuss in \cref{quantitative_analysis} and which any conclusion drawn from these scores should account for. As with any public leaderboard, repeated evaluation against the released test sets also carries a risk of overfitting to them, and we encourage users to be transparent about how many evaluation rounds a reported number reflects.

The same design choices also affect reproducibility. To ensure reproducibility of the work presented in this paper, we provide comprehensive details on the different steps of our study. The complexity framework implementation is detailed in \cref{complexity_prediction_framework} with technical specifics in \cref{implementation_details}. Our dataset construction and pre-processing procedures are documented in \cref{benchmark_data_release}. Benchmark design methodologies and theoretical foundations for our evaluation metrics are described in \cref{format_metrics}, while time-space complexity definitions are formalized in \cref{time_space_complexity_definitions}. Ablation studies validating our benchmark design choices are presented in \cref{ablations_prompts}. The robustness of the evaluation to label error is analysed in \cref{app:label_error}, the failure modes behind the gap between complexity prediction and generation in \cref{app:generation_failures}, and the sensitivity of the framework to its parameters and to its compute budget in \cref{app:parameter_sensitivity,app:cost_quality}.

The computational infrastructure and experimental setup used for our evaluations are documented in \cref{benchmark_experimental_setup}. Main experimental results are presented in \cref{evaluation_section}, with extended multi-sample analysis in \cref{multi_sample_results_scores}. Representative examples of benchmark prompts for both complexity prediction and generation tasks are provided in \cref{appendix:complexity_prediction_example,appendix:complexity_generation_example}. All open-source models and their repositories are listed in \cref{licence_section}, while we also evaluate several closed-source models accessible through vendor platforms. The released codebase, dataset, and interactive leaderboard are provided in the header links to facilitate future research.

\clearpage
\newpage

\bibliographystyle{assets/plainnat}
\bibliography{paper}

\clearpage
\newpage

\beginappendix

\section{Formal Metric Definitions}
\label{format_metrics}

\subsection{Dataclass Generation Metrics}
\label{dataclass_metrics}

In order to measure the quality of the dataclasses as generated by an LLM, we introduce two metrics \textsc{\textbf{Corr}} and \textsc{\textbf{BckTr}}, used over a test set of code problems and solutions to compare performance of LLMs on the dataclass generation task.

We introduce the following variables, and the structure of a problem dataclass:
\begin{itemize}
    \item \( P \) = Problem description
    \item \( S \) = Example solution
    \item \( D \) = LLM-generated dataclass (from \( P \) and \( S \)), that has the following class template (see \cref{alg:dataclass_template}):
\[
D = \begin{cases}
D\text{.from\_str}(\cdot) : \text{str} \to \text{dict} \\
D\text{.\_\_repr\_\_}() : \text{dict} \to \text{str}
\end{cases}
\]
    \item \( \text{input} \) = Input string, that corresponds to the input of an Input/Output test case pair of the code problem
    \item \( D.\text{from\_str}(\text{input}) \) = Dictionary of parsed arguments
    \item \( D.\_\_\text{repr}\_\_() \) = String representation of parsed arguments
    \item \( \mathcal{P} \): Distribution over problems \( P \)
    \item \( \mathcal{S}(P) \): Distribution over human solutions \( S \) for problem \( P \)
    \item \( \mathcal{D}_{LLM}(P, S) \): LLM's output distribution for dataclass \( D \) given \( P \) and \( S \).
    \item \( \mathcal{I}(P) \): Set of test case inputs for problem \( P \).
\end{itemize}

Based on these definitions, we define the following two metrics:
\begin{enumerate}
    \item \textbf{Correction (Corr)}: Executability of both methods
\[
\text{Corr} = \mathbb{E}_{P \sim \mathcal{P},\, S \sim \mathcal{S}(P),\, D \sim \mathcal{D}_{LLM}(P, S)} \mathbb{I}(D.\text{from\_str}(\cdot) \text{ and } D.\_\_\text{repr}() \text{ are executable})
\]
    \item \textbf{Backtranslation Accuracy (BckTr)}: Round-trip consistency
\end{enumerate}
\[
\text{BckTr} = \mathbb{E}_{P \sim \mathcal{P},\, S \sim \mathcal{S}(P),\, D \sim \mathcal{D}_{LLM}(P, S)} \mathbb{I}\left(\bigcap_{\text{input} \in \mathcal{I}(P)} \left( \text{input} == D.\text{from\_str}(\text{input}).\_\_\text{repr}() \right)\right)
\]

In practice, we approximate these metrics, especially as we operate under a fixed sampling budget for the LLM. We define a sampling budget parametrized by $n$, the number of times we sample a dataclass $D$ from $\mathcal{D}_{LLM}(P, S)$. Each problem is sampled with equal probability, and for the sake of limiting the compute budget only the first human solution of the attached set of solutions $\mathcal{S}(P)$ is used to compute the metrics. Then, for each  \( (P, S) \), we use the model to generate \( n \) samples \( D_1, \dots, D_n \sim \mathcal{D}_{LLM}(P, S) \).

We generalize the above metrics to the case of $k$-success, where we record the probability that at least one of $k$ generated dataclasses \( \{D_i\}_{i=1}^k \) satisfies the metric. We use the full compute budget of $n$ possible samples (in practice we take $n = 2 \times k$) to compute the unbiased $pass@k$ estimator from \citet{chen2021evaluating}.

This leads us to the following definitions of the estimators \textbf{Corr}$@k$ and \textbf{BckTr}$@k$:

\begin{enumerate}
    \item \textbf{Corr}$@k$ estimator that at least one of k LLM attempts satisfies dataclass correction
\[
\text{Corr} = \mathbb{E}_{\substack{P \sim \mathcal{P} \\ S_1 \in \mathcal{S}(P)}}\left[ 1 - \frac{\binom{n - c}{k}}{\binom{n}{k}} \right],
\quad \text{where }
\begin{cases}
c = \sum_{i=1}^n \mathbb{I}(\text{Corr is satisfied for } D_i), \\
D_i \sim \mathcal{D}_{LLM}(P, S_1).
\end{cases}
\]

    \item \textbf{BckTr}$@k$ estimator that at least one of k LLM attempts satisfies backtranslation of the dataclass methods
\[
\text{BckTr} = \mathbb{E}_{\substack{P \sim \mathcal{P} \\ S_1 \in \mathcal{S}(P)}}\left[ 1 - \frac{\binom{n - c}{k}}{\binom{n}{k}} \right],
\quad \text{where }
\begin{cases}
c = \sum_{i=1}^n \mathbb{I}(\text{BckTr is satisfied for } D_i), \\
D_i \sim \mathcal{D}_{LLM}(P, S_1).
\end{cases}
\]
\end{enumerate}

\subsection{Complexity Metrics}
\label{complexity_metrics}

\subsubsection{Complexity Prediction Metrics}

The first evaluation task, \textbf{Complexity Prediction}, involves predicting the time and space complexity of a problem given its description and a human solution. The metrics for this task are defined as follows:

\begin{itemize}
    \item \textbf{Pass@k}: Measures the accuracy of correctly predicting the complexity class for each problem, macro-averaged across complexity classes and then across problems. The unbiased estimator among $k$ samples is used.\\

    \item \textbf{Best@k}: Measures accuracy only for the most optimized complexity class of each problem, using the unbiased estimator among $k$ samples.\\

    \item \textbf{All@k}: Requires correct complexity output across all complexity classes simultaneously for each problem, using the unbiased estimator among $k$ samples.\\
\end{itemize}

We share below more formal definitions of these metrics. First, we define the following setup for the task, on top of the general set up of coding problems involving problems \(P \sim \mathcal{P}\) and corresponding human solutions \( S \sim \mathcal{S}(P)\) as introduced in \cref{dataclass_metrics}:
\begin{itemize}
    \item \( \mathcal{C} \): Set of complexity classes (e.g., time, space).\\
    \item For each problem \( P \), \( \mathcal{C}(P) \subseteq \mathcal{C} \): Subset of complexity classes relevant to \( P \). This set is determined in our case using the complexity framework on the ground truth human solutions, and post-processing the distribution of complexity classes (typically to remove any obvious outlier).\\
    \item \( \mathcal{H}(P, c) \): Set of human solutions for problem \( P \) and complexity class \( c \in \mathcal{C}(P) \).\\
    \item \( S' \): LLM-generated solution containing the predicted complexity for the given human code \( S \) and problem \( P \). This solution is generated by the LLM \(S' \sim \mathcal{D}_{LLM}(P, S) \).\\
    \item \( G \) is the function that extracts the ground-truth label assigned to the human code, the very same complexity that we are trying to predict using the LLM.\\
    \item \( \text{parse}(S') \): Parsing function that extracts the predicted complexity of class \( c' \) from \( S' \).\\
    \item \( \mathbb{I}_{\text{correct}}(S', S, P) \): Indicator function equal to 1 if \( \text{parse}(S') = G(S) \), else 0.\\
\end{itemize}

That said, we define the above introduced metrics as the following:

\begin{enumerate}
\item \textbf{Pass@k}: Measures correctness for each complexity class of each problem independently, averaged over all classes and problems. Correctness of having at least one correct solution out of $k$ attempts, using a sampling budget of $n > k$ (usually using $n = 2\times k$):\\
   \[
   \text{Pass@k} = \mathbb{E}_{P \sim \mathcal{P}}\left[ \frac{1}{|\mathcal{C}(P)|} \sum_{c \in \mathcal{C}(P)} \left[ 1 - \frac{\dbinom{n - c_c}{k}}{\dbinom{n}{k}} \right] \right],
   \]
   where \(c_c\) is the count of times the LLM found the correct solution, that is to say labeled the complexity class of the human solution correctly:
   \[
   c_c = \sum_{i=1}^n \mathbb{I}_{\text{correct}}(S'_i, S_1, P),
   \]
   and \( S_1 \) is a human solution from the same problem and complexity class \(S_1 \in \mathcal{H}(P, c)\).\\

\item \textbf{Best@k}: Evaluates correctness only for the best (e.g., most efficient) complexity class per problem, in particular the probability of getting one correct answer among k answers:\\
   \[
   \text{Best@k} = \mathbb{E}_{P \sim \mathcal{P}}\left[ 1 - \frac{\dbinom{n - c_{c_{\text{best}}}}{k}}{\dbinom{n}{k}} \right],
   \]
   where \(c_{c_{\text{best}}}\) is the count of times the LLM found the correct solution, that is to say labeled the complexity class of the human solution correctly, only for the best class of complexity \(c_{\text{best}}\) for each problem:
   \[
   c_{c_{\text{best}}} = \sum_{i=1}^n \mathbb{I}_{\text{correct}}(S'_i, S_1, P),
   \]
   where \( c_{\text{best}}(P) \) is the most optimized complexity class for \( P \), and \( S_1 \in \mathcal{H}(P, c_{\text{best}}(P))\).\\

\item \textbf{All@k}: Joint accuracy across all complexity classes. It requires correctness for all complexity classes simultaneously:\\
   \[
   \text{All@k} = \mathbb{E}_{P \sim \mathcal{P}}\left[ 1 - \frac{\dbinom{n - c_{\text{all}}}{k}}{\dbinom{n}{k}} \right],
   \]
   where
   \[
   c_{\text{all}} = \sum_{i=1}^n \mathbb{I}\left( \bigcap_{c \in \mathcal{C}(P)} \mathbb{I}_{\text{correct}}(S'_i, S_c, P) \right)
   \]
   with \(S_c \in \mathcal{H}(P, c)\).\\
\end{enumerate}

These metrics all use the unbiased estimator \( 1 - \frac{\dbinom{n - c_.}{k}}{\dbinom{n}{k}} \) from \citet{chen2021evaluating}, where \( n \) is the total number of samples, and \( c_. \) is the count of valid samples, usually parametrized by a complexity class for instance.

\subsubsection{Complexity Generation Metrics}

The second task, Complexity Generation, involves generating solutions that meet specific complexity requirements. The metrics are defined as follows:

\begin{itemize}
    \item \textbf{Pass@k}: Measures performance of generating a correct solution that meets each complexity class requirement independently, averaged over all classes and problems.\\

    \item \textbf{Best@k}: Measures generation correctness for the most optimized complexity class per problem.  \\

    \item \textbf{All@k}: Requires the solution to meet all complexity classes simultaneously for each problem.\\
\end{itemize}

For each problem \( P \sim \mathcal{P} \):
\begin{itemize}
    \item \( \mathcal{C}(P) \): Set of complexity classes (e.g., time and/or space) that exist for a particular problem.\\
    \item \( S' \): Generated solution by the LLM. We can write \( S'(P, c)\) or \(S' \sim \mathcal{D}_{LLM}(P, c)\) to specify that the generation of the LLM is conditioned on a particular problem \( P \) and a requested complexity class \(c\) for the solution.\\
    \item \( \mathbb{I}_{\text{correct}}(S', P) \): 1 if \( S' \) passes all tests associated with problem \(P\) which means it is evaluated as a correct solution to the coding challenge, 0 otherwise (therefore this does not take into account the complexity of the generated solution).\\
    \item \( \mathbb{I}_{\text{class}}(S', c, P) \): 1 if \( S' \) meets complexity class \( c \in \mathcal{C}(P) \), 0 otherwise. This does not depend on the correctness of the solution with respect to the coding challenge \( P \) alone. So a solution that compiles, runs and produces wrong results can potentially meet the complexity requirement.\\

\end{itemize}

In this context, we can define the following metrics associated with the task of complexity generation:\\

\begin{itemize}
    \item \textbf{Pass@k}:
    \[
    \text{Pass@k} = \mathbb{E}_{P \sim \mathcal{P}}\left[ \frac{1}{|\mathcal{C}(P)|} \sum_{c \in \mathcal{C}(P)} \left( 1 - \frac{\dbinom{n - c_c}{k}}{\dbinom{n}{k}} \right) \right],
    \]
    where \( c_c = \sum_{i=1}^n \mathbb{I}_{\text{class}}(S'_i, c, P) \times \mathbb{I}_{\text{correct}}(S'_i, P) \).\\

    \item \textbf{Best@k}:
    \[
    \text{Best@k} = \mathbb{E}_{P \sim \mathcal{P}}\left[ 1 - \frac{\dbinom{n - c_{\text{best}}}{k}}{\dbinom{n}{k}} \right],
    \]
    where \( c_{\text{best}} = \sum_{i=1}^n \mathbb{I}_{\text{class}}(S'_i, c_{\text{best}}(P), P) \times \mathbb{I}_{\text{correct}}(S'_i, P) \), and \( c_{\text{best}}(P) \) is the most optimized class in \( \mathcal{C}(P) \).\\

    \item \textbf{All@k}:
    \[
    \text{All@k} = \mathbb{E}_{P \sim \mathcal{P}}\left[ 1 - \frac{\dbinom{n - c_{\text{all}}}{k}}{\dbinom{n}{k}} \right],
    \]
    where \( c_{\text{all}} = \sum_{i=1}^n \prod_{c \in \mathcal{C}(P)} \mathbb{I}_{\text{correct}}(S'_{i, c}, P) \times  \mathbb{I}_{\text{class}}(S'_{i, c}, c, P) \) with \(S'_{i, c} \sim \mathcal{D}_{LLM}(P, c)\).\\
\end{itemize}

\subsubsection{Complexity Ranking Metrics}

Concerning the Complexity Ranking Task, as explained in the main manuscript, the prompt is modified so as to query the model for the complexity generation task, with an optimization addition: the model needs to generate correct code, with correct complexity, as optimized as possible while adhering to this complexity requirement.

To formalize the metrics of this task, we add the following elements to the formalism defined in the previous subsections:

\begin{itemize}
    \item \( \mathcal{H}(P, c) \): Set of human solutions for problem \( P \) and complexity class \( c \in \mathcal{C}(P) \).\\

    \item \( \text{coeff}(S, P, c) \): Complexity coefficient of solution \( S \) for problem \( P \) and class \( c \) (lower is better).\\

    \item \( \text{best\_coeff}_{LLM}(P, c) \): Best coefficient from \( n = 20 \) LLM-generated solutions for \( P \) and \( c \). This is obtained by sampling \(S'_1, S'_2, \dots S'_n \sim \mathcal{D}_{LLM}(P, c)\) and ranking \(\text{coeff}(S'_1, P, c), \text{coeff}(S'_2, P, c), \dots \text{coeff}(S'_n, P, c)\).\\

    \item \( \mathbb{I}_{\text{correct}}(S', P) \) has the exact same definition as for the complexity generation task. It measures whether the generated solution is correct for a given problem given its I/O test cases.\\

    \item \( \mathbb{I}_{\text{class}}(S', c, P) \) has the exact same definition as for the complexity generation task. It measures whether a generated solution respects the complexity class it was conditioned upon. \\

\end{itemize}

We now introduce the following metrics. \textbf{Pass@k}, \textbf{Best@k} and \textbf{All@k} have the same definitions as in the complexity generation task:

\begin{itemize}
    \item \textbf{Pass@k}:
    \[
    \text{Pass@k} = \mathbb{E}_{P \sim \mathcal{P}}\left[ \frac{1}{|\mathcal{C}(P)|} \sum_{c \in \mathcal{C}(P)} \left( 1 - \frac{\dbinom{n - c_c}{k}}{\dbinom{n}{k}} \right) \right],
    \]
    where \( c_c = \sum_{i=1}^n \mathbb{I}_{\text{class}}(S'_i, c, P) \times \mathbb{I}_{\text{correct}}(S'_i, P) \).\\

    \item \textbf{Best@k}:
    \[
    \text{Best@k} = \mathbb{E}_{P \sim \mathcal{P}}\left[ 1 - \frac{\dbinom{n - c_{\text{best}}}{k}}{\dbinom{n}{k}} \right],
    \]
    where \( c_{\text{best}} = \sum_{i=1}^n \mathbb{I}_{\text{class}}(S'_i, c_{\text{best}}(P), P) \times \mathbb{I}_{\text{correct}}(S'_i, P) \), and \( c_{\text{best}}(P) \) is the most optimized class in \( \mathcal{C}(P) \).\\

    \item \textbf{All@k}:
    \[
    \text{All@k} = \mathbb{E}_{P \sim \mathcal{P}}\left[ 1 - \frac{\dbinom{n - c_{\text{all}}}{k}}{\dbinom{n}{k}} \right],
    \]
    where \( c_{\text{all}} = \sum_{i=1}^n \prod_{c \in \mathcal{C}(P)} \mathbb{I}_{\text{correct}}(S'_{i, c}, P) \times  \mathbb{I}_{\text{class}}(S'_{i, c}, c, P) \) with \(S'_{i, c} \sim \mathcal{D}_{LLM}(P, c)\).\\

    \item \textbf{COEFF}: Percentile ranking against human solutions of the same problem and complexity class. It compares the LLM's best coefficient to human solutions in the same class. This metric uses a fixed sampling budget of size $n$ (in our experiments we take $n=20$):
   \[
   \text{COEFF} = \mathbb{E}_{P \sim \mathcal{P},\, c \in \mathcal{C}(P)}\left[ \text{rank}_{LLM}(P, c) \right],
   \]
   where:
   \[
   \text{rank}_{LLM}(P, c) =
   \begin{cases}
   \frac{|\{ h \in \mathcal{H}(P, c) \mid \text{coeff}(h, P, c) > \text{best\_coeff}_{LLM}(P, c) \}|}{|\mathcal{H}(P, c)|} \times 100 & \text{if } \exists S' \text{ valid}, \\
   0 & \text{otherwise}.
   \end{cases}
   \]\\

   In addition, we also introduce \(\text{COEFF}_{intersect}\) (and distinguish it from \(\text{COEFF}\) by denoting the latter one as \(\text{COEFF}_{full}\)) as the expectation used in the definition of \(\text{COEFF}\) restricted to a subset of problems and complexity classes. For the set \(\mathcal{M^*}\) of starred models of the complexity ranking table, we define the set of problems \(\mathcal{P'}\) as the collection of problems \( P \) from \( \mathcal{P} \) where there exists at least one complexity class \( c \) associated with \( P \), such that for every model \( M \) in the set \( \mathcal{M}^* \), sampling \( n \) solutions from \( M \) on \( P \) and \( c \) yields at least one valid solution. Formally:\\

\[
\begin{split}
\mathcal{P}' = \Bigl\{ P \in \mathcal{P} \mid \exists c \in \mathcal{C}(P), \forall M \in \mathcal{M}^*, \sum_{i=1}^n & \mathbb{I}_{\text{correct}}(S'_i, P) \times \mathbb{I}_{\text{class}}(S'_i, c, P) \geq 1 \\
& \text{where } S'_i \sim \mathcal{D}_M(P, c) \Bigr\}
\end{split}
\]\\

    Similarly, we define the set of corresponding complexity classes \( \mathcal{C}'(P) \) where all models are indeed producing at least one correct sample code. For any \( P \in \mathcal{P}' \), we define \( \mathcal{C}'(P) \):\\

\[
\begin{split}
\mathcal{C}'(P) = \Bigl\{ c \in \mathcal{C}(P) \mid \forall M \in \mathcal{M}^*, \sum_{i=1}^n & \mathbb{I}_{\text{correct}}(S'_i, P) \times \mathbb{I}_{\text{class}}(S'_i, c, P) \geq 1 \\
& \text{where } S'_i \sim \mathcal{D}_M(P, c) \Bigr\}
\end{split}
\]\\

    Based on that, \(\text{COEFF}_{intersec}\) can be formalized as:\\

\[
\text{COEFF}_{intersect} = \mathbb{E}_{P \sim \mathcal{P'},\, c \in \mathcal{C'}(P)}\left[ \text{rank}_{LLM}(P, c) \right]
\] where we are using the above definition of \(\text{rank}_{LLM}(P, c)\).\\

\end{itemize}

Note that the COEFF metric uses the best coefficient from \(n\) (20 in practice) LLM attempts and ranks it against human solutions in the same complexity class. A higher percentile means the LLM's solution is more optimized than most human solutions. The \textbf{Intersec} subset ensures comparisons are only made when all models have at least one valid solution, avoiding skewed rankings from partial failures. We are selecting models that are good enough in the first place so as to make the subset viable and interesting (it has to contain enough samples in the first place).
Finally, the unbiased estimator \( 1 - \frac{\dbinom{n - c}{k}}{\dbinom{n}{k}} \) remains consistent with prior work \citep{chen2021evaluating}.

\newpage

\section{Time and Space Complexity: Definitions and Conventions}
\label{time_space_complexity_definitions}
\titleformat*{\paragraph}{\normalfont\sffamily\bfseries}

The complexity labels released with \benchmark aim at the theoretical worst-case time and space complexity of each solution. Deriving that quantity by hand for 1,190,250 solutions was not an option, and the level of expertise it demands from annotators was out of reach (\cref{quantitative_analysis}), so we estimate it instead: the framework runs a solution over a range of input sizes and fits the observed runtime and memory curves against a fixed set of candidate growth functions, with no static analysis of the source involved at any point. The theoretical complexity remains the target throughout, and \cref{complexity_framework_performance} measures how close the estimate comes to it, reporting 84\% agreement with human annotation on time complexity and 82\% on space. What the framework returns is therefore an estimate drawn from finitely many executions rather than a proof about the program text, which is the sense in which \cref{introduction} describes these labels as performance profiles and not theoretical complexities.

We first discuss the mathematical definitions of asymptotic notation and the notion of running case needed to carry them over to programs (\cref{app:running_cases}), then the convention commonly assumed in programming practice and the sense in which we adopt it (\cref{app:convention}), and its extension to complexities in several variables (\cref{app:multivariable}). We then turn to the labels themselves: what \benchmark discards from the competitive programming setting its data comes from (\cref{app:contest_constraints}), the finite space of classes a label is drawn from (\cref{app:label_space}), and what a measured label does and does not pin down (\cref{app:label_object}). We finally share a few examples to illustrate the notions of complexity (\cref{app:complexity_examples}).

\subsection{Big-O in Mathematics and Algorithmics}
\label{app:running_cases}

Big-O, big-Theta and big-Omega are defined in mathematical analysis \citep{cormen2022introduction,sipser2013introduction} for real- or complex-valued functions on a domain that is typically the real or complex numbers, or tuples of real or complex numbers; they denote respectively an asymptotic upper bound, a tight bound and an asymptotic lower bound, each up to a constant factor. Programs can be defined on trickier domains: their inputs are integers, lists, strings, dictionaries and mixtures of these, not necessarily carrying a natural ordering or sense of growing to infinity. One can of course always fall back on the total number of bits of all the inputs, which does give an ordering and a sense of growing to infinity, but that map is not injective: many different inputs share the same bit count while costing wildly different amounts, so the cost is not a function of that size alone. Consider a function taking an integer and a dictionary, or a function taking an integer and a list, where it is not clear whether the list gets ``bigger'' when its elements grow in size or when the length of the list itself grows, not to mention how the content of the list may intertwine with the integer.

\paragraph{Running cases.}
The gap between the mathematical definitions and the programming case can be closed with what we call a \textbf{running case}: a generator $g$ that maps a single integer parameter $n$ to an input of the program, obtained by scaling designated arguments of a given example input. The generator need not be defined at every $n$, since structural constraints on a problem's inputs may leave some sizes unrealisable; it is enough that it be defined on an infinite set $S_g \subseteq \mathbb{N}$, along which the asymptotic statements below are read. The parameter is the size: a running case fixes what size means for the program at hand (we will see in \cref{app:multivariable} that in the multivariate case we can use tuples of size parameters). Write $f$ for the program and $T$ for the cost of the program we are interested in, either its runtime or its memory footprint, as typically recorded by a sandbox equipped with a clock and a memory profiler. Then $T(f \circ g)$ is that cost as a function of $n$, for $f$ run on the input $g(n)$. A running case makes it so we can apply the mathematical definitions: $T(f \circ g)$ is a real-valued function of the single integer variable $n$ ranging over $S_g$, to which $O$, $\Theta$ and $\Omega$ apply verbatim. The generated input is also required to grow without bound as $n$ increases: a generator allowed to return the same input at every size would trace a flat curve for any program, and would give every program a best case in $O(1)$.

Writing $\mathcal{G}_n$ for the running cases of $f$ that are defined at $n$, the familiar cases are three aggregations over that set:
\begin{equation}
T_{\text{best}}(n) = \inf_{g \in \mathcal{G}_n} T(f \circ g)(n),
\qquad
T_{\text{avg}}(n) = \mathbb{E}_{g \sim \mu}\bigl[T(f \circ g)(n)\bigr],
\qquad
T_{\text{worst}}(n) = \sup_{g \in \mathcal{G}_n} T(f \circ g)(n).
\label{eq:running_case_aggregations}
\end{equation}
Therefore $T_{\text{worst}}$ is the worst cost among the possible inputs of a given size $n$ when running the program on them. The expectation is different in kind, as it requires a distribution $\mu$ over running cases. We do not use it. No principled $\mu$ is available here, since weighting the running cases amounts to assuming a likelihood of seeing this or that shape of inputs when deploying the piece of code in real environments, and the answer depends entirely on that assumption: a distribution concentrated on random permutations gives Quick Sort $\Theta(n \cdot \mathrm{log}(n))$, while one concentrated on nearly sorted inputs gives $\Theta(n^2)$. In general, because it takes many assumptions to reason about the average case, one estimates the best and the worst case instead, and the average is then either one of them or somewhere in between. \benchmark uses the worst running case throughout.

Complexity analysis then amounts to characterizing how $T_{\text{worst}}$ grows with $n$, applying the bounds to that cost curve rather than to the program itself. Of course, in the above definitions the growth rates of the different cases are loosely defined, depending on the exact parametrization of the generator function: the input can grow as fast as $n$, twice as fast, or along any ``up to a constant factor'' schedule. This is why big-O, big-Theta and big-Omega are the perfect tools for complexity analysis, characterizing growth rates ``up to a constant factor''. Concretely, for a running case $g$ and a growth function $h$:
\begin{itemize}
    \item $T(f \circ g)(n) = O(h(n))$ when there exist $c > 0$ and $n_0$ such that $T(f \circ g)(n) \leq c \, h(n)$ for every $n \geq n_0$, so that $h$ bounds the growth of the cost along $g$ from above;
    \item $T(f \circ g)(n) = \Omega(h(n))$ when there exist $c > 0$ and $n_0$ such that $T(f \circ g)(n) \geq c \, h(n)$ for every $n \geq n_0$, so that $h$ bounds it from below;
    \item $T(f \circ g)(n) = \Theta(h(n))$ when both hold, so that $h$ gives the growth of the cost exactly, up to a constant factor.
\end{itemize}
The same three definitions apply verbatim to $T_{\text{best}}$, $T_{\text{avg}}$ and $T_{\text{worst}}$, which are themselves functions of $n$. Finding an $h$ with $T_{\text{worst}}(n) = O(h(n))$ is useful because it bounds every input at once: no input of size $n$ costs more than on the order of $h(n)$. It does not say that the bound is reached, since the program may do better and a slower-growing $h$ may bound it as well, so a big-O statement about the worst case leaves room above the actual behavior. Only $T_{\text{worst}}(n) = \Theta(h(n))$ fixes the growth rate.

\paragraph{Cases and bounds are independent.}
Selecting a running case determines which function of $n$ is under discussion; choosing between $O$, $\Theta$ and $\Omega$ determines how that function is bounded. The two choices are orthogonal, and every combination of them is meaningful, as is standard in the algorithms literature \citep{cormen2022introduction} and as shown in \cref{tab:cases_and_bounds}. The lower bound on the worst case in the bottom-right cell of that table is the decision-tree bound on comparison sorting, one of the standard results of the field. Three running cases and three bounds give nine statements available about a program, not three.

\begin{table}[h]
\centering\small
\caption{Running case and asymptotic bound are independent choices. The running case selects which function of $n$ is being bounded, while the bound describes how that function is bounded; each cell is a statement that is standard in the algorithms literature. Insertion sort attains its best running case on already sorted inputs and its worst running case on reversed inputs. The bottom-right cell is the decision-tree lower bound on comparison sorting, which is not expressible if big-Omega is read as denoting the best case.}
\label{tab:cases_and_bounds}
\begin{tabular}{llll}
\toprule
 & \textbf{Upper bound} ($O$) & \textbf{Tight bound} ($\Theta$) & \textbf{Lower bound} ($\Omega$) \\
\midrule
$T_{\text{best}}$ & insertion sort is $O(n)$ & Quick Sort is $\Theta(n \cdot \mathrm{log}(n))$ & any comparison sort is $\Omega(n)$\footnotemark{} \\
$T_{\text{worst}}$ & insertion sort is $O(n^2)$ & insertion sort is $\Theta(n^2)$ & any comparison sort is $\Omega(n \cdot \mathrm{log}(n))$\footnotemark{} \\
\bottomrule
\end{tabular}
\end{table}
\addtocounter{footnote}{-2}
\stepcounter{footnote}\footnotetext{A correct comparison sort must perform at least $n-1$ comparisons on every input, including the easiest one, in order to certify that the output is ordered.}
\stepcounter{footnote}\footnotetext{This is the decision-tree lower bound on comparison sorting, Theorem 8.1 of \citet{cormen2022introduction}.}

\subsection{Commonly Assumed Big-O Definitions in Programming}
\label{app:convention}

Programming practice keeps only a narrow, and kind of incorrect, version of these statements, by letting the letter determine the running case: big-O is identified with the worst case and big-Omega with the best case. In this usage, big-Theta reads as ``the tight one'', which is what it means, though without saying which running case it is tight on. \citet{Knuth1976} introduced big-Theta into computer science precisely to express tightness, because big-O was already being used as though it were tight, and the standard texts keep the case and the bound as separate axes.

Nevertheless, even if the popular programming practice uses big-O and the other notations in an improper manner, and it goes against the definitions we introduced in \cref{app:running_cases}, we try to adopt these conventions, for two reasons. The first is that it is the vocabulary in which the question is actually asked. Complexity is discussed among software engineers in the register of a coding interview for example, where a candidate is asked for ``the complexity of this function'' and is expected to answer with a single big-O expression, and \benchmark measures how LLMs perform on the question as it is posed in practice. Restating the query in the terms of \cref{app:running_cases} would measure something else, namely how well a model follows the formal mathematical definitions, rather than what developers commonly expect.

The second reason is also that \benchmark has to attach (within each best/worst/average case) a single label to each program, and that label has to be tight: without a tightness requirement, any program could be labeled $O(n!)$ and no prediction would ever be wrong. The prompts state this directly, asking for the tightest possible upper bound in big-O notation.

These conventions can nevertheless be defined according to the precise terms of \cref{app:running_cases}:
\begin{align*}
O_{\text{conv}}(f) &= h \quad \text{such that } T_{\text{worst}}(n) = \Theta(h(n)), \\
\Omega_{\text{conv}}(f) &= h \quad \text{such that } T_{\text{best}}(n) = \Theta(h(n)), \\
\Theta_{\text{conv}}(f) &= h \quad \text{defined only when } O_{\text{conv}}(f) = \Omega_{\text{conv}}(f) = h.
\end{align*}
Written this way the convention is self-consistent, and the sense in which it departs from the analytic definitions becomes explicit: it uses the symbol $O$ to state a tight, $\Theta$-style fact about one particular running case. For big-Theta, since $T_{\text{best}}(n) \leq T(f \circ g)(n) \leq T_{\text{worst}}(n)$ holds for every $g$, requiring $O_{\text{conv}}$ and $\Omega_{\text{conv}}$ to agree pins every running case to the same growth rate, so $\Theta_{\text{conv}}$ exists exactly when the program is insensitive to which input of a given size it receives. Quick Sort, with a best running case in $\Omega(n \cdot \mathrm{log}(n))$ and a worst running case in $O(n^2)$, has no $\Theta_{\text{conv}}$, and the query ``predict the big-Theta complexity of this program'' becomes ill-defined.

\Cref{ablations_prompts} measures what changes of formulations imply, by rewriting the prompt in terms of big-Theta: the reasoning model moves from 40.3 to 39.9 $All@1$ on time complexity prediction, scored against the same worst-case labels. Models appear to read big-Theta as a request for a tight bound and to apply it to the same running case they would have used for big-O: being trained on data produced by developers who assume the conventions above, they probably adhere to them anyway.

\subsection{Multi-Variable Complexity}
\label{app:multivariable}

Several \benchmark classes take more than one variable, such as $O(n + m)$, $O(n \times m)$ and $O((n+m) \cdot \mathrm{log}(n+m))$, and a running case as defined above would not produce them, since $T(f \circ g)$ is a function of $n$ alone. The input generators make it possible to grow several variables together at different rates, but in practice people find it useful, where it is defined, to recover the dependency on each variable separately. We therefore also index running cases by which arguments they grow. For a program taking arguments $a_1, \dots, a_k$, a running case grows a chosen subset of them jointly at a common rate while holding the others fixed. Growing subsets one at a time gives the dependence of runtime and memory on each argument separately, and growing them together exposes interactions that the separate measurements miss, such as a loop nested over two arguments. The global label is composed from the per-subset fits, which is how a product term such as $n \times m$ is told apart from a sum such as $n + m$.

Asymptotic notation over several variables carries subtleties that the single-variable definitions do not, since the domain admits several inequivalent notions of tending to infinity \citep{Howell2008}. We do not dig too much into the subtleties of the multivariate cases in this paper, and stick to the above approach together with the finite label space of \cref{app:label_space}.

\subsection{Asymptotic Complexity Versus Code Competition Constraints}
\label{app:contest_constraints}

A code competition problem, for instance from \vendor, comes with input bounds, along with a time limit. That limit exists for a practical reason: the judge has to evaluate every submission under a finite compute budget, which is only possible if each execution terminates within a reasonable bounded time. The input bounds are then chosen together with the timeout, so that a solution of the complexity the problem is designed around runs within it while a less efficient one exceeds it. A problem stating $n \leq 3 \times 10^5$ under a one-second limit is asking for an $O(n \cdot \mathrm{log}(n))$ solution: $O(n^2)$ would amount to roughly $10^{11}$ operations and time out, whereas $O(n \cdot \mathrm{log}(n))$ comfortably fits. Asymptotic complexity is being enforced in a limited compute setting using input bounds and timeouts.

\paragraph{Dropping the timeout.}
\benchmark gets rid of these limits, starting with the timeout. Fitting a complexity class to empirical runtime curves is already delicate, and asking the framework to also reproduce the pass or fail behavior of a problem-specific timeout would add a second and largely independent source of error. Timeouts are noisy quantities, and in code competition datasets they can even be improperly calibrated for a given problem in a given language \citep{chambon2026reinforcementlearningcodeoptimization}. The framework therefore measures runtime and memory without reference to any time limit. It does bound its own executions, for the same practical reason any judge does, but those limits guard against runaway runs rather than serving as a criterion the solution is judged against.

\paragraph{Dropping the input bounds.}
Discarding the timeout entails discarding the input bounds, since the bounds only carry information about complexity in conjunction with it: they were chosen to make a target complexity fit within a given limit, and taken alone they say nothing about how a program scales. In fact, they can only alter the complexity label being computed, and growing beyond them helps better determine the asymptotic behavior. The framework consequently grows inputs well past the stated bounds, up to $3000\times$ the original input size (\cref{benchmark_experimental_setup}). The cost is that the framework deliberately runs code outside the regime its author had in mind, which is why the prompts are updated to specify that inputs are assumed to potentially grow indefinitely. Code that silently relies on the stated bounds can fail once they are exceeded, which is in fact what our benchmark favors over solutions that enumerate cases to save a bit of runtime on a bounded subset of inputs, instead of achieving real asymptotic complexity improvements.

\paragraph{Evaluating beyond the intended complexity.}
A further reason to detach a problem from its published constraints is that \benchmark does not restrict itself to the complexity that problem was designed around. Each retained problem can carry several complexity classes among its solutions (\cref{composition}), and the prediction, generation and ranking tasks address the less efficient ones as much as the most optimized one, precisely in order to observe how models navigate between them. The generation task in particular can ask a model for a solution in a class the original competition would have rejected: a correct $O(n^2)$ program for a problem whose intended solution is $O(n \cdot \mathrm{log}(n))$ is a valid answer here and a timeout for the judge. Bounds and a time limit calibrated for one target complexity cannot serve a setting that evaluates several.

\paragraph{Adapting the query.}
The bounds affect the models as well. We noticed in preliminary experiments that, because the bounds appear in the problem description handed to the model, some LLMs treat the input as bounded and conclude that the program runs in constant time. When asked to generate, they can also `cheat' in the same way, writing code that only has to handle the finite set of inputs the bounds allow, enumerating or precomputing cases rather than implementing an algorithm of the requested complexity. Since the bounds exist for the purposes of the competition, and only in conjunction with its time limit, we state explicitly in the prompt that input sizes are to be assumed unbounded, which matches what the framework does for the reasons above. This keeps the stated constraints from steering the reasoning, and leaves the model to reason about the asymptotic complexity itself.

\paragraph{Consequences for the benchmark.}
A label and a contest verdict can then disagree. A solution accepted by the original judge because the bounds were loose enough for it to fit within the time limit is still labeled by its asymptotic behavior, so an accepted $O(n^2)$ solution is labeled $O(n^2)$, and would be rejected in our evaluation for not matching the best complexity. That disagreement is what makes the benchmark discriminative: several complexity classes can coexist among accepted solutions to the same problem when the bounds leave enough slack for a suboptimal algorithm to pass, but \benchmark can still differentiate them by looking at their complexity directly.

\subsection{The Space of Possible Complexity Labels in \benchmark}
\label{app:label_space}

The framework fits a fixed library of base growth functions, combined with addition and multiplication, and returns the member of the resulting finite set $\mathcal{C}$ of complexity classes that minimizes the fitting residual under a simplicity bias (\cref{implementation_details}). Multi-variable classes such as $O(n + m)$ and $O(n \times m)$ arise from the same construction applied across several arguments (\cref{app:multivariable}).

A \benchmark label is a minimizer over $\mathcal{C}$, not the result of a search over all growth functions, so a program whose true growth lies outside $\mathcal{C}$ receives the nearest representable class, which gets considered as a fitting error. \Cref{implementation_details} lists growth functions that the current library does not represent. The time complexity test set exercises 11 classes and the space complexity test set 5 (\cref{composition}), though the training set is much larger and therefore covers many more classes. Model predictions are compared against these labels after normalization, so that different ways of writing a complexity, by changing letters, order or notation, do not affect the match; \cref{complexity_metrics} defines the metrics computed on top of that comparison.

The classes reachable for a given problem also depend on its dataclass (\cref{dataclass_generation}). The dataclass determines how the input string is decomposed into separate arguments, and the framework can only attribute growth to arguments that the dataclass exposes. A dataclass parsing a graph problem into a vertex count and an edge list allows a label in $O(n + m)$, whereas a correct but coarser dataclass returning a single argument restricts the same problem to a single-variable label. Dataclass generation therefore bounds not only which problems enter the benchmark, but also which labels those problems can receive.

\subsection{What a \benchmark Label Is}
\label{app:label_object}

\paragraph{From the definition to the measurement.}
\Cref{app:running_cases} defines the worst running case as the upper bound over all admissible generators, whereas the framework computes the maximum over the running cases it implements, a finite subset $\mathcal{G}_{\text{impl}} \subset \mathcal{G}$. The measured label is therefore a lower bound on the true worst case, and the two coincide only when $\mathcal{G}_{\text{impl}}$ is rich enough to reach the worst behavior of the program. \Cref{app:complexity_examples} gives the standard example of a program where this can happen.

\paragraph{Input size.}
Complexity is expressed in terms of the size of each argument, and size is defined per argument type: for a list, a string or another sequence, its length; for a dictionary or a set, its number of entries; for an integer, its value. These are defaults rather than a fixed rule. The ambiguity raised in \cref{app:running_cases}, where a list can be said to grow either by holding more elements or by holding larger ones, is handled through the dataclass (\cref{dataclass_generation}), which can either expose one growth axis or several: a list of strings can be decomposed into the length of the list and the maximum length of the strings it contains, so that growth along each is measured separately. Composite arguments are where the notion is least clear, and they are also where coverage is weakest: \cref{complexity_framework_performance} reports nested structures such as triple-nested lists among the infrequent input types the framework fails to handle.

\paragraph{Cost model.}
Because the label comes from running the code, \benchmark assumes no cost model for individual Python operations: whatever CPython does at runtime is what the curve fit sees. Amortized costs, constant factors coming from C-level implementations, and interpreter-level optimizations are all reflected in the label. This can separate the measured label from the one a static reading produces, in both directions -- a solution that reads as linear may pay a hidden cost inside a built-in call, and a solution that reads as quadratic may benefit from an optimization in the interpreter. \Cref{app:complexity_examples} gives a concrete case. The choice was made so to also reflect interesting language-specific optimizations, which are definitely playing a big role on the problems we chose to work on (from top code competitions).

\paragraph{Space semantics.}
Space complexity in \benchmark refers to the peak memory allocated during execution, as recorded by \textit{tracemalloc}, not counting the initial space used by the inputs. The prompts state the convention the models are asked to follow: reading the input consumes no memory, any operation performed on top of the parsed input is counted, and printed output counts towards memory. This is the notion of \emph{extra} space complexity, and it is the one against which the space labels, the space prediction task and the space generation task are all defined. \textit{tracemalloc} can have limitations in accounting for all the memory a program uses, and improving the measurement tools is left for future iterations of the benchmark.

\paragraph{Constants and lower-order terms.}
A complexity class discards constant factors and lower-order terms: an algorithm performing $2n$ operations and one performing $100n$ operations are both $O(n)$, and $n^2 + n$ is $O(n^2)$, since the quadratic term dominates. \benchmark measures the discarded constant separately. The framework returns the coefficient of the fitted curve alongside the class, and the ranking task of \cref{results:third_task} compares solutions within a single class by that coefficient. The class captures how a solution scales, the coefficient captures what the asymptotic statement leaves out, and both contribute to what separates an optimized solution from a merely correct one.

\subsection{Examples}
\label{app:complexity_examples}

\paragraph{Complexity on graphs.}
Graphs are a good example to look at to illustrate complexities. They are characterized by their number of vertices $n$ and of edges $m$, and running cases can take a lot of shapes.

Take a search that stops as soon as it finds what it is looking for. If the target sits right next to the starting vertex, the search costs almost nothing; if it sits as far away as possible, the search goes through the whole graph and costs $\Theta(n + m)$. Same program, two very different running cases. A full traversal is the opposite: breadth-first and depth-first search go through every vertex and every edge whatever the graph looks like, so the best and the worst running case are the same, and the program does have a $\Theta_{\text{conv}}$ (\cref{app:convention}).

That full traversal is also where one variable stops being enough. It runs in $O(n + m)$, and we cannot drop either term. If we grow the graph keeping $m$ around $n$, which is a sparse graph, the curve looks linear and we would say $O(n)$. If we grow it keeping $m$ around $n^2$, which is a dense graph, the curve looks quadratic and we would say $O(n^2)$. Neither answer is wrong, they just describe different running cases, and only the two-variable label describes the program.

Note that $n$ and $m$ are tied to each other: a simple graph has $m \leq n(n-1)/2$ and a connected one has $m \geq n - 1$, so on connected graphs $n = O(m)$ and $O(n + m)$ becomes $O(m)$. Even having two variables depends on the graphs we feed the program. This is also where the bounds of \cref{app:contest_constraints} play a role: a problem saying $n \leq 10^5$ and $m \leq 2 \times 10^5$ is telling the competitor that the graph is sparse, so the traversal can be treated as linear. Drop the bounds and we are back to two variables.

Time and space do not always move together. With an adjacency matrix the graph takes $O(n^2)$ memory, checking whether an edge exists is immediate, and the traversal takes $O(n^2)$ whatever $m$ is, because it reads a full row per vertex. With an adjacency list the graph takes $O(n + m)$ memory and the traversal takes $O(n + m)$, but checking an edge now costs the degree of the vertex. Same algorithm from the textbook, different time and space labels, which is one reason we label both. And which of the two sizes the framework can even see depends on the dataclass (\cref{app:label_space}).

What a measurement can tell apart also depends on the graphs it runs on. Dijkstra with a binary heap is $O((n + m) \cdot \mathrm{log}(n))$, and with a Fibonacci heap it is $O(m + n \cdot \mathrm{log}(n))$. On sparse graphs both are $O(n \cdot \mathrm{log}(n))$ and nothing measured there will separate them; on dense graphs they differ by a log factor. Even when the class is the same, the coefficient of the curve is not, and that coefficient is what the ranking task of \cref{results:third_task} looks at.

Finally, some costs only make sense over a whole sequence of operations, and some differences are too small to ever show up. A single union-find operation with path compression can cost $O(\mathrm{log}(n))$, but $m$ of them cost $O(m \cdot \alpha(n))$ in total, so only the amortized cost says anything useful. And $\alpha$, the inverse Ackermann function, stays below 4 for any $n$ we could ever run, so no curve fitting will ever tell $O(m \cdot \alpha(n))$ from $O(m)$: those two only come apart at input sizes that do not exist.

\paragraph{Interpreter behavior against static reading.}
The cost model of \cref{app:label_object} follows the interpreter rather than an idealized machine, and competitive programmers rely on this heavily. Each snippet below carries a complexity that a naive reading of the source would get wrong.

\begin{lstlisting}[language=Python, frame=single, captionpos=t, title=\textbf{(1) Strings that resize in place}]
s = ""
for c in chars:
    s += c
\end{lstlisting}
This reads as quadratic, since each concatenation appears to copy the string accumulated so far, but CPython resizes the target in place when its reference count allows it, and the loop is linear in practice.

\begin{lstlisting}[language=Python, frame=single, captionpos=t, title=\textbf{(2) The same loop, one reference later}]
s = ""
for c in chars:
    t = s
    s += c
\end{lstlisting}
Keeping a second reference alive is enough to switch the optimization off, and the same loop is quadratic again. The complexity class here depends on a line that does no work at all.

\begin{lstlisting}[language=Python, frame=single, captionpos=t, title=\textbf{(3) A bitset inside an integer}]
dp = 1
for w in weights:
    dp |= dp << w
\end{lstlisting}
A whole subset-sum table carried inside one integer, where bit $i$ is set when the sum $i$ can be reached. Read statically the loop does one operation per item and looks like $O(n)$, while each shift and or actually moves $W/64$ machine words for a total weight $W$, so the loop is $O(nW/64)$. The same table written as a Python list of booleans is $O(nW)$ with an interpreter step per cell.

\begin{lstlisting}[language=Python, frame=single, captionpos=t, title=\textbf{(4) The container decides the class}]
hits = sum(x in seen for x in xs)
\end{lstlisting}
This one changes class with the type of the container alone: membership in a list is a linear scan while membership in a set is constant on average, so the same expression is quadratic or linear depending on a declaration made elsewhere in the program. The same trap exists for \texttt{x in range(n)}, which is constant because a range answers arithmetically, against \texttt{x in list(range(n))}, which scans.

\begin{lstlisting}[language=Python, frame=single, captionpos=t, title=\textbf{(5) Hashes an adversary can pick}]
seen = set()
for x in xs:
    seen.add(x)
\end{lstlisting}
Constant on average per insertion, so linear overall, but only on average. CPython hashes a small integer to itself, so integer keys can be chosen to collide in the table and drive the loop to quadratic. Competitive programmers guard against it by mixing keys with a random constant, and it is a case a generator that grows inputs at random will never produce.

\begin{lstlisting}[language=Python, frame=single, captionpos=t, title=\textbf{(6) A list used as a queue}]
while queue:
    v = queue.pop(0)
\end{lstlisting}
Here the interpreter is slower than the source suggests, as \texttt{list.pop(0)} shifts every remaining element: a breadth-first search written with a list as its queue is $O(n^2 + m)$ instead of the $O(n + m)$ its structure suggests, which is why \texttt{collections.deque} is more standard.

\begin{lstlisting}[language=Python, frame=single, captionpos=t, title=\textbf{(7) Arithmetic that is not constant}]
fact = 1
for i in range(1, n):
    fact *= i
\end{lstlisting}
Here arithmetic itself is not constant time, since Python integers are arbitrary precision, so the running product costs more at every step and the loop is superlinear. The cost of one multiplication is not fixed either: CPython switches from schoolbook multiplication to Karatsuba past a threshold, so two $d$-digit integers cost $O(d^{1.585})$ and not $O(d^2)$.

Other examples can be given. \texttt{sum(chunks, [])} looks linear and is quadratic, since every addition copies the list accumulated so far. Slicing copies too, so a recursion written on \texttt{arr[:mid]} pays for the copies on top of the recursion itself. Two further cases matter in practice without changing the class. Vectorized operations through \textsc{NumPy} keep the same asymptotic behavior while moving the work into C, which can shift the coefficient by more than an order of magnitude and is exactly what the ranking task of \cref{results:third_task} measures.
\clearpage

\section{Robustness of the Evaluation to Label Error}
\label{app:label_error}

The complexity labels of \benchmark are produced by measurement, and \cref{complexity_framework_performance} reports that they agree with human annotation on 84\% of time complexity samples and 82\% of space complexity samples. A natural concern is whether an error rate of that size can distort the comparison between models of \cref{table:complexityeval}. We address it in three steps: we measure the random component of the label error, we then corrupt the labels deliberately at rates well above it, and we report what happens to the rankings.

\subsection{Label Error in Benchmarks}
\label{app:label_error_related}

Benchmarks whose labels are imperfect are the norm rather than the exception, and the rates they report can be larger than the error rates measured for \benchmark. \citet{northcutt2021pervasive} estimate at least 3.3\% label errors on average across ten widely used test sets, including 6\% of the ImageNet validation set and 10\% of QuickDraw. \citet{beyer2020imagenet} find that roughly 29\% of re-annotated ImageNet validation images either contain several objects or fall under a category with multiple synonym labels. \citet{gema2025mmlu} re-annotate 5,700 MMLU questions with 14 experts and estimate 6.49\% erroneous questions overall, rising to 57\% on one subject. In each case the response was to correct and continue rather than to discard: \citet{beyer2020imagenet} conclude that their re-annotation reinforces ImageNet as a benchmark, and \textsc{HumanEval} remains the most cited code generation benchmark despite \textsc{HumanEval+} showing that its improved test suites cut reported $pass@k$ by up to 19.3--28.9\%, while the ordering of models barely moves and the few flips that do occur are confined to near-ties \citep{liu2023codegeneratedchatgptreally}. The same holds for \vendor, which \benchmark itself as well as a number of other code-competition test sets are built on: \citet{wang2025codecontestsplus} re-execute 1.72 million labeled submissions against its test suites and find that, among the 10,166 problems with enough submissions to score, over 4,000 have test cases that reject almost every correct solution, which they trace to incorrect test cases and to the absence of checkers for problems admitting several valid outputs. We do account for this when building the time and space complexity test sets: as described in \cref{composition}, we only keep problems with at least 50 human solutions passing the \vendor tests, so a problem whose suite rejects nearly every correct submission never reaches our test sets.

Rankings can move under label error, but only under specific circumstances that are not at stake here. \citet{northcutt2021pervasive} report that model rankings on the full ImageNet and CIFAR-10 test sets are unaffected by correcting the labels, and that inversions appear only once the test set is artificially enriched in the examples that had been mislabeled, all of them scored with their corrected labels: ResNet-18 overtakes ResNet-50 when those examples make up 6 points more of the test set, and VGG-11 overtakes VGG-19 at 5 points. But this happens because the models were trained in the first place on data carrying the same labels. The mechanism they identify is that higher-capacity models reproduce the distribution of \emph{systematic} label errors they learned from that training data. That mechanism needs the model to have learned the annotator's mistakes, which is not what we do here: the models compared in \cref{table:complexityeval} are all directly evaluated on the test set, without ever being confronted with the labels output by the complexity framework during any training stage. Therefore, none of them is given the chance to overfit to the potential systematic errors of the labeler.

Getting human-derived labels has its own share of issues, both in annotator agreement and in scaling, and using an automatic labeler is not necessarily detrimental as long as it is clearly detailed and explained. \citet{gema2025mmlu} report inter-annotator agreement between Cohen's $\kappa = 0.64$ and $0.96$ on the binary judgment of whether a question is erroneous, a considerably simpler call than assigning a complexity class to top code competition problems. Automatically derived labels are common wherever expert annotation does not scale: in \textsc{CheXpert}, labels extracted from radiology reports reach a macro-F1 between 0.34 and 0.44 against image-based expert annotation across 14 pathologies, against 0.65 to 0.93 when scored on the text they were derived from, and the dataset is a standard resource nonetheless \citep{chambon2024chexpertplus}.

The specific field of automated complexity analysis is itself a difficult task, which we believe also mitigates the severity of the error rates we observed in \benchmark. \textsc{SPEED} derives complexity bounds by static analysis of the source, without ever running the program, and only manages a precise bound for ``more than half of the loops that we have seen in practice'' \citep{gulwani2009speed}, returning nothing usable on the rest. \textsc{GuessCompx}, an empirical estimator, reports per-algorithm accuracies from 38\% to 99\% over five algorithms \citep{agenisnevers2019guesscompx}. \textsc{Trend-Prof}, the reference work on empirical complexity measurement, treats the divergence between measured and theoretical complexity as its contribution rather than as error, reporting programs that beat their worst-case theoretical bound and arguing that the measured curve gives ``a more accurate working performance model'' \citep{goldsmith2007measuring}. Part of any disagreement between our framework and a human annotator could be seen as that same divergence rather than a mistake, and \cref{time_space_complexity_definitions} further details what the \benchmark labels are and how they relate to, and differ from, the theoretical target.

\subsection{What the Error Rate Actually Is}
\label{app:label_error_rate}

We measured three quantities to get a sense of potential framework errors.

\begin{table}[!ht]
\centering\small
\caption{Three different measurements of label quality. The 84\%/82\% figure asks whether a label is \emph{correct} in absolute terms, against human annotation of human solutions. On the generation task for instance, what \benchmark scores is not the absolute label but a verdict: an attempt counts as a success only when the generated program passes the tests \emph{and} its measured complexity class matches the requested one. Verdict reproducibility measures how often the framework returns the same verdict twice on the same generated program, which bounds how often that comparison can flip and therefore how much label error can reorder models.}
\label{tab:label_error_rates}
\resizebox{\textwidth}{!}{%
\begin{tabular}{llll}
\toprule
\textbf{Quantity} & \textbf{Value} & \textbf{What it measures} & \textbf{Population} \\
\midrule
Framework accuracy & 84\% / 82\% & is the label correct & human solutions, 125 samples per axis \\
Self-consistency & 91.9\% / 89.1\% & is the label stable across reruns & human solutions \\
Verdict reproducibility & ${\sim}97\%$ & is the verdict stable across reruns & model-generated code, 12,800 attempts \\
\bottomrule
\end{tabular}%
}
\end{table}

\paragraph{Which error rate would flip benchmark rankings.}
Label error splits into a systematic part and a random part. The systematic part -- the framework consistently reading some pattern as $O(n)$ when a human would call it $O(n \cdot \mathrm{log}(n))$ -- applies to every model identically, since every model is scored against the same labels. The question as to whether it can flip some models comes down to whether it impacts models differently, in a systematic manner. The random part would impact models differently, but would it systematically shift one model compared to another over the range of test problems? The 84\%/82\% figure is the right number for asking whether the labels are true; it is not the right number for asking whether label error can flip a ranking.

\paragraph{Measuring the random part.}
Random noise mostly comes from the noise of the execution measures, both time and space. The framework was run three times over an identical set of \textsc{Gemma 3 27B} generations, so that sampling noise is held fixed and the only thing varying is the framework itself. Over all 12,800 attempts, pairs of replicates return an equal complexity class 96.7\% to 97.1\% of the time, and 7.8\% of attempts have at least one of the three replicates disagreeing. The disagreements are concentrated on adjacent classes and on permutations of the same variables rather than spread at random, as shown in \cref{tab:replicate_disagreements}. Downstream the effect is small: across the three replicates $Pass@1$ is 15.2 / 15.1 / 15.2, $Best@1$ is 17.7 / 17.8 / 17.6 and $All@1$ is 1.8 / 1.8 / 1.8, so roughly 3\% per-attempt verdict noise moves $Pass@1$ by about 0.05 points, an order of magnitude below the differences reported in \cref{table:complexityeval}.

\begin{table}[!ht]
\centering\small
\caption{The most frequent disagreements between three framework replicates over identical \textsc{Gemma 3 27B} generations, out of 12,800 attempts. Disagreement is concentrated on neighboring complexity classes and on permutations of the same variables. Such permutations are in principle handled by the framework, except when it cannot tell whether they translate into genuinely different complexities: do $n$ and $m$ interchangeably designate two lists, or the number of edges and of vertices of a graph?}
\label{tab:replicate_disagreements}
\begin{tabular}{ll}
\toprule
\textbf{Count} & \textbf{Disagreeing pair} \\
\midrule
242 & $O(n+m\,\mathrm{log}(m))$ \; vs \; $O(n\,\mathrm{log}(n)+m)$ \\
114 & $O(n)$ \; vs \; $O(n \cdot \mathrm{log}(n))$ \\
69 & $O(n)$ \; vs \; $O(n^2)$ \\
66 & $O(n \cdot m\,\mathrm{log}(m))$ \; vs \; $O(n\,\mathrm{log}(n) \cdot m)$ \\
65 & $O(1)$ \; vs \; $O(\mathrm{log}(n))$ \\
51 & $O(n^2+m)$ \; vs \; $O(n+m^2)$ \\
\bottomrule
\end{tabular}
\end{table}

\subsection{Ranking Stability Under Simulated Systematic Label Corruption}
\label{app:label_error_corruption}

We explore two forms of label perturbations. On prediction we flip the \emph{ground-truth} label of a (problem, class) pair; that label is a property of the test set, so the flip is \textbf{shared} across models. On generation we flip the framework's verdict on a model's own generated program; each model's code gets its own framework call, so the flip is \textbf{independent} across models (on generation the target complexity is given in the prompt, so there is no hidden ground-truth label to corrupt in the first place, only the framework's measurement of what the model produced). Flips can happen in three modes: \emph{random} (transform into any potential class), \emph{pessimistic} (one step slower) and \emph{optimistic} (one step faster). Boundary classes have no neighbor in one direction and can sometimes be left uncorrupted, so the realized rate can fall below the nominal one.

Since rank inversions occur under sampling noise alone, showing that corruption causes some inversions would prove nothing. The question is whether corruption is model-dependent, and therefore biases the evaluation. We therefore use a paired bootstrap over 1000 replicates, running three experiments: experiment A resamples problems without corrupting, experiment B corrupts without resampling, and experiment C does both, with experiments A and C sharing the same problem resample within a replicate so that their difference isolates the corruption effect from sampling noise.

\begin{figure*}[t!]
\centering
\includegraphics[width=\textwidth]{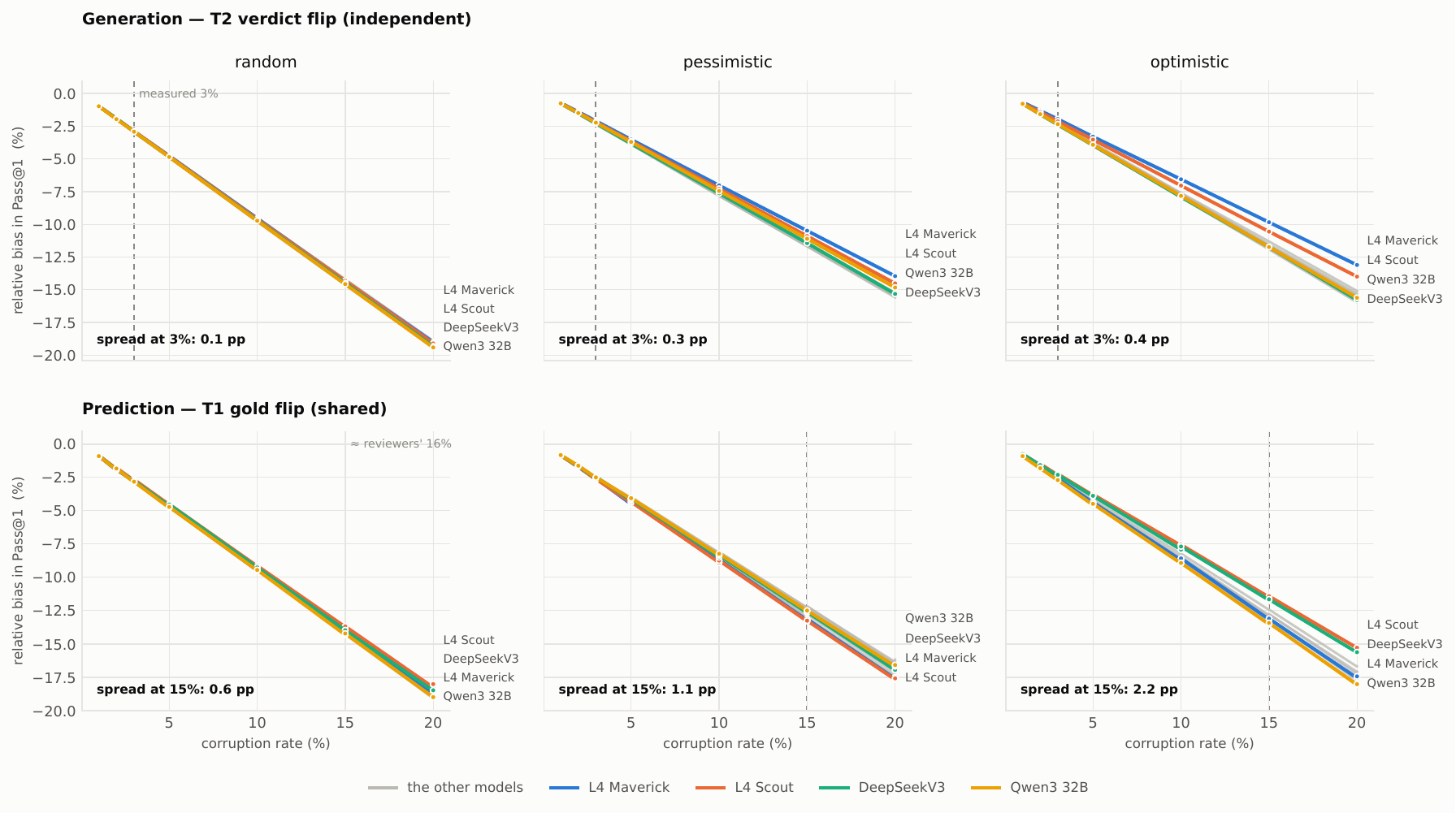}
\caption{Relative bias in $Pass@1$ against corruption rate on time complexity, for generation (top, verdict flips, independent across models) and prediction (bottom, ground-truth flips, shared across models), under the three corruption modes. Models that stay within the band are drawn as one grey series and only the deviating models are named. Dashed lines mark the anchor rates of 3\% on generation and 15\% on prediction, and each panel is annotated with the spread across models at that anchor. Label corruption moves every model by nearly the same proportion, so that the band collapses onto a single line. Since an inversion requires two models to be scaled by different factors, and that spread is of the order of a percentage point, only already-close pairs can reorder.}
\label{fig:corruption_time}
\end{figure*}

\begin{figure*}[t!]
\centering
\includegraphics[width=\textwidth]{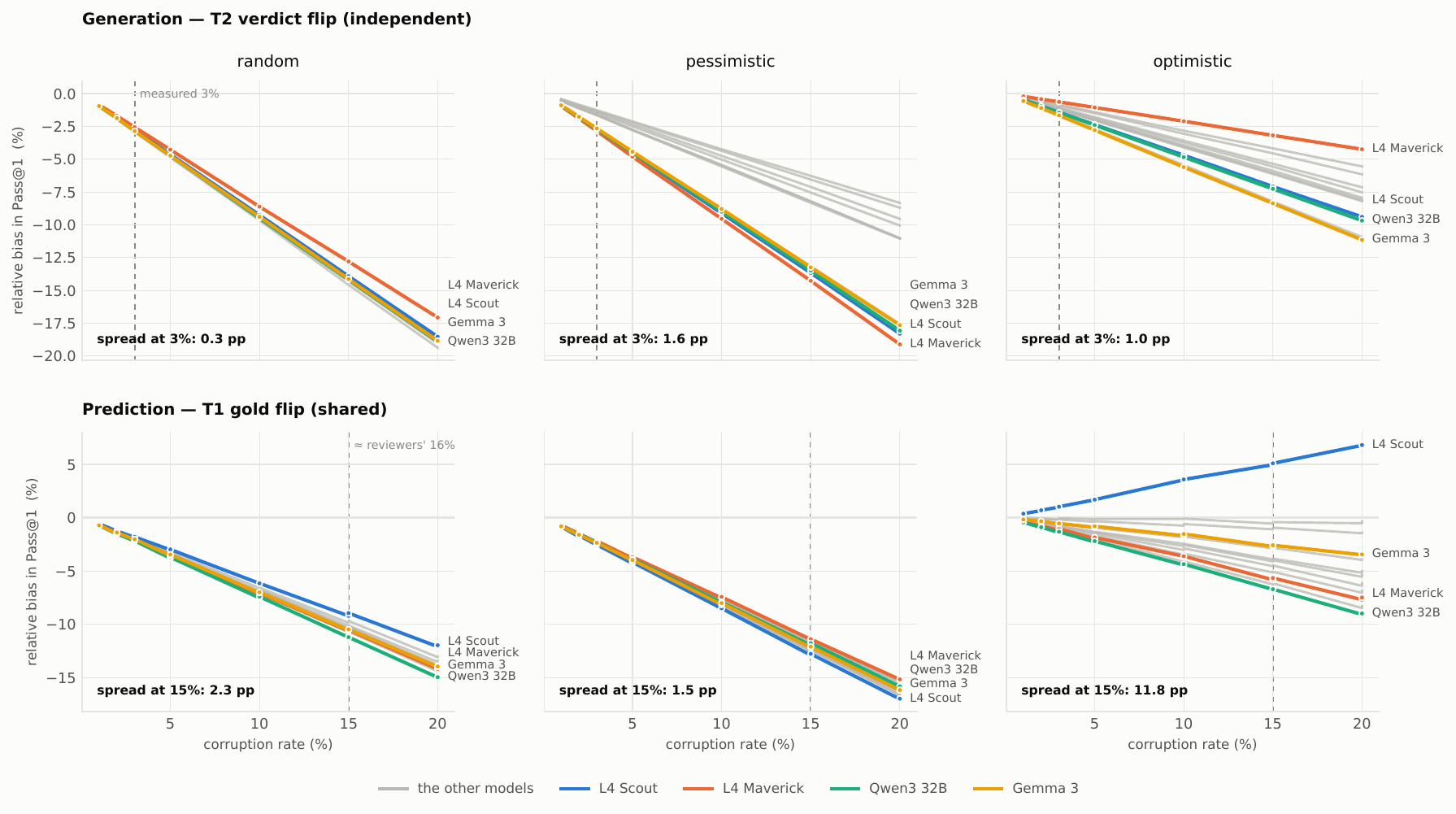}
\caption{The same sweep on space complexity, with conventions as in \cref{fig:corruption_time}. The one cell where models separate is space prediction under optimistic corruption, bottom right, where \textsc{Llama 4 Scout} gains while every other model loses; \cref{tab:scout_cell} further shares more numbers on this specific case.}
\label{fig:corruption_space}
\end{figure*}

\paragraph{Rankings under corruption.}
\Cref{fig:corruption_time,fig:corruption_space} show the relative bias in $Pass@1$ against corruption rate for every model. The lines collapse onto each other: corruption at rate $p$ scales every model's score by very nearly the same factor. \Cref{tab:corruption_summary} gives scores for two particular rates: 3\% on the generation task, the measured verdict reproducibility error rate, and more aggressively 15\% for the prediction task, the closest sampled rate to the 16\% label error rate measured against humans.

The spread across models is at most 2.3 pp in eleven of the twelve cells. Since a rank inversion requires that factor to differ between two models, inversions are rare and confined to pairs that were already close: no cell has more than 3 inverting pairs, out of 45 or 66 depending on the task, except space prediction under optimistic error, where 9 of 66 invert.

The same sweep can also be run on generation at a 15\% target rate, and \cref{tab:corruption_generation_15} reports the outcome. The findings do not change: no cell has more than 3 inverting pairs out of 66. Spreads widen, up to 8.0 pp on space under pessimistic corruption, but a 15\% verdict error rate on generation is a particularly conservative hypothesis.

\begin{table}[!ht]
\centering\small
\caption{Effect of label corruption at the anchor rates: 3\% on generation, the measured verdict-reproducibility rate, and 15\% on prediction, the error rate of labels when compared to human labels. \emph{Spread} is the range of relative bias across models. A pair of models can only invert if the relative gap between their clean scores is smaller than this, so the smaller the spread, the closer two models must already have been for corruption to reorder them. For instance, on time generation the gap between \textsc{Llama 3.1 Nemotron-Ultra 253B} at 33.5 $Pass@1$ and \textsc{Qwen3 32B} at 29.1 is a relative 15\%, against a spread of 0.1 to 0.4 pp. Generation compares 12 models (66 pairs), time prediction 10 (45 pairs), space prediction 12 (66 pairs).}
\label{tab:corruption_summary}
\begin{tabular}{lllrrr}
\toprule
\textbf{Task / axis} & \textbf{Mode} & \textbf{Realized} & \textbf{Relative bias} & \textbf{Spread} & \textbf{Inversions} \\
\midrule
generation / time & random & 2.9\% & $-2.9$ to $-2.9$\% & 0.1 pp & 1/66 \\
& pessimistic & 2.3\% & $-2.3$ to $-2.1$\% & 0.3 pp & 2/66 \\
& optimistic & 2.2\% & $-2.4$ to $-1.9$\% & 0.4 pp & 1/66 \\
\midrule
generation / space & random & 2.7\% & $-2.9$ to $-2.6$\% & 0.3 pp & 0/66 \\
& pessimistic & 2.3\% & $-2.9$ to $-1.3$\% & 1.6 pp & 2/66 \\
& optimistic & 1.2\% & $-1.7$ to $-0.6$\% & 1.0 pp & 1/66 \\
\midrule
prediction / time & random & 15.1\% & $-14.3$ to $-13.7$\% & 0.5 pp & 0/45 \\
& pessimistic & 11.1\% & $-13.3$ to $-12.3$\% & 1.0 pp & 0/45 \\
& optimistic & 13.7\% & $-13.5$ to $-11.4$\% & 2.1 pp & 2/45 \\
\midrule
prediction / space & random & 15.1\% & $-11.2$ to $-8.9$\% & 2.3 pp & 0/66 \\
& pessimistic & 11.8\% & $-12.9$ to $-11.4$\% & 1.5 pp & 3/66 \\
& optimistic & 11.5\% & $-6.7$ to $+5.0$\% & 11.7 pp & 9/66 \\
\bottomrule
\end{tabular}
\end{table}

\begin{table}[!ht]
\centering\small
\caption{Effect of label corruption on the generation task at a 15\% target rate, five times the measured verdict-reproducibility rate of 3\%. Columns as in \cref{tab:corruption_summary}, over the same 12 models and 66 model pairs. Even at that rate no cell has more than 3 inverting pairs.}
\label{tab:corruption_generation_15}
\begin{tabular}{lllrrr}
\toprule
\textbf{Task / axis} & \textbf{Mode} & \textbf{Realized} & \textbf{Relative bias} & \textbf{Spread} & \textbf{Inversions} \\
\midrule
generation / time & random & 14.6\% & $-14.6$ to $-14.2$\% & 0.4 pp & 1/66 \\
& pessimistic & 11.4\% & $-11.7$ to $-10.5$\% & 1.2 pp & 2/66 \\
& optimistic & 11.2\% & $-11.9$ to $-9.8$\% & 2.1 pp & 1/66 \\
\midrule
generation / space & random & 13.5\% & $-14.6$ to $-12.8$\% & 1.8 pp & 1/66 \\
& pessimistic & 11.6\% & $-14.3$ to $-6.3$\% & 8.0 pp & 3/66 \\
& optimistic & 6.1\% & $-8.4$ to $-3.2$\% & 5.2 pp & 1/66 \\
\bottomrule
\end{tabular}
\end{table}

\paragraph{Space prediction under optimistic corruption.}
One cell of \cref{tab:corruption_summary}, space prediction under optimistic corruption with a relatively high rate of 15\%, seems more impacted than the other cases.

\Cref{tab:scout_cell} looks at that cell in detail. \textsc{Llama 4 Scout} is the only model that gains, and it has by far the largest share of failures one step below gold, 59.7\% against a field of 40.9 to 53.3\%; the ordering is near-monotone. The bias is therefore a measurable property of the kind of errors this model makes: it systematically under-predicts space complexity, so moving the gold downward through the optimistic corruption consistently gives it more credit. In addition, the top four models on space prediction sit within about 3 points of each other, against a bootstrap standard deviation of roughly 1.4, so that ranking was never well separated to begin with. Space complexity labels concentrate on $O(n)$ and $O(1)$, and one optimistic step on space complexity collapses $O(n)$ straight into $O(1)$: weaker models are precisely the ones that answer $O(1)$ and miss the subtlety that makes a solution linear, and the collapse hands them a large amount of credit at once. The task still holds at its extremes, since the best and worst of the twelve models, \textsc{Qwen3 32B} at 47.7 and \textsc{Llama 4 Scout} at 37.4 $Pass@1$, are 27\% apart in relative terms, more than twice the 11.7 pp spread. But the middle of the field sits between 43.2 and 45.6, and under this corruption those models become hard to tell apart. Space prediction is therefore the one task of \benchmark where we would read the ranking with more caution.

\begin{table}[!ht]
\centering\small
\caption{Space prediction under optimistic corruption at 15\%. Column two is the share of each model's failed predictions that sit one step below the ground truth, computed before any corruption. It orders the models almost exactly as the realized bias does, as the corruption transformation in this specific case corrects the by far most present source of error.}
\label{tab:scout_cell}
\begin{tabular}{lrr}
\toprule
\textbf{Model} & \textbf{Failures one step below gold} & \textbf{Relative bias} \\
\midrule
Llama 4 Scout & 59.7\% & $+5.0$\% \\
DeepSeek V3 671B & 53.3\% & $-1.1$\% \\
Codestral 22B & 52.8\% & $-2.9$\% \\
Gemma 3 27B & 52.8\% & $-2.6$\% \\
Llama 3.3 70B & 51.9\% & $-0.4$\% \\
Qwen2.5-Coder 32B & 49.9\% & $-3.8$\% \\
DeepSeekCoderV2 236B & 46.9\% & $-4.1$\% \\
DeepSeekR1 Qwen 32B & 45.0\% & $-4.5$\% \\
DeepSeekR1 Llama 70B & 44.6\% & $-5.1$\% \\
Llama 4 Maverick & 43.8\% & $-5.7$\% \\
Nemotron-Ultra 253B & 42.5\% & $-6.3$\% \\
Qwen3 32B & 40.9\% & $-6.7$\% \\
\bottomrule
\end{tabular}
\end{table}

\FloatBarrier

\subsection{Sharing Label Errors Across Model Evaluations}
\label{app:label_error_shared}

\Cref{tab:corruption_summary} confirmed that a very limited number of inversions happen when corruption rates of the ground truth labels increase, even up to rates around 15\%. Why? If we look at the complexity prediction task, a framework complexity label is a property of the (problem, solution) pair and not of the model being evaluated, so a wrong label is wrong for every model at once. An error of that kind is therefore a shared burden: it moves all scores down together and largely cancels in the pairwise difference that a ranking actually depends on. Sharing does not make the effect identical across models, and \cref{tab:scout_cell} is the case where it does not, but it is what keeps the residual differences at the scale of a percentage point rather than at the scale of the error rate itself.

The concern that a 16\% error rate could swamp a few points between two models would be assuming the opposite, that each model draws its own set of bad labels. To get a sense of how much inflation this creates on score deviation, we also measured corruption under the independence hypothesis, and compared it with our shared-error experiments. \Cref{tab:shared_vs_independent} reports the standard deviation of the difference between two models' scores across replicates, with no problem resampling so that corruption is the only thing varying. Treating the error as independent per model inflates that standard deviation by a factor of 1.8 to 3.0, and comparing a per-item error rate (such as the 16\% implied by the accuracy reported in \cref{complexity_framework_performance}) directly against a gap between two models is exactly that independent reading.

In the case of the complexity generation task, the label attaches to each model's own program, so it is not shared in the same literal sense. It is produced by the same framework, though: any systematic bias in how the framework reads a given pattern is the same bias for every model rather than an independent draw per model, and only the run-to-run randomness is genuinely independent. That randomness is the 3\% measured in \cref{app:label_error_rate}, not the 84\%/82\% figure, which counts systematic and random error together. On both tasks, a per-label error rate therefore overstates the level of potential ranking errors.

\begin{table}[!ht]
\centering\small
\caption{Standard deviation of the difference in $Pass@1$ between two models, under corruption of a shared ground-truth label against corruption drawn independently for each model. Prediction task, 15\% corruption, 400 replicates, averaged over all 45 model pairs. Treating label error as independent, which is what comparing a per-item error rate to a model gap implicitly does, overstates the uncertainty on that gap by two to three times.}
\label{tab:shared_vs_independent}
\begin{tabular}{llrrr}
\toprule
\textbf{Axis} & \textbf{Mode} & \textbf{shared} & \textbf{independent} & \textbf{inflation} \\
\midrule
time & random & 0.57 & 1.52 & $\times 2.6$ \\
time & pessimistic & 0.48 & 1.42 & $\times 3.0$ \\
time & optimistic & 0.57 & 1.46 & $\times 2.6$ \\
space & random & 0.70 & 1.46 & $\times 2.1$ \\
space & pessimistic & 0.62 & 1.33 & $\times 2.1$ \\
space & optimistic & 0.74 & 1.36 & $\times 1.8$ \\
\bottomrule
\end{tabular}
\end{table}

\subsection{Putting Together All Sources of Variance}
\label{app:label_error_variance}

Finally, we measure label error against the other sources of uncertainty present in the same experiment. We isolate three of them by resampling one source at a time over 1000 replicates: the finite number of problems, the 20 generations per task, and label noise at the anchor rate.

\begin{table}[!ht]
\centering\small
\caption{Contribution of each source of uncertainty to the standard deviation of $Pass@1$, averaged over models and reported separately for each task and axis. Problem sampling resamples the 311 (time) or 308 (space) problems; the generation axis resamples the 20 attempts per task; label noise corrupts at 15\% on prediction, and on generation is given at both the measured 3\% and at 15\%, written 3\% / 15\%. Twelve models per row, except time prediction which has ten. On generation, label noise contributes roughly one seventeenth of the variability that the finite test set does; on prediction it is larger but still the smaller of the two.}
\label{tab:variance_budget}
\begin{tabular}{llrrrr}
\toprule
\textbf{Task} & \textbf{Axis} & \textbf{problem sampling} & \textbf{20 generations} & \textbf{label noise} & \textbf{ratio} \\
\midrule
generation & time & 1.23 & 0.21 & 0.07 / 0.15 & 0.06 / 0.12 \\
generation & space & 1.13 & 0.19 & 0.07 / 0.14 & 0.06 / 0.13 \\
\midrule
prediction & time & 1.89 & 0.14 & 1.10 & 0.58 \\
prediction & space & 1.42 & 0.15 & 1.03 & 0.72 \\
\bottomrule
\end{tabular}
\end{table}

Looking at \cref{tab:variance_budget}, on generation label noise at 3\% contributes about one seventeenth of the standard deviation that finite test-set size does, and still only about one eighth of it at 15\%. On prediction it is a larger term but still smaller than problem sampling, at 0.58 of it on time and 0.72 on space, and that comparison is made at 15\%, which is a conservative hypothesis. The 20-generation term is negligible throughout, which incidentally supports the choice of 20 samples for the $@k$ estimators. These are means over models, and the underlying values are tight: the ratio in the last column spans 0.05 to 0.07 at the measured rate across all twelve models on generation, 0.56 to 0.62 on time prediction and 0.62 to 0.80 on space prediction.

Overall, the 84\%/82\% figure is the right number for asking whether a label is true, and the wrong one for asking whether the benchmark ranks models correctly. Label error at that scale is acceptable for a widely used benchmark, and larger rates elsewhere have rarely reordered anything. Corrupting the labels deliberately, at rates up to 20\% and in three directions, scales every model by nearly the same factor and leaves at most 3 of 66 pairs inverted, the single exception being space prediction under optimistic corruption. The robustness of the benchmark comes from the fact that the error is carried by the labels and by the framework, both shared across models (not trained on the framework, and only exposed to it at test time), so it largely cancels in the differences a ranking depends on. Against the other sources of noise in the same experiment it is the smallest term on generation, and still the smaller of the two on prediction at a conservative 15\%. A larger expert-annotated test set remains the way forward for future and more important efforts in this research direction, but the model comparisons of \cref{table:complexityeval} do not depend on it.

\clearpage

\section{Ablations on Task Prompts}
\label{ablations_prompts}

We ran these prompt ablations with \textsc{Qwen QwQ 32B} as the reasoning model and \textsc{Llama 3.3 70B} as the non-reasoning model. This section is therefore a prompt-sensitivity study: unlike the main benchmark tables, which report \textsc{Qwen3 32B} among the evaluated reasoning models, these ablation results were not recomputed for \textsc{Qwen3 32B}. We keep the QwQ scores here only where they correspond to actually run ablations. The following subsections provide the types of prompts that were used (every time the example corresponds to the time complexity variant, given that the space complexity variant can easily be derived from it; similarly, given the complexity prediction variant, the complexity generation variant can also be derived in a simple manner).

\subsection{Original BigOBench Prompt - Time Complexity Prediction}

Provide the time complexity for the following competitive programming question and corresponding solution.

When analyzing the complexity of an algorithm, consider the worst-case scenario where all possible input combinations are tried, given the following conditions: 1. the inputs must adhere to the specified data types of the problem; 2. the inputs should not cause the code to crash or exit on an exception; 3. the inputs do not necessarily need to satisfy additional constraints that are potentially mentioned in the problem statement; 4. calling input() does not consume runtime nor memory, but of course any operations on top of it or afterwards will be counted towards runtime and memory footprint; 5. Anything printed gets added to the memory. You can take advantage of Python-specific optimizations provided by the underlying CPython interpreter or compiler to achieve the desired complexity, and you must account for them when analyzing the complexity.

Here is the programming question: {context}.

Here is the corresponding Python solution: {code\_content}.

Please ignore any constraints on the input sizes that may have been previously mentioned in the problem description. Compute the big-O complexity as if inputs can be as large as possible.

Output the big-O time complexity only, no explanation needed, no other words needed.

\subsection{Big-Theta Variant - Time Complexity Prediction}

Provide the time complexity for the following competitive programming question and corresponding solution.

When analyzing the complexity of an algorithm, consider the worst-case scenario where all possible input combinations are tried, given the following conditions: 1. the inputs must adhere to the specified data types of the problem; 2. the inputs should not cause the code to crash or exit on an exception; 3. the inputs do not necessarily need to satisfy additional constraints that are potentially mentioned in the problem statement; 4. calling input() does not consume runtime nor memory, but of course any operations on top of it or afterwards will be counted towards runtime and memory footprint; 5. Anything printed gets added to the memory. You can take advantage of Python-specific optimizations provided by the underlying CPython interpreter or compiler to achieve the desired complexity, and you must account for them when analyzing the complexity.

Here is the programming question: {context}.

Here is the corresponding Python solution: {code\_content}.

Please ignore any constraints on the input sizes that may have been previously mentioned in the problem description. Compute the big-Theta complexity as if inputs can be as large as possible.

Output the big-Theta time complexity only, no explanation needed, no other words needed.

\subsection{Detailed Big-O variant - Time Complexity Prediction}

Provide the time complexity for the following competitive programming question and corresponding solution.

When analyzing the complexity of an algorithm, consider the worst-case scenario where all possible input combinations are tried, given the following conditions: 1. the inputs must adhere to the specified data types of the problem; 2. the inputs should not cause the code to crash or exit on an exception; 3. the inputs do not necessarily need to satisfy additional constraints that are potentially mentioned in the problem statement; 4. calling input() does not consume runtime nor memory, but of course any operations on top of it or afterwards will be counted towards runtime and memory footprint; 5. Anything printed gets added to the memory. You can take advantage of Python-specific optimizations provided by the underlying CPython interpreter or compiler to achieve the desired complexity, and you must account for them when analyzing the complexity.

Here is the programming question: {context}.

Here is the corresponding Python solution: {code\_content}.

Please ignore any constraints on the input sizes that may have been previously mentioned in the problem description. Compute the big-O complexity as if inputs can be as large as possible.

The big-O complexity is the time complexity of the program when running on the worst case, which means you may want to find the worst running case first, and then find its corresponding time complexity, that you need to output in big-O format. For example, if the worst case is in the order of nlogn, do output precisely O(nlogn), as O(n**2) will be considered too high of an upper bound (though mathematically speaking one could say that what is O(nlogn) is also O(n**2)). Find the tightest possible upper bound in big-O notation.

Output the big-O time complexity only, no explanation needed, no other words needed.

\subsection{COT Big-O variant - Time Complexity Prediction}

Provide the time complexity for the following competitive programming question and corresponding solution.

When analyzing the complexity of an algorithm, consider the worst-case scenario where all possible input combinations are tried, given the following conditions: 1. the inputs must adhere to the specified data types of the problem; 2. the inputs should not cause the code to crash or exit on an exception; 3. the inputs do not necessarily need to satisfy additional constraints that are potentially mentioned in the problem statement; 4. calling input() does not consume runtime nor memory, but of course any operations on top of it or afterwards will be counted towards runtime and memory footprint; 5. Anything printed gets added to the memory. You can take advantage of Python-specific optimizations provided by the underlying CPython interpreter or compiler to achieve the desired complexity, and you must account for them when analyzing the complexity.

Here is the programming question: {context}.

Here is the corresponding Python solution: {code\_content}.

Please ignore any constraints on the input sizes that may have been previously mentioned in the problem description. Compute the big-O complexity as if inputs can be as large as possible.

Break down the analysis into these steps: 1. Identify all loops, recursive calls, and operations dependent on input size; 2. Explicitly state the time complexity of each Python-specific operation (e.g., list appends, dictionary lookups, sorting); 3. Assume all inputs are adversarially chosen to maximize runtime (e.g., hash collisions, worst-case comparisons); 4. Account for memory allocations (e.g., dynamic array resizes, string concatenation costs); 5. If recursion is used, include the cost of stack frames and potential tail-call optimizations; 6. Treat all arithmetic operations as O(1), but flag if arbitrary-precision integers could introduce hidden costs; 7. For nested operations, multiply complexities conservatively (e.g., O(n) loops inside O(n) loops = O(n²)); 8. Ignore Python's global interpreter lock (GIL) and concurrency effects; 9. Explicitly confirm whether built-in functions like sorted() or re.search() are treated as black-box with known complexities; 10. Final answer must be the tightest possible upper bound in big-O notation, even if the problem's original constraints imply smaller inputs.

Output the big-O time complexity at the end.

\subsection{Original BigOBench Prompt - Time Complexity Ranking}

Provide a Python solution for the following competitive programming question: {context}.

Output the code only. Generate code that has an algorithmic time complexity of {time\_complexity}. Try to optimize the runtime of your code as much as you can, while respecting the time complexity requirement.

When analyzing the complexity of an algorithm, consider the worst-case scenario where all possible input combinations are tried, given the following conditions: 1. the inputs must adhere to the specified data types of the problem; 2. the inputs should not cause the code to crash or exit on an exception; 3. the inputs do not necessarily need to satisfy additional constraints that are potentially mentioned in the problem statement; 4. calling input() does not consume runtime nor memory, but of course any operations on top of it or afterwards will be counted towards runtime and memory footprint; 5. Anything printed gets added to the memory. You can take advantage of Python-specific optimizations provided by the underlying CPython interpreter or compiler to achieve the desired complexity, and you must account for them when analyzing the complexity.

Your code should be enclosed in triple backticks like so: ```python YOUR CODE HERE ```. Use the backticks for your code only.

\subsection{Python Specifications Big-O variant - Time Complexity Ranking}

Provide a Python solution for the following competitive programming question: {context}.

Output the code only. Generate code that has an algorithmic time complexity of {time\_complexity}. Try to optimize the runtime of your code as much as you can, while respecting the time complexity requirement.

When analyzing the complexity of an algorithm, consider the worst-case scenario where all possible input combinations are tried, given the following conditions: 1. the inputs must adhere to the specified data types of the problem; 2. the inputs should not cause the code to crash or exit on an exception; 3. the inputs do not necessarily need to satisfy additional constraints that are potentially mentioned in the problem statement; 4. calling input() does not consume runtime nor memory, but of course any operations on top of it or afterwards will be counted towards runtime and memory footprint; 5. Anything printed gets added to the memory. You can take advantage of Python-specific optimizations provided by the underlying CPython interpreter or compiler to achieve the desired complexity, and you must account for them when analyzing the complexity.

As additional specifications: 1. Python Version - The code must be compatible with Python 3.10.0, avoid using features deprecated in this version or introduced in later versions, you can leverage the optimization that are proper to this version; 2. Libraries Allowed - You may use standard libraries (e.g., itertools, collections, math) and third-party libraries (e.g., NumPy, pandas) to optimize performance (For example, use numpy for vectorized operations on numerical data (e.g., np.where, np.concatenate), use collections.defaultdict for faster dictionary-like structures with default values…); 3. Memory Constraints - Avoid unnecessary copies of large data structures, Prefer in-place operations where possible (e.g., list.sort() instead of sorted() for large lists); 4. Edge Cases - Assume inputs adhere to the problem's data types and constraints (e.g., non-negative integers, valid strings) but do not need to handle invalid cases (e.g., non-integer inputs where integers are expected); 5. Output Handling - Minimize printing intermediate results to reduce memory overhead. Use generators or lazy evaluation (e.g., yield) for large datasets. Your code must still adhere strictly to the specified time complexity and avoid unnecessary operations.

Your code should be enclosed in triple backticks like so: ```python YOUR CODE HERE ```. Use the backticks for your code only.

\subsection{Ablation Results}

Taking all these variations of the tasks' formulations into account, \benchmark scores can be re-evaluated for each model. \cref{table:complexityeval_prompt_ablations} compares prompt results across complexity prediction and generation, while \cref{table:complexityeval_prompt_ablations_ranking} looks at the complexity ranking task. For all tasks, only the time complexity version was tried out.

As seen in these tables, the reasoning model's performance on Time Complexity Prediction decreased slightly with the big-Theta and detailed big-O variants, and more sharply with the COT-prompted variant, reaching an All@1 score of 31.8 (compared to the main \benchmark score of 40.4). The non-reasoning model's performance also decreased slightly with the big-Theta variant, but improved with the detailed big-O and COT variants, reaching 36.1 All@1. We believe that sharing more details with the reasoning model leads to overthinking, while the non-reasoning model benefits from the added context. The non-reasoning model's performance, even with COT, remains lower than the reasoning model's.

In the case of Time Complexity Generation, the non-reasoning model falls short of seeing any improvements with the different prompt variants. We believe the performance is so low (3.3 All@1 for Llama 3.3 70B) that even sharing slightly more details cannot help a model that simply does not grasp the objective of the task. In the case of the reasoning model, we do see an improvement using the variants, up to scores of 11.7 All@1 with the big-Theta variant. The base reported score of Qwen QwQ on this task is 9.6 All@1, which suggests that the model starts to grasp the task, though it lacks guidance compared to the Complexity Prediction equivalent (40.4 All@1). In this case, we believe the added details to the prompt help the model perform better while staying far from overthinking. It also helps put more focus on the complexity requirement compared to code correctness, given that a manual error analysis underlined the models often lose focus from the first objective during the reasoning process.

Finally, adding Python specifications to the prompt resulted in a marginal improvement in performance (up to ~+2 coeffFULL for Qwen QwQ).

\begin{table*}[h]
\centering\small
\caption{
\benchmark benchmark results for variation of the task prompts on a reasoning model (\textsc{Qwen QwQ 32B}) and a non-reasoning model (\textsc{Llama 3.3 70B}). \textbf{Program Synthesis} checks the correctness of model-generated solutions to given programming problems, not taking into account any complexity requirement. \textbf{Complexity Prediction} measures whether a model can find the time-space complexity of an existing code snippet. \textbf{Complexity Generation} evaluates whether a model can output a working code snippet to a given problem that meets a time-space complexity requirement. $Pass@k$ considers each complexity class of all problems independently and calculates a macro-average between them. $Best@k$ is a refinement of $Pass@k$, focusing only on the most optimized complexity class for each problem. $All@k$ checks if all complexity classes for each problem are correctly predicted or generated simultaneously, then macro-averages across all problems.
}
\label{table:complexityeval_prompt_ablations}
\begin{sc}
\resizebox{0.95\linewidth}{!}{
\begin{tabular}{llcclllllllll}
\toprule
\multirow{3}{*}{\textbf{Model}} & ~ & \multicolumn{2}{c}{\textbf{Prog. Synthesis}} & ~ & \multicolumn{3}{c}{\textbf{Complexity Pred.}} & ~ & \multicolumn{4}{c}{\textbf{Complexity Gen.}} \\
 & & \multicolumn{1}{c}{Pass} & \multicolumn{1}{c}{Pass} & & \multicolumn{1}{c}{Pass} & \multicolumn{1}{c}{Best} & \multicolumn{1}{c}{All} & & \multicolumn{1}{c}{Pass} & \multicolumn{1}{c}{Pass} & \multicolumn{1}{c}{Best}& \multicolumn{1}{c}{All} \\
 & & \multicolumn{1}{c}{\bf $@1$} & \multicolumn{1}{c}{\bf $@10$} & & \multicolumn{1}{c}{\bf $@1$}  &\multicolumn{1}{c}{\bf $@1$}&\multicolumn{1}{c}{\bf $@1$}& & \multicolumn{1}{c}{\bf $@1$} & \multicolumn{1}{c}{\bf $@10$} &\multicolumn{1}{c}{\bf $@1$} & \multicolumn{1}{c}{\bf $@1$} \\
\hline\vspace{-0.1in} \\
\textbf{Qwen QwQ} & & \\
~~BigOBench Prompt & & 60.7 & 75.8  & & 62.7 & 72.2 & 40.3 & & 26.9 & 50.9 & 37.8 & 8.3  \\
~~Big Theta & & 58.4 & 74.8 & & 62.0 & 71.4 & 39.9  & & 28.8 & 52.1 & 37.3 & 11.7 \\
~~BigO with More Details & & 58.4 & 74.7 & & 61.9 & 70.8 & 39.5 & & 26.7 & 51.3 & 36.7 & 8.8 \\
~~BigO with COT  & & 57.9 & 74.1 & & 57.1 & 62.9 & 31.8  & & 27.0 & 51.6 & 36.3 & 9.7 \\
\hline\vspace{-0.1in} \\
\textbf{Llama 3.3} & & \\
~~BigOBench Prompt  & & 34.0 & 54.9 & & 58.0 & 72.3 & 33.7 & & 13.1 & 31.3 & 19.1 & 2.1\\
~~Big Theta  & & 33.5 & 54.8 & & 57.2 & 72.7 & 32.1  & & 12.3 & 30.8 & 18.5 & 1.8 \\
~~BigO with More Details & & 33.5 & 54.5 & & 59.9 & 73.6 & 35.5 & & 12.5 & 31.1 & 18.3 & 1.9 \\
~~BigO with COT  & & 37.0 & 62.4 & &  59.3 & 68.1 & 36.1 & & 14.1 & 37.5 & 20.3 & 2.2  \\
\bottomrule
\end{tabular}
}

\end{sc}\end{table*}

\begin{table*}[h]
\centering\small
\caption{
\benchmark benchmark results for variation of the task prompts on a reasoning model (\textsc{Qwen QwQ 32B}) and a non-reasoning model (\textsc{Llama 3.3 70B}). \textbf{Program Synthesis} checks the correctness of model-generated solutions to given programming problems, not taking into account any complexity requirement. \textbf{Complexity Ranking} evaluates whether a model can output a working code snippet to a given problem that meets a time-space complexity requirement, while being as optimized as possible within this complexity class of solutions. $Pass@k$ considers each complexity class of all problems independently and calculates a macro-average between them. $Best@k$ is a refinement of $Pass@k$, focusing only on the most optimized complexity class for each problem. $All@k$ checks if all complexity classes for each problem are correctly predicted or generated simultaneously, then macro-averages across all problems. Finally for $CoeffFull$, using the complexity framework, the best measured coefficient of the complexity curve, out of 20 attempts, is used to rank LLM-generated code among human solutions from the same problem and time-space complexity class. Ranking is percentile based, n\% ranking score amounts for n\% human solutions having worse complexity coefficient. If no LLM solution passes correctness tests, ranking score is set to 0.
}
\label{table:complexityeval_prompt_ablations_ranking}
\begin{sc}
\resizebox{0.95\linewidth}{!}{
\begin{tabular}{llcclllllc}
\toprule
\multirow{3}{*}{\textbf{Model}} & ~ & \multicolumn{2}{c}{\textbf{Program Synthesis}} & ~ & \multicolumn{5}{c}{\textbf{Complexity Ranking}} \\
 & & \multicolumn{1}{c}{Pass} & \multicolumn{1}{c}{Pass} & & \multicolumn{1}{c}{Pass} & \multicolumn{1}{c}{Pass} & \multicolumn{1}{c}{Best} & \multicolumn{1}{c}{All} & \multicolumn{1}{c}{CoeffFull} \\
 & & \multicolumn{1}{c}{\bf $@1$} & \multicolumn{1}{c}{\bf $@10$} & & \multicolumn{1}{c}{\bf $@1$} & \multicolumn{1}{c}{\bf $@10$} &\multicolumn{1}{c}{\bf $@1$} & \multicolumn{1}{c}{\bf $@1$} & \\
\hline\vspace{-0.1in} \\
\textbf{Qwen QwQ} & & \\
~~BigOBench Prompt & & 68.8 & 82.8 & & 27.7 & 54.8 & 40.6 & 6.9 & 48.1 \\
~~BigO with Python Details & & 68.9 & 82.5 & & 26.6 & 55.3 & 37.7 & 6.6 & 49.8 \\
\hline\vspace{-0.1in} \\
\textbf{Llama 3.3} & & \\
~~BigOBench Prompt  & & 36.3 & 58.8 & & 13.6 & 33.4 & 20.1 & 2.1 & 29.5 \\
~~BigO with Python Details & & 33.8 & 57.0 & & 10.9 & 32.8 & 16.4 & 1.4 & 30.2 \\
\bottomrule
\end{tabular}
}

\end{sc}\end{table*}

\clearpage

\section{Multi-sample Results Scores}
\label{multi_sample_results_scores}

For any of the metrics defined for the tasks of \textbf{Complexity Prediction}, \textbf{Complexity Generation} and \textbf{Complexity Ranking}, the models are queried multiple times to estimate performance under a larger sampling budget. In the specific case of \textbf{Complexity Prediction}, we only report $@1$ scores: this task remains a classification problem with a small and fixed output space, so repeated sampling would mostly conflate model uncertainty with true performance, instead of reflecting exploration capabilities such as in code generation.

As defined in \cref{evaluation_section} and in \cref{format_metrics}, we define \textbf{Pass@k} for Complexity Generation for example following the definition of \citet{chen2021evaluating}:
\[
\text{Pass@k} = \mathbb{E}_{P \sim \mathcal{P}}\left[ \frac{1}{|\mathcal{C}(P)|} \sum_{c \in \mathcal{C}(P)} \left( 1 - \frac{\dbinom{n - c_c}{k}}{\dbinom{n}{k}} \right) \right],
\]
where \( c_c = \sum_{i=1}^n \mathbb{I}_{\text{class}}(S'_i, c, P) \times \mathbb{I}_{\text{correct}}(S'_i, P) \).\\

\begin{table*}[h]
\centering\small
\caption{
\benchmark benchmark results for popular LLMs on the tasks of \textbf{Program Synthesis} and \textbf{Complexity Generation}, with multiple $@k$ estimators. \textbf{Program Synthesis} checks the correctness of model-generated solutions to given programming problems, not taking into account any complexity requirement. \textbf{Complexity Generation} evaluates whether a model can output a working code snippet to a given problem that meets a time-space complexity requirement. $Pass@k$ considers each complexity class of all problems independently and calculates a macro-average between them. $Best@k$ is a refinement of $Pass@k$, focusing only on the most optimized complexity class for each problem. $All@k$ checks if all complexity classes for each problem are correctly predicted or generated simultaneously, then macro-averages across all problems.
}
\label{table:complexityeval_more_samples}
\begin{sc}
\resizebox{\linewidth}{!}{
\begin{tabular}{llcccllllllllllll}
\toprule
\multirow{3}{*}{\textbf{Model}} & ~ & \multicolumn{3}{c}{\textbf{Prog. Synthesis}} & ~ & \multicolumn{11}{c}{\textbf{Complexity Gen.}} \\
 & & \multicolumn{1}{c}{Pass} & \multicolumn{1}{c}{Pass} & \multicolumn{1}{c}{Pass} & & \multicolumn{1}{c}{Pass} & \multicolumn{1}{c}{Pass} & \multicolumn{1}{c}{Pass} & & \multicolumn{1}{c}{Best} & \multicolumn{1}{c}{Best}  & \multicolumn{1}{c}{Best} & & \multicolumn{1}{c}{All} & \multicolumn{1}{c}{All} & \multicolumn{1}{c}{All} \\
 & & \multicolumn{1}{c}{\bf $@1$} & \multicolumn{1}{c}{\bf $@5$} & \multicolumn{1}{c}{\bf $@10$} & & \multicolumn{1}{c}{\bf $@1$} & \multicolumn{1}{c}{\bf $@5$} &\multicolumn{1}{c}{\bf $@10$} & & \multicolumn{1}{c}{\bf $@1$} & \multicolumn{1}{c}{\bf $@5$} & \multicolumn{1}{c}{\bf $@10$} & & \multicolumn{1}{c}{\bf $@1$} & \multicolumn{1}{c}{\bf $@5$} & \multicolumn{1}{c}{\bf $@10$} \\
\hline\vspace{-0.1in} \\
\textbf{Time} & & \\
~~Baselines & &  30.3 & 48.5 & 55.4  & &  12.1 & 23.7 & 29.7 & & 19.0 & 33.5 & 40.3 & & 0.9 & 3.7 & 6.0 \\
~~Llama 3.3 70B & &   43.4 & 60.7 & 66.0  & &  17.7 & 32.9 & 40.0 & & 25.7 & 41.8 & 47.8 & & 3.3 & 9.5 & 13.9 \\
~~Llama 4 Scout 17Bx16E & &  61.7 & 76.9 & 80.1  & &  22.8 & 40.5 & 48.0 & & 31.8 & 50.8 & 58.2 & & 3.5 & 10.7 & 16.2 \\
~~Llama 4 Maverick 17Bx128E & &   59.3 & 74.6 & 78.9  & &  19.9 & 37.0 & 44.6 & & 27.0 & 45.9 & 52.4 & & 5.3 & 12.1 & 17.1 \\
~~Gemma 3 27B & &   37.7 & 43.7 & 45.6  & &  15.2 & 19.3 & 20.8 & & 17.6 & 21.9 & 23.4 & & 1.8 & 3.2 & 3.9 \\
~~Codestral 22B & &  23.7 & 40.5 & 47.5  & &  10.6 & 21.2 & 26.6 & & 14.9 & 27.6 & 33.8 & & 1.3 & 4.3 & 6.4 \\
~~Qwen2.5-Coder 32B & &  30.5 & 45.3 & 50.8  & &  12.2 & 22.0 & 26.5 & & 15.2 & 26.8 & 31.5 & & 3.1 & 6.6 & 8.7 \\
~~GPT-4o & &  51.0 & 73.1 & 78.3  & &  20.6 & 37.5 & 44.7 & & 30.2 & 51.9 & 59.5 & & 4.3 & 11.7 & 16.4 \\
~~DeepSeekCoderV2 236B & &  44.1 & 60.5 & 65.5  & &  19.5 & 32.3 & 38.0 & & 27.6 & 43.0 & 48.7 & & 3.3 & 8.1 & 11.8 \\
~~DeepSeekV3 671B & &  41.4 & 58.4 & 63.6  & &  17.7 & 31.5 & 37.7 & & 23.0 & 39.6 & 46.2 & & 3.4 & 8.9 & 12.2 \\
~~DeepSeekR1 Qwen 32B & &  70.1 & 81.2 & 83.7  & &  29.0 & 44.1 & 49.9 & & 46.1 & 61.7 & 66.0 & & 4.8 & 14.7 & 21.0 \\
~~DeepSeekR1 Llama 70B & &  70.1 & 81.3 & 83.8  & &  29.2 & 45.3 & 51.6 & & 46.5 & 63.4 & 68.4 & & 4.8 & 15.5 & 22.6 \\
~~Llama 3.1 Nemotron-Ultra 253B & &  80.0 & 88.1 & 89.4  & &  33.5 & 49.0 & 54.9 & & 51.8 & 66.5 & 71.5 & & 6.1 & 16.8 & 23.2 \\
~~Qwen3 32B & & 70.0 & 79.4 & 81.3  & &  29.1 & 47.2 & 53.8 & & 43.5 & 61.9 & 67.2 & & 6.5 & 19.0 & 26.1\\
\hline\vspace{-0.1in} \\
\textbf{Space} & & \\
~~Baselines & &  30.1 & 46.8 & 52.6  & &  12.2 & 25.3 & 32.4 & & 17.8 & 32.7 & 40.0 & & 1.3 & 5.4 & 8.9 \\
~~Llama 3.3 70B & &  42.6 & 58.1 & 62.5  & &  15.0 & 30.6 & 37.7 & & 21.9 & 38.7 & 45.2 & & 1.8 & 6.2 & 10.0\\
~~Llama 4 Scout 17Bx16E & &  56.7 & 70.3 & 73.5  & &  20.0 & 34.3 & 40.1 & & 31.1 & 47.9 & 53.9 & & 2.2 & 8.2 & 12.8 \\
~~Llama 4 Maverick 17Bx128E & &  58.4 & 72.0 & 75.7  & &  16.8 & 24.7 & 28.2 & & 30.4 & 41.2 & 45.2 & & 0.8 & 3.0 & 4.6 \\
~~Gemma 3 27B & &  40.3 & 46.9 & 49.0  & &  16.2 & 22.1 & 24.3 & & 22.5 & 29.8 & 31.9 & & 1.4 & 3.5 & 5.1 \\
~~Codestral 22B & &  25.7 & 41.4 & 47.6  & &  11.0 & 23.3 & 29.4 & & 16.7 & 31.1 & 37.1 & & 1.3 & 5.3 & 8.8 \\
~~Qwen2.5-Coder 32B & &  31.1 & 44.6 & 49.2  & &  10.1 & 19.2 & 23.3 & & 15.3 & 26.4 & 30.5 & & 1.2 & 4.0 & 6.2 \\
~~GPT-4o & &  51.6 & 70.8 & 74.4  & &  18.1 & 33.7 & 39.9 & & 28.0 & 47.0 & 53.0 & & 1.4 & 6.0 & 10.1 \\
~~DeepSeekCoderV2 236B & &  43.1 & 58.8 & 63.8  & &  16.7 & 29.0 & 34.5 & & 25.6 & 40.0 & 45.0 & & 1.0 & 4.1 & 7.0 \\
~~DeepSeekV3 671B & & 41.8 & 57.9 & 62.5  & &  15.0 & 29.1 & 35.4 & & 22.6 & 40.4 & 46.8 & & 1.6 & 5.6 & 8.8 \\
~~DeepSeekR1 Qwen 32B & &  68.0 & 78.7 & 80.6  & &  24.8 & 41.9 & 48.6 & & 38.6 & 58.7 & 64.3 & & 3.1 & 11.4 & 17.8\\
~~DeepSeekR1 Llama 70B & &  68.8 & 79.3 & 81.2  & &  25.6 & 43.4 & 50.0 & & 38.7 & 59.1 & 64.6 & & 3.4 & 12.2 & 18.8\\
~~Llama 3.1 Nemotron-Ultra 253B & &  77.7 & 85.2 & 86.4  & &  30.4 & 48.7 & 55.5 & & 45.3 & 65.6 & 70.9 & & 5.6 & 16.4 & 23.7 \\
~~Qwen3 32B & & 65.9 & 75.5 & 77.7  & &  25.5 & 41.9 & 47.8 & & 39.6 & 58.7 & 64.2 & & 5.1 & 14.0 & 19.5\\
\bottomrule
\end{tabular}
}

\end{sc}\end{table*}

\begin{table*}[h]
\centering\small
\caption{
\benchmark benchmark results for popular LLMs on the tasks of \textbf{Complexity Generation}, with multiple $@k$ estimators, where $@5$ and $@10$ are displayed relative to the corresponding $@1$ score. \textbf{Complexity Generation} evaluates whether a model can output a working code snippet to a given problem that meets a time-space complexity requirement. $Pass@k$ considers each complexity class of all problems independently and calculates a macro-average between them. $All@k$ checks if all complexity classes for each problem are correctly predicted or generated simultaneously, then macro-averages across all problems.
}
\label{table:complexityeval_more_samples_relative}
\begin{sc}
\resizebox{\linewidth}{!}{
\begin{tabular}{lllrrllrr}
\toprule
\multirow{3}{*}{\textbf{Model}} & ~ & \multicolumn{7}{c}{\textbf{Complexity Gen.}} \\
 & & \multicolumn{1}{c}{Pass} & \multicolumn{1}{c}{Pass} & \multicolumn{1}{c}{Pass} & & \multicolumn{1}{c}{All} & \multicolumn{1}{c}{All} & \multicolumn{1}{c}{All} \\
 & & \multicolumn{1}{c}{\bf $@1$} & \multicolumn{1}{c}{\bf $@5$} & \multicolumn{1}{c}{\bf $@10$} & & \multicolumn{1}{c}{\bf $@1$} & \multicolumn{1}{c}{\bf $@5$} & \multicolumn{1}{c}{\bf $@10$} \\
\hline\vspace{-0.1in} \\
\textbf{Time} & & \\
~~Llama 3.3 70B & & 17.7 & +85.8\% & +126.2\% & & 3.3 & +183.4\% & +315.0\% \\
~~Llama 4 Scout 17Bx16E & & 22.8 & +77.4\% & +110.4\% & & 3.5 & +206.8\% & +365.2\% \\
~~Llama 4 Maverick 17Bx128E & & 19.9 & +86.4\% & +124.8\% & & 5.3 & +127.9\% & +221.0\% \\
~~Gemma 3 27B & & 15.2 & +27.1\% & +37.5\% & & 1.8 & +81.7\% & +122.6\% \\
~~Codestral 22B & & 10.6 & +100.5\% & +151.6\% & & 1.3 & +222.8\% & +382.2\% \\
~~Qwen2.5-Coder 32B & & 12.2 & +80.0\% & +116.6\% & & 3.1 & +114.1\% & +182.4\% \\
~~GPT-4o & & 20.6 & +81.5\% & +116.4\% & & 4.3 & +170.8\% & +280.0\% \\
~~DeepSeekCoderV2 236B & & 19.5 & +66.0\% & +95.0\% & & 3.3 & +141.1\% & +252.6\% \\
~~DeepSeekV3 671B & & 17.7 & +78.2\% & +113.3\% & & 3.4 & +157.4\% & +254.9\% \\
~~DeepSeekR1 Qwen 32B & & 29.0 & +52.1\% & +72.1\% & & 4.8 & +202.8\% & +333.3\% \\
~~DeepSeekR1 Llama 70B & & 29.2 & +55.2\% & +76.8\% & & 4.8 & +223.2\% & +371.3\% \\
~~Llama 3.1 Nemotron-Ultra 253B & & 33.5 & +46.4\% & +63.8\% & & 6.1 & +175.1\% & +279.8\% \\
~~Qwen3 32B & & 29.1 & +62.0\% & +84.7\% & & 6.5 & +190.7\% & +299.7\% \\
\hline\vspace{-0.1in} \\
\textbf{Space} & & \\
~~Llama 3.3 70B & & 15.0 & +103.9\% & +151.1\% & & 1.8 & +256.0\% & +471.5\% \\
~~Llama 4 Scout 17Bx16E & & 20.0 & +71.4\% & +100.5\% & & 2.2 & +279.3\% & +490.7\% \\
~~Llama 4 Maverick 17Bx128E & & 16.8 & +47.2\% & +68.5\% & & 0.8 & +266.9\% & +471.5\% \\
~~Gemma 3 27B & & 16.2 & +36.5\% & +50.1\% & & 1.4 & +145.5\% & +253.5\% \\
~~Codestral 22B & & 11.0 & +111.5\% & +166.7\% & & 1.3 & +319.6\% & +592.4\% \\
~~Qwen2.5-Coder 32B & & 10.1 & +90.3\% & +131.3\% & & 1.2 & +245.5\% & +434.2\% \\
~~GPT-4o & & 18.1 & +86.3\% & +120.5\% & & 1.4 & +318.7\% & +608.4\% \\
~~DeepSeekCoderV2 236B & & 16.7 & +73.1\% & +106.1\% & & 1.0 & +298.8\% & +572.4\% \\
~~DeepSeekV3 671B & & 15.0 & +93.4\% & +135.3\% & & 1.6 & +256.8\% & +459.8\% \\
~~DeepSeekR1 Qwen 32B & & 24.8 & +68.7\% & +95.6\% & & 3.1 & +264.3\% & +470.9\% \\
~~DeepSeekR1 Llama 70B & & 25.6 & +69.8\% & +95.3\% & & 3.4 & +259.0\% & +452.6\% \\
~~Llama 3.1 Nemotron-Ultra 253B & & 30.4 & +60.4\% & +82.6\% & & 5.6 & +192.3\% & +322.7\% \\
~~Qwen3 32B & & 25.5 & +63.9\% & +87.2\% & & 5.1 & +176.3\% & +286.0\% \\
\bottomrule
\end{tabular}
}
\end{sc}\end{table*}

The main results are summarized in \cref{table:complexityeval} which includes pass@10 scores for program synthesis and complexity generation scores. On top of it, we present in \cref{table:complexityeval_more_samples} more details on performance when the generation budget allows for multiple attempts (note that for \textsc{o1-mini}, due to limited inference budget, we could not retrieve the scores at higher $@k$ values). For \textbf{Program Synthesis}, we include $pass@\{1, 5, 10\}$ scores, along with $best@\{1, 5, 10\}$ and $all@\{1, 5, 10\}$ scores for \textbf{Complexity Generation} and \textbf{Complexity Ranking}.

When pushing all@k scores to k=10, the new best scores are obtained by Qwen3 32B on time complexity generation with 26.1 all@10 (also best model for all@1 with 6.5) and Llama 3.1 Nemotron-Ultra 253B on space complexity generation with 23.7 all@10 (also best model for all@1 with 5.6).

\begin{figure*}[h]

\begin{center}
\centerline{\includegraphics[width=\textwidth]{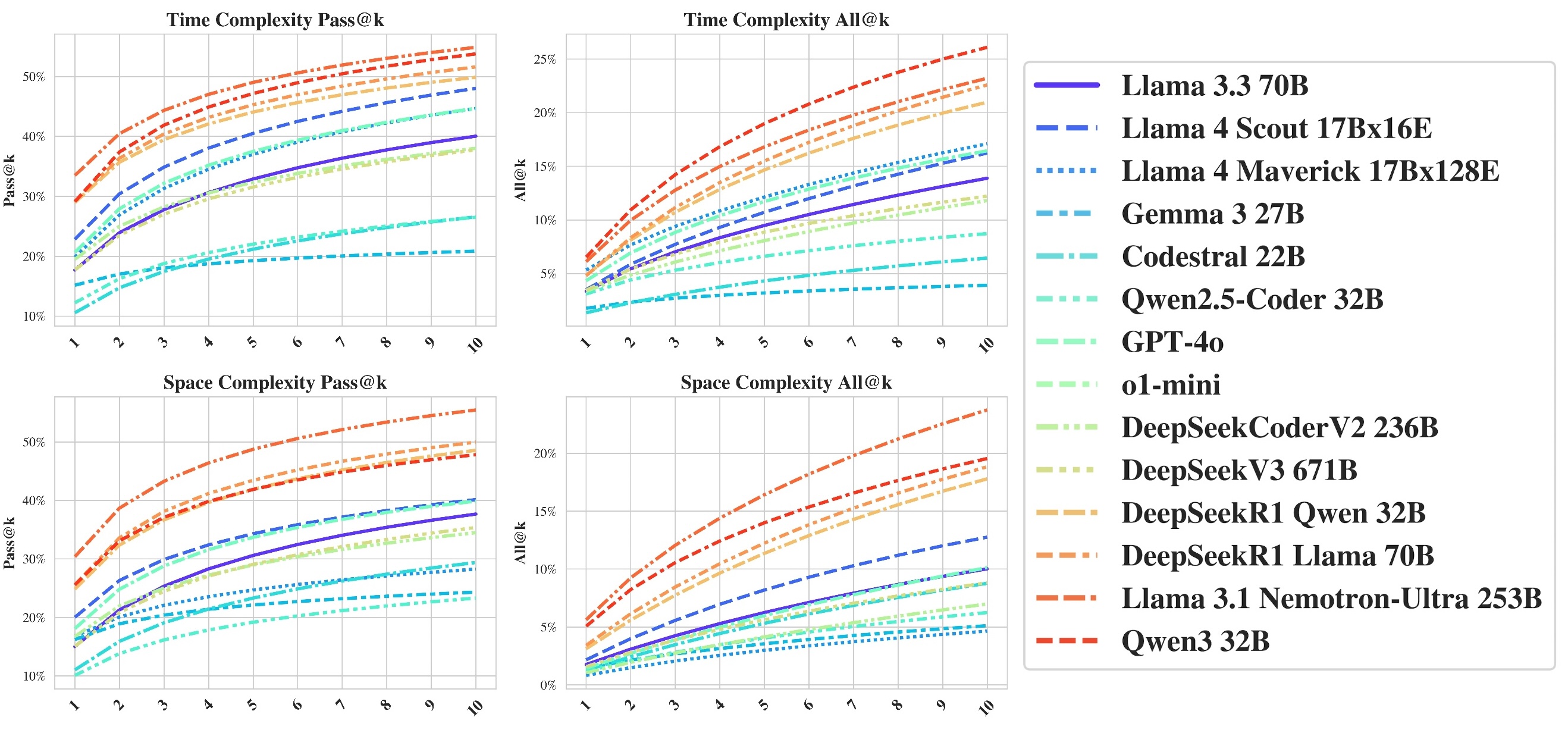}}
\caption{
Comparing gains of $Pass@k$ and $All@k$ across all models on the task of \textbf{Complexity Generation}.
}
\label{pass_at_experiments_figure}
\end{center}
\end{figure*}

In general, the order of models does not change much when pushing @k scores to higher values, except for Gemma 3, which shows less performance return. This is especially visible in \cref{pass_at_experiments_figure}, which provides the score evolution for all models across 1-step increments of k on time and space complexity tasks. Reasoning models benefit more from higher @k generations than non-reasoning models, especially on time complexity generation. Across all models, we notice higher gains for higher @k metrics when using all@k, and on space complexity generation.

Across all models, we notice higher gains for higher @k metrics when using all@k (compared to pass@k), and on space complexity generation (compared to time complexity generation). On average, models get +100\% performance for time pass@1 to pass@10, +275\% time all@1 to all@10, and on space complexity generation respectively +110\% and up to +450\%. In general, the more challenging the tasks, the higher the gains for higher @k values.

Finally, for time complexity generation, we observe more marginal gains the further we improve @k values on all@k scores, compared to pass@k scores. For example, on average across models, pass@5 is 70\% higher than pass@1, whereas pass@10 is only 15\% higher than pass@5. In comparison, all@5 is a staggering 160\% higher than all@1, and all@10 remains 40\% higher than all@5. Looking at the difference between @10 and @9, all@k still grows two times faster than pass@k.

These results give more extensive details on the strategies to adopt to use the compute budget where it matters the most in the context of complexity related tasks. Though some tasks remain very challenging at the single-sample level (all models on time-space complexity generation keep all@1 score below 10), even just doubling the sampling budget already leads to substantial performance gains.

\vfill

\newpage

\section{Fine-Tuning Experiments More Results}

In this section, we provide more extensive results about the SFT experiments. These tables decompose the compact fine-tuning results from the main paper by showing how each variant behaves when the sampling budget increases. \cref{table:complexityeval_sft_more_results_time} displays multiple @k values for the \textsc{Llama 3.1 70B} model queried for zero-shot, few-shot, and then further trained with either \textbf{Time Prediction Fine-tuning}, \textbf{Space Prediction Fine-tuning}, \textbf{Time Generation Fine-tuning}, or \textbf{Space Generation Fine-tuning}, evaluated for the \textbf{Time Complexity} tasks. Similarly, \cref{table:complexityeval_sft_more_results_space} displays the same methods but evaluated on \textbf{Space Complexity} tasks.

\cref{table:complexityeval_more_samples_relative_sft_time} provides relative scores for @5 and @10 on \textbf{Time Complexity}, and \cref{table:complexityeval_more_samples_relative_sft_space} for \textbf{Space Complexity}.

\begin{figure*}[h]

\begin{center}
\centerline{\includegraphics[width=\textwidth]{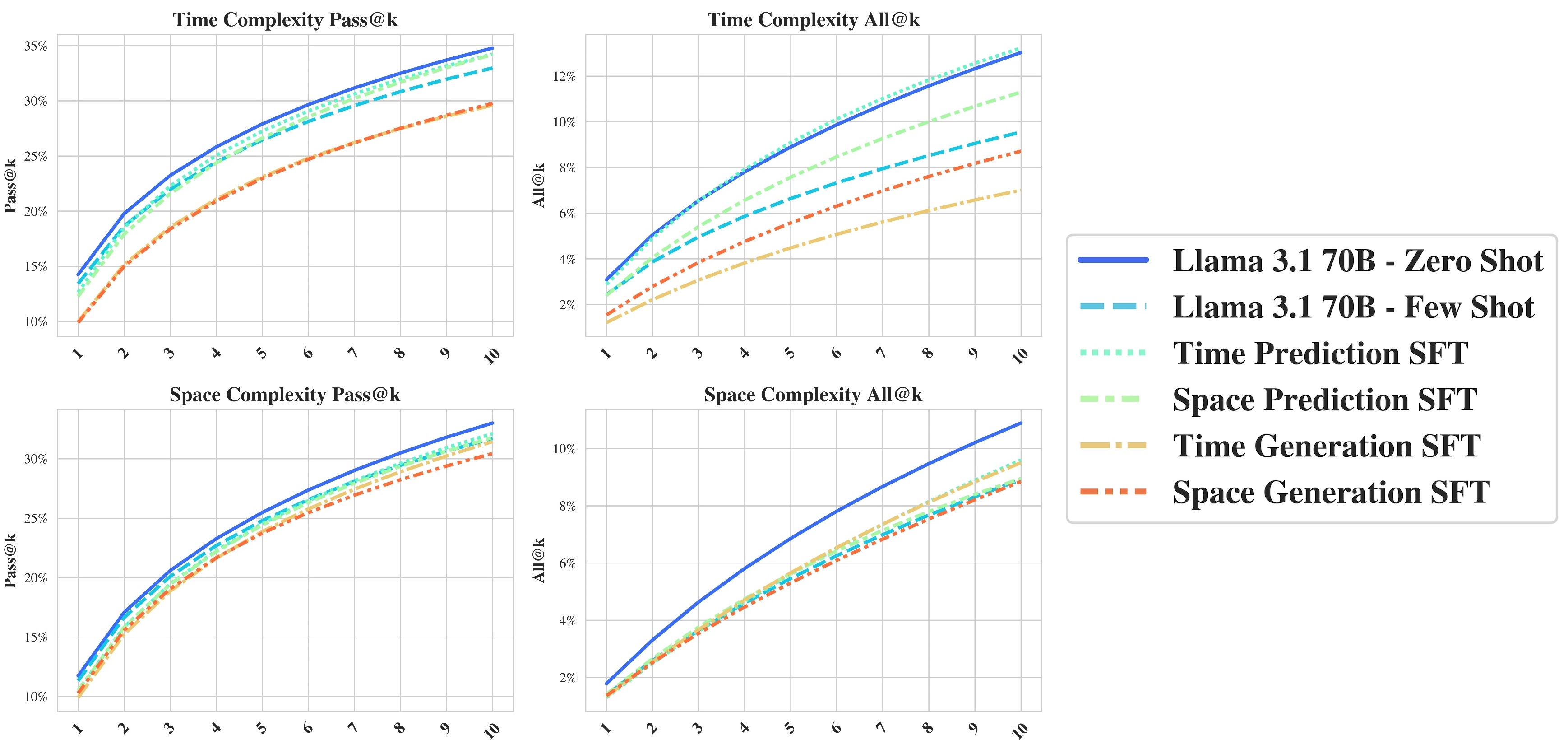}}
\caption{
Comparing gains of $Pass@k$ and $All@k$ across \textsc{Llama 3.1 70B} variants on the task of \textbf{Complexity Generation}.
}
\label{pass_at_experiments_figure_sft}
\end{center}
\end{figure*}

Finally, \cref{pass_at_experiments_figure_sft} displays visually the gain on the task of \textbf{Complexity Generation}.

\begin{table*}[h]
\centering\small
\caption{
Time Complexity \benchmark benchmark results when fine-tuning \textsc{Llama 3.1 70B} on the tasks of \textbf{Time Complexity Prediction}, \textbf{Space Complexity Prediction}, \textbf{Time Complexity Generation} and \textbf{Space Complexity Generation}, then evaluated on the tasks of \textbf{Program Synthesis} and \textbf{Time Complexity Generation}, with multiple $@k$ estimators. \textbf{Program Synthesis} checks the correctness of model-generated solutions to given programming problems, not taking into account any complexity requirement. \textbf{Complexity Generation} evaluates whether a model can output a working code snippet to a given problem that meets a time-space complexity requirement. $Pass@k$ considers each complexity class of all problems independently and calculates a macro-average between them. $Best@k$ is a refinement of $Pass@k$, focusing only on the most optimized complexity class for each problem. $All@k$ checks if all complexity classes for each problem are correctly predicted or generated simultaneously, then macro-averages across all problems.
}
\label{table:complexityeval_sft_more_results_time}
\begin{sc}
\resizebox{\linewidth}{!}{
\begin{tabular}{llcccllllllllllll}
\toprule
\multirow{3}{*}{\textbf{Method}} & ~ & \multicolumn{3}{c}{\textbf{Prog. Synthesis}} & ~ & \multicolumn{11}{c}{\textbf{Time Complexity Gen.}} \\
 & & \multicolumn{1}{c}{Pass} & \multicolumn{1}{c}{Pass} & \multicolumn{1}{c}{Pass} & & \multicolumn{1}{c}{Pass} & \multicolumn{1}{c}{Pass} & \multicolumn{1}{c}{Pass} & & \multicolumn{1}{c}{Best} & \multicolumn{1}{c}{Best}  & \multicolumn{1}{c}{Best} & & \multicolumn{1}{c}{All} & \multicolumn{1}{c}{All} & \multicolumn{1}{c}{All} \\
 & & \multicolumn{1}{c}{\bf $@1$} & \multicolumn{1}{c}{\bf $@5$} & \multicolumn{1}{c}{\bf $@10$} & & \multicolumn{1}{c}{\bf $@1$} & \multicolumn{1}{c}{\bf $@5$} &\multicolumn{1}{c}{\bf $@10$} & & \multicolumn{1}{c}{\bf $@1$} & \multicolumn{1}{c}{\bf $@5$} & \multicolumn{1}{c}{\bf $@10$} & & \multicolumn{1}{c}{\bf $@1$} & \multicolumn{1}{c}{\bf $@5$} & \multicolumn{1}{c}{\bf $@10$} \\
\hline\vspace{-0.1in} \\
~~Zero-shot & &  29.6 & 47.2 & 54.0  & &  14.2 & 27.9 & 34.8 & & 20.3 & 35.0 & 41.2 & & 3.1 & 8.9 & 13.0 \\
~~Few-shot & & 28.9 & 45.9 & 52.8  & &  13.4 & 26.4 & 33.0 & & 19.6 & 35.0 & 41.8 & & 2.4 & 6.6 & 9.6 \\
\textbf{Prediction Fine-tuning} & & \\
~~On Time & &  27.4 & 46.4 & 53.6  & &  12.6 & 27.2 & 34.2 & & 17.2 & 32.6 & 39.0 & & 2.9 & 9.1 & 13.2 \\
~~On Space & &  26.6 & 46.4 & 54.3  & &  12.3 & 26.6 & 34.2 & & 17.3 & 33.5 & 40.9 & & 2.4 & 7.6 & 11.3 \\
\textbf{Generation Fine-tuning} & & \\
~~On Time & &  23.2 & 42.0 & 48.5  & &  10.0 & 23.1 & 29.6 & & 13.3 & 27.6 & 33.6 & & 1.2 & 4.5 & 7.0 \\
~~On Space & &  23.4 & 42.7 & 50.0  & &  9.9 & 23.0 & 29.8 & & 13.1 & 27.8 & 34.9 & & 1.5 & 5.6 & 8.7 \\
\bottomrule
\end{tabular}
}

\end{sc}\end{table*}

\begin{table*}[h]
\centering\small
\caption{
Time Complexity \benchmark benchmark results when fine-tuning \textsc{Llama 3.1 70B} on the tasks of \textbf{Time Complexity Prediction}, \textbf{Space Complexity Prediction}, \textbf{Time Complexity Generation} and \textbf{Space Complexity Generation}, then evaluated on the tasks of \textbf{Time Complexity Generation}, with multiple $@k$ estimators, where $@5$ and $@10$ are displayed relative to the corresponding $@1$ score. \textbf{Complexity Generation} evaluates whether a model can output a working code snippet to a given problem that meets a time-space complexity requirement. $Pass@k$ considers each complexity class of all problems independently and calculates a macro-average between them. $All@k$ checks if all complexity classes for each problem are correctly predicted or generated simultaneously, then macro-averages across all problems.
}
\label{table:complexityeval_more_samples_relative_sft_time}
\begin{sc}
\resizebox{0.8\linewidth}{!}{
\begin{tabular}{lllrrllrr}
\toprule
\multirow{3}{*}{\textbf{Model}} & ~ & \multicolumn{7}{c}{\textbf{Time Complexity Gen.}} \\
 & & \multicolumn{1}{c}{Pass} & \multicolumn{1}{c}{Pass} & \multicolumn{1}{c}{Pass} & & \multicolumn{1}{c}{All} & \multicolumn{1}{c}{All} & \multicolumn{1}{c}{All} \\
 & & \multicolumn{1}{c}{\bf $@1$} & \multicolumn{1}{c}{\bf $@5$} & \multicolumn{1}{c}{\bf $@10$} & & \multicolumn{1}{c}{\bf $@1$} & \multicolumn{1}{c}{\bf $@5$} & \multicolumn{1}{c}{\bf $@10$} \\
\hline\vspace{-0.1in} \\
~~Zero-shot & & 14.2 & +95.9\% & +144.2\% & & 3.1 & +188.1\% & +322.1\% \\
~~Few-shot & & 13.4 & +97.0\% & +145.6\% & & 2.4 & +173.4\% & +293.4\% \\
\textbf{Prediction Fine-tuning} & & \\
~~On Time & & 12.6 & +115.6\% & +170.8\% & & 2.9 & +215.8\% & +359.6\% \\
~~On Space & & 12.3 & +116.9\% & +178.6\% & & 2.4 & +216.0\% & +371.6\% \\
\textbf{Generation Fine-tuning} & & \\
~~On Time & & 10.0 & +131.4\% & +196.3\% & & 1.2 & +271.3\% & +481.0\% \\
~~On Space & & 9.9 & +132.1\% & +200.8\% & & 1.5 & +260.8\% & +464.4\% \\
\bottomrule
\end{tabular}
}
\end{sc}\end{table*}

\begin{table*}[h]
\centering\small
\caption{
Space Complexity \benchmark benchmark results when fine-tuning \textsc{Llama 3.1 70B} on the tasks of \textbf{Time Complexity Prediction}, \textbf{Space Complexity Prediction}, \textbf{Time Complexity Generation} and \textbf{Space Complexity Generation}, then evaluated on the tasks of \textbf{Program Synthesis} and \textbf{Space Complexity Generation}, with multiple $@k$ estimators. \textbf{Program Synthesis} checks the correctness of model-generated solutions to given programming problems, not taking into account any complexity requirement. \textbf{Complexity Generation} evaluates whether a model can output a working code snippet to a given problem that meets a time-space complexity requirement. $Pass@k$ considers each complexity class of all problems independently and calculates a macro-average between them. $Best@k$ is a refinement of $Pass@k$, focusing only on the most optimized complexity class for each problem. $All@k$ checks if all complexity classes for each problem are correctly predicted or generated simultaneously, then macro-averages across all problems.
}
\label{table:complexityeval_sft_more_results_space}
\begin{sc}
\resizebox{\linewidth}{!}{
\begin{tabular}{llcccllllllllllll}
\toprule
\multirow{3}{*}{\textbf{Method}} & ~ & \multicolumn{3}{c}{\textbf{Prog. Synthesis}} & ~ & \multicolumn{11}{c}{\textbf{Space Complexity Gen.}} \\
 & & \multicolumn{1}{c}{Pass} & \multicolumn{1}{c}{Pass} & \multicolumn{1}{c}{Pass} & & \multicolumn{1}{c}{Pass} & \multicolumn{1}{c}{Pass} & \multicolumn{1}{c}{Pass} & & \multicolumn{1}{c}{Best} & \multicolumn{1}{c}{Best}  & \multicolumn{1}{c}{Best} & & \multicolumn{1}{c}{All} & \multicolumn{1}{c}{All} & \multicolumn{1}{c}{All} \\
 & & \multicolumn{1}{c}{\bf $@1$} & \multicolumn{1}{c}{\bf $@5$} & \multicolumn{1}{c}{\bf $@10$} & & \multicolumn{1}{c}{\bf $@1$} & \multicolumn{1}{c}{\bf $@5$} &\multicolumn{1}{c}{\bf $@10$} & & \multicolumn{1}{c}{\bf $@1$} & \multicolumn{1}{c}{\bf $@5$} & \multicolumn{1}{c}{\bf $@10$} & & \multicolumn{1}{c}{\bf $@1$} & \multicolumn{1}{c}{\bf $@5$} & \multicolumn{1}{c}{\bf $@10$} \\
\hline\vspace{-0.1in} \\
~~Zero-shot & &  29.7 & 45.7 & 52.3  & &  11.7 & 25.5 & 33.0 & & 17.2 & 32.4 & 39.1 & & 1.8 & 6.9 & 10.9\\
~~Few-shot & &  29.8 & 46.5 & 53.3  & &  11.3 & 24.8 & 31.7 & & 16.4 & 31.6 & 38.6 & & 1.4 & 5.5 & 8.9\\
\textbf{Prediction Fine-tuning} & & \\
~~On Time & &  27.3 & 44.7 & 51.4  & &  10.5 & 24.5 & 32.1 & & 15.0 & 30.2 & 36.8 & & 1.3 & 5.6 & 9.6 \\
~~On Space & &  27.0 & 45.0 & 51.9  & &  10.5 & 24.4 & 31.8 & & 15.2 & 30.9 & 38.1 & & 1.4 & 5.6 & 9.0 \\
\textbf{Generation Fine-tuning} & & \\
~~On Time & &   23.9 & 42.6 & 49.5  & &  9.9 & 23.9 & 31.4 & & 14.6 & 31.1 & 38.3 & & 1.3 & 5.7 & 9.5\\
~~On Space & &  24.2 & 42.1 & 48.8  & &  10.3 & 23.7 & 30.4 & & 15.0 & 31.1 & 37.6 & & 1.4 & 5.3 & 8.8 \\
\bottomrule
\end{tabular}
}

\end{sc}\end{table*}

\begin{table*}[h]
\centering\small
\caption{
Space Complexity \benchmark benchmark results when fine-tuning \textsc{Llama 3.1 70B} on the tasks of \textbf{Time Complexity Prediction}, \textbf{Space Complexity Prediction}, \textbf{Time Complexity Generation} and \textbf{Space Complexity Generation}, then evaluated on the tasks of \textbf{Space Complexity Generation}, with multiple $@k$ estimators, where $@5$ and $@10$ are displayed relative to the corresponding $@1$ score. \textbf{Complexity Generation} evaluates whether a model can output a working code snippet to a given problem that meets a time-space complexity requirement. $Pass@k$ considers each complexity class of all problems independently and calculates a macro-average between them. $All@k$ checks if all complexity classes for each problem are correctly predicted or generated simultaneously, then macro-averages across all problems.
}
\label{table:complexityeval_more_samples_relative_sft_space}
\begin{sc}
\resizebox{0.8\linewidth}{!}{
\begin{tabular}{lllrrllrr}
\toprule
\multirow{3}{*}{\textbf{Model}} & ~ & \multicolumn{7}{c}{\textbf{Space Complexity Gen.}} \\
 & & \multicolumn{1}{c}{Pass} & \multicolumn{1}{c}{Pass} & \multicolumn{1}{c}{Pass} & & \multicolumn{1}{c}{All} & \multicolumn{1}{c}{All} & \multicolumn{1}{c}{All} \\
 & & \multicolumn{1}{c}{\bf $@1$} & \multicolumn{1}{c}{\bf $@5$} & \multicolumn{1}{c}{\bf $@10$} & & \multicolumn{1}{c}{\bf $@1$} & \multicolumn{1}{c}{\bf $@5$} & \multicolumn{1}{c}{\bf $@10$} \\
\hline\vspace{-0.1in} \\
~~Zero-shot & & 11.7 & +117.3\% & +181.5\% & & 1.8 & +284.0\% & +509.9\% \\
~~Few-shot & & 11.3 & +119.3\% & +180.5\% & & 1.4 & +291.1\% & +538.4\% \\
\textbf{Prediction Fine-tuning} & & \\
~~On Time & & 10.5 & +133.8\% & +206.0\% & & 1.3 & +332.0\% & +638.9\% \\
~~On Space & & 10.5 & +133.2\% & +203.4\% & & 1.4 & +298.3\% & +533.7\% \\
\textbf{Generation Fine-tuning} & & \\
~~On Time & & 9.9 & +141.2\% & +217.7\% & & 1.3 & +324.9\% & +614.1\% \\
~~On Space & & 10.3 & +131.2\% & +196.4\% & & 1.4 & +289.1\% & +548.8\% \\
\bottomrule
\end{tabular}
}
\end{sc}\end{table*}

\clearpage

\section{Tasks Correlations}

The correlation figures in this appendix are computed on the model sweep used for this appendix analysis, which includes \textsc{Qwen QwQ 32B}. They are not recomputed with the \textsc{Qwen3 32B} rows reported in the main benchmark tables, so the values below should be read as an analysis of that sweep rather than as additional Qwen3 results.

\begin{figure*}[h]

\begin{center}
\centerline{\includegraphics[width=\textwidth]{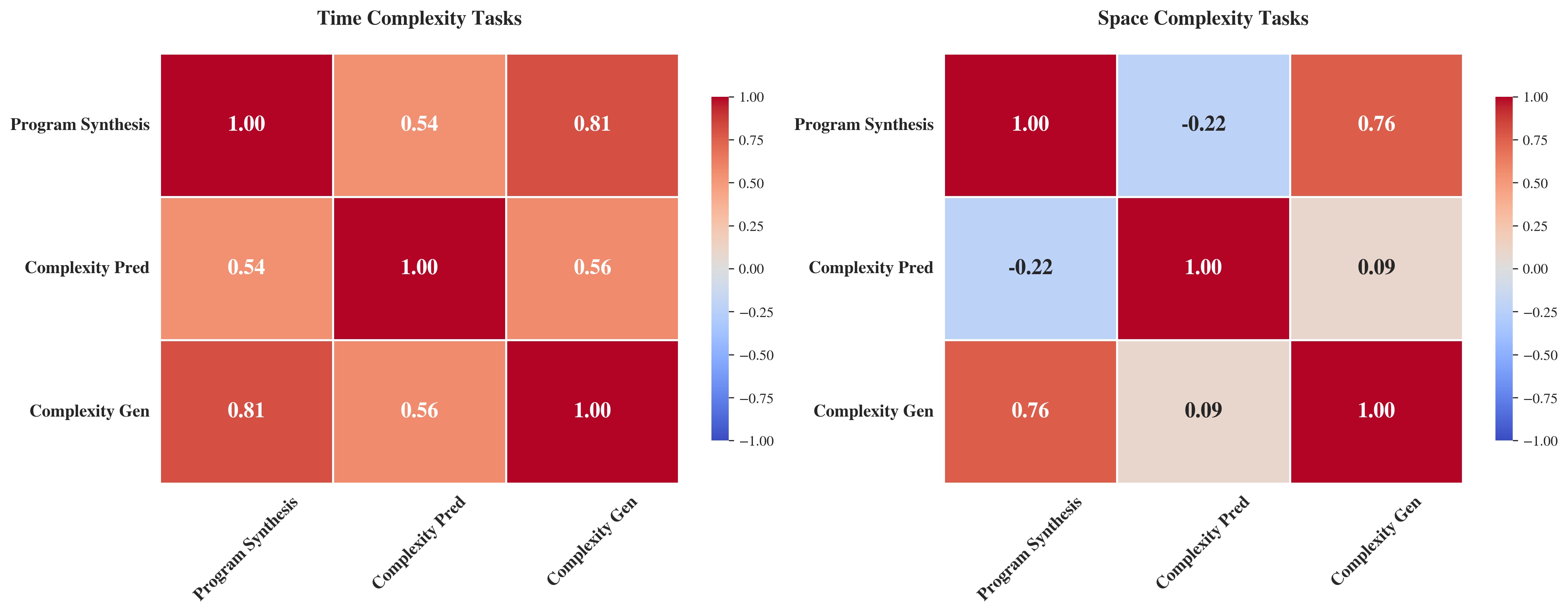}}
\caption{
Correlation matrices comparing the distribution of scores across all models in this appendix sweep for the tasks of \textbf{Program Synthesis} (on the corresponding time or space complexity test set), \textbf{Complexity Prediction} and \textbf{Complexity Generation}, the two latter being respectively on the time and space complexity test sets. Correlations are computed with the Pearson standard correlation coefficient.
}
\label{correlation_time_complexity_correlation_space_complexity}
\end{center}
\end{figure*}

\begin{figure*}[h]
\begin{center}
\centerline{\includegraphics[width=0.5\textwidth]{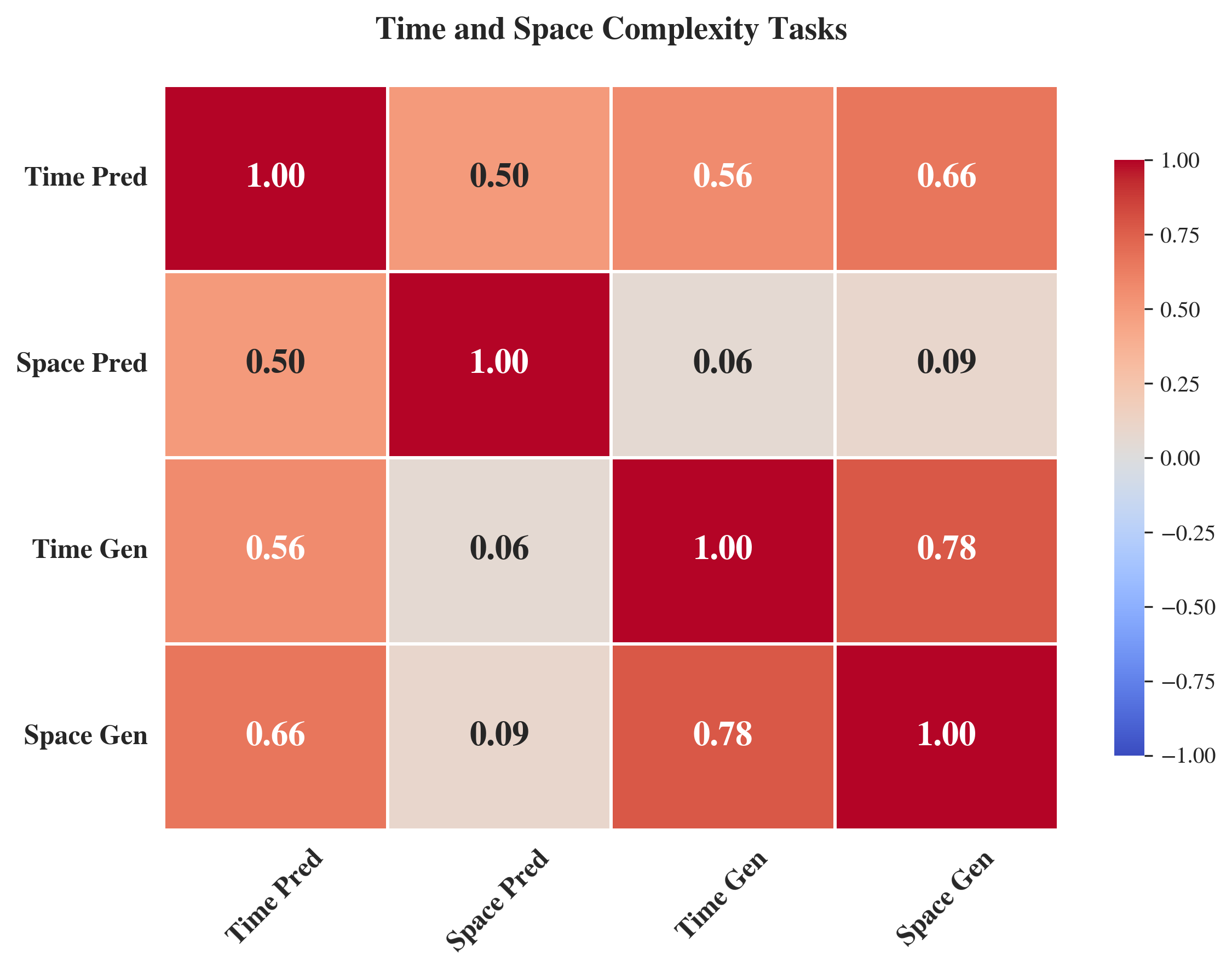}}
\caption{
Correlation matrix comparing the distribution of scores across all models in this appendix sweep for the tasks of \textbf{Time Complexity Prediction} and \textbf{Time Complexity Generation}, and \textbf{Space Complexity Prediction} and \textbf{Space Complexity Generation}. Correlations are computed with the Pearson correlation coefficient.
}
\label{correlation_time_space_complexity}
\end{center}
\end{figure*}

Using the many scores computed across models in this sweep on the different tasks of \benchmark, correlations can be measured between tasks: program synthesis, complexity prediction, and complexity generation, for both time and space. \Cref{correlation_time_complexity_correlation_space_complexity} reports the correlation matrices within time-related tasks and within space-related tasks. \Cref{correlation_time_space_complexity} then reports the correlations between time and space prediction and generation tasks. All of these measures use the Pearson correlation coefficient \citep{Pearson1895}.

Overall, the strongest correlations are between program synthesis and complexity generation: 0.81 for time complexity and 0.76 for space complexity, across all models in this appendix sweep. A likely explanation is that models are more frequently exposed to time optimization objectives when trained to generate code, since these objectives are more common on code competition platforms and more extensively documented than their space counterparts. Time prediction has a correlation coefficient of 0.56 with time complexity generation, suggesting that the program synthesis objective may still dominate this double-requirement task. For space generation, where we observe very low All@1 scores, the correlation between prediction and generation drops to 0.1. This is consistent with the low absolute performance on space generation: models often struggle with the space constraint and therefore behave more like generic program synthesis systems than like systems optimizing for a specified complexity target.

Finally, when comparing time and space tasks, we observe that time prediction correlates more with time generation than with space prediction. The converse does not hold: space prediction has a correlation coefficient of 0.5 with time prediction and only 0.1 with space generation. This again suggests that space generation remains dominated by the program synthesis objective, because the double objective is still too hard for most models. Time generation and space generation also correlate more strongly with one another than with their respective prediction counterparts.

Notice that the fine-tuning experiments presented \cref{table:finetuning_experiments} and discussed in \cref{quantitative_analysis} can also help better understand the dependencies between the tasks introduced in \benchmark.

\clearpage

\section{Where Do the Differences in Complexity Prediction and Generation Come From}
\label{app:generation_failures}

\Cref{table:complexityeval} shows a large gap between what models can predict and what they can generate, and \cref{quantitative_analysis} offers a qualitative explanation of why. Here we try to further refine that analysis with more measurements. Every generation attempt carries two independent bits: whether the program passes the tests, and whether the framework measures it in the requested complexity class. Separating them says what the failures are actually made of.

\subsection{What Generation Failures Are Made Of}
\label{app:generation_failures_decomposition}

\paragraph{Four-way decomposition.}
Crossing the two bits splits every attempt into a success and three kinds of failure: correct code in the wrong complexity class, incorrect code that happens to land in the right class (though this is limited in some sense, if the code crashes at the beginning it gets logged as $O(1)$, though from the structure of the code that's not really what the theoretical complexity label would be), and code that is wrong on both counts. We report numbers for all of that for time complexity generation in \cref{tab:failure_decomposition}.

The dominant failure mode is a working program in the wrong complexity class. Its share of all failures ranges from 14\% for \textsc{Codestral 22B} to 70\% for \textsc{Llama 3.1 Nemotron-Ultra 253B}, and it tracks program synthesis ability: the better a model is at writing correct code, the more of its complexity failures are purely complexity failures rather than coding failures, and the space axis behaves the same way, from 17\% to 68\%. Complexity generation is therefore more than a correlate of program synthesis skill: for the models strongest at the latter, the \emph{right code, wrong class} error does not diminish as fast as the purely wrong-code error. This has to be qualified, though. We cannot conclude that the best models at code synthesis are no better than the worst ones at complexity control, since the weakest models only face the complexity requirement on the problems they can solve at all, which are the easiest ones, whereas the strongest models face it on harder problems as well, precisely because they are able to write working solutions for them.

\begin{table}[!ht]
\centering\small
\caption{Four-way decomposition of generation attempts on time complexity, as a percentage of all attempts, with models ordered by the final column. Working code is the share of that model's failures which consist of a correct program in the wrong complexity class. The remaining failure mass is code that fails the tests, with or without matching the requested class.}
\label{tab:failure_decomposition}
\begin{tabular}{lrrrrr}
\toprule
\textbf{Model} & \textbf{success} & \textbf{right code,} & \textbf{wrong code,} & \textbf{both} & \textbf{working code} \\
 & & \textbf{wrong class} & \textbf{right class} & \textbf{wrong} & \textbf{share of failures} \\
\midrule
Llama 3.1 Nemotron-Ultra 253B & 33.2 & 46.8 & 6.2 & 13.8 & 70.0\% \\
DeepSeekR1 Llama 70B & 28.8 & 41.2 & 3.9 & 26.1 & 57.9\% \\
DeepSeekR1 Qwen 32B & 28.6 & 41.3 & 4.1 & 26.0 & 57.8\% \\
Qwen3 32B & 28.9 & 41.0 & 6.7 & 23.4 & 57.6\% \\
Llama 4 Scout 17Bx16E & 22.6 & 38.8 & 13.3 & 25.3 & 50.1\% \\
Llama 4 Maverick 17Bx128E & 19.7 & 39.1 & 12.7 & 28.4 & 48.7\% \\
Llama 3.3 70B & 17.5 & 25.4 & 20.4 & 36.7 & 30.8\% \\
DeepSeekCoderV2 236B & 19.4 & 24.6 & 22.4 & 33.7 & 30.5\% \\
DeepSeekV3 671B & 17.5 & 23.4 & 22.5 & 36.7 & 28.3\% \\
Gemma 3 27B & 14.8 & 22.1 & 15.6 & 47.5 & 25.9\% \\
Qwen2.5-Coder 32B & 11.9 & 18.0 & 22.5 & 47.5 & 20.5\% \\
Codestral 22B & 10.4 & 12.9 & 29.5 & 47.3 & 14.4\% \\
\bottomrule
\end{tabular}
\end{table}

\begin{figure*}[t!]
\centering
\includegraphics[width=\textwidth]{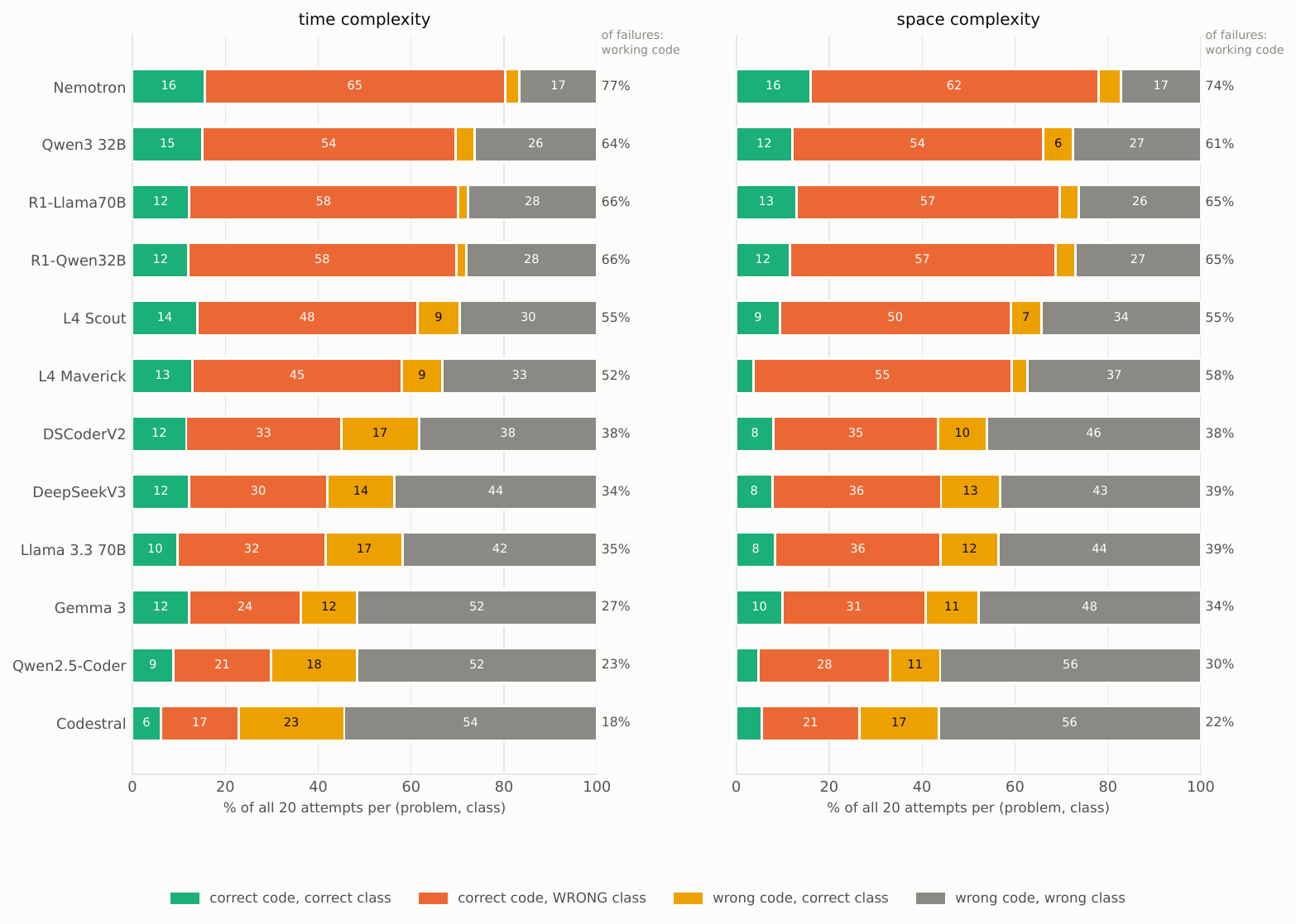}
\caption{Generation attempts on suboptimal requested classes (which means that for a given problem we know that there is a better class of complexity of solutions, but we asked for one that is less optimal), split four ways by whether the tests pass and whether the framework-measured class matches the requested one, for 12 models on both axes. On suboptimal targets, most failures are working programs in the wrong complexity class: the share of \emph{failures} that consist of correct code in the wrong class rises with program synthesis ability, from 18\% to 77\% on time complexity.}
\label{fig:failure_decomposition}
\end{figure*}

\paragraph{Optimal versus suboptimal targets.}
\Cref{quantitative_analysis} observes that models do worse on the non-optimal complexity classes of a problem and suggests that models can have the tendency to decline producing less-optimized solutions, and rather output the best complexity class only. We test this directly by splitting the decomposition on whether the requested class is the most optimized one for that problem, and \cref{tab:optimal_suboptimal} shows two findings on every model, no exception. Success collapses on suboptimal targets, by 36 points for \textsc{Nemotron-Ultra}, from 51.8 to 15.6, while the share of failures that are working code rises, from 58.0\% to 76.5\% for the same model. The models are not losing the ability to write correct programs when asked for a slower class; they keep writing correct programs and decline to hit the class. \Cref{fig:failure_decomposition} shows the same four-way split restricted to suboptimal requests, for all 12 models on both axes.

\begin{table}[!ht]
\centering\small
\caption{Generation on time complexity, split by whether the requested class is the most optimized one for the problem. Every model shows the same two movements: success drops sharply on suboptimal targets, while the fraction of failures that consist of working code in the wrong class rises. Models ordered by the size of the success drop.}
\label{tab:optimal_suboptimal}
\begin{tabular}{lrrrrrr}
\toprule
\multirow{2}{*}{\textbf{Model}} & \multicolumn{3}{c}{\textbf{success}} & \multicolumn{3}{c}{\textbf{working-code share of failures}} \\
\cmidrule(lr){2-4}\cmidrule(lr){5-7}
 & optimal & suboptimal & $\Delta$ & optimal & suboptimal & $\Delta$ \\
\midrule
Llama 3.1 Nemotron-Ultra 253B & 51.8 & 15.6 & $-36.2$ & 58.0\% & 76.5\% & $+18.5$ \\
DeepSeekR1 Llama 70B & 46.5 & 12.1 & $-34.4$ & 43.7\% & 66.0\% & $+22.3$ \\
DeepSeekR1 Qwen 32B & 46.1 & 12.1 & $-34.0$ & 44.6\% & 65.5\% & $+20.9$ \\
Qwen3 32B & 43.5 & 15.1 & $-28.4$ & 47.4\% & 64.1\% & $+16.7$ \\
Llama 4 Scout 17Bx16E & 31.8 & 13.9 & $-17.9$ & 43.4\% & 55.2\% & $+11.8$ \\
DeepSeekCoderV2 236B & 27.6 & 11.6 & $-16.1$ & 21.1\% & 37.7\% & $+16.7$ \\
Llama 3.3 70B & 25.7 & 9.8 & $-15.9$ & 25.0\% & 35.3\% & $+10.3$ \\
Llama 4 Maverick 17Bx128E & 27.0 & 12.9 & $-14.0$ & 44.8\% & 51.8\% & $+7.0$ \\
DeepSeekV3 671B & 23.0 & 12.2 & $-10.8$ & 21.6\% & 33.8\% & $+12.2$ \\
Codestral 22B & 14.9 & 6.1 & $-8.8$ & 10.3\% & 17.9\% & $+7.6$ \\
Qwen2.5-Coder 32B & 15.2 & 8.8 & $-6.4$ & 17.6\% & 23.1\% & $+5.5$ \\
Gemma 3 27B & 17.6 & 12.2 & $-5.4$ & 24.2\% & 27.4\% & $+3.2$ \\
\bottomrule
\end{tabular}
\end{table}

\subsection{Which Way Models Miss}
\label{app:generation_failures_overshoot}

\begin{figure*}[t!]
\centering
\includegraphics[width=\textwidth]{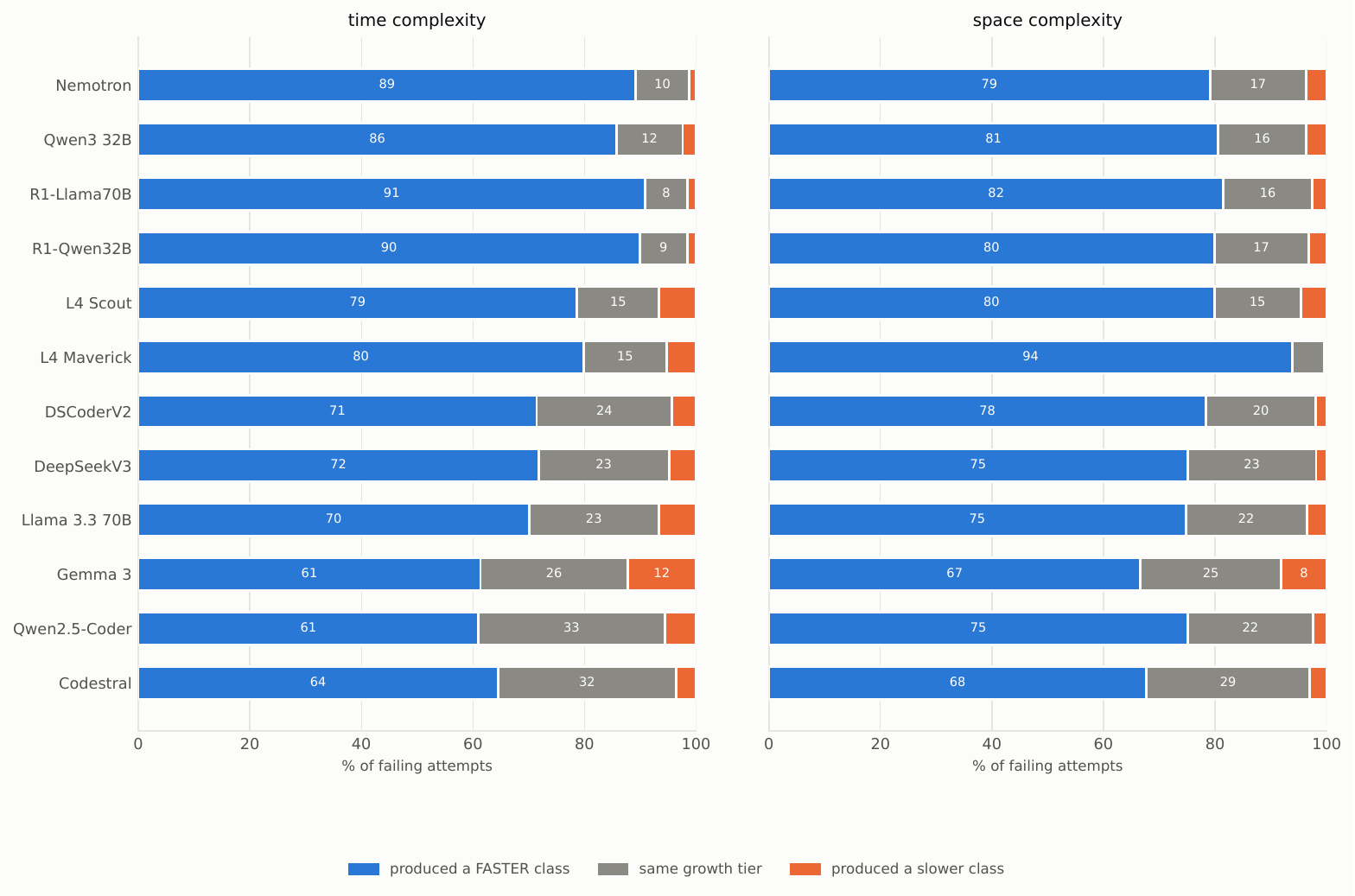}
\caption{Among failing attempts on suboptimal complexity classes for the task of complexity generation, whether the framework-measured class of the produced program is faster than, slower than, or at the same complexity level as the class that was requested but still different, for instance comparing $O(n\,\mathrm{log}(n)+m)$ with $O(n+m\,\mathrm{log}(m))$. Asked for a suboptimal complexity and failing, models overshoot toward faster code: faster outnumbers slower by between 5:1 and over 200:1 for every model on both axes.}
\label{fig:overshoot}
\end{figure*}

Given that models fail on suboptimal targets by writing working code in the wrong class, the next question is which wrong class. As displayed in \cref{fig:overshoot}, they overshoot toward faster code. Among failing attempts on suboptimal complexity classes for the task of complexity generation, the produced class is faster than the requested one for 61\% to 94\% of attempts depending on the model, against 0.4\% to 12\% slower. The ratio ranges from 5:1 to over 200:1, and the direction is the same for every model on both axes.

Asked for a deliberately slow solution and failing, models almost never miss the complexity class by being too slow; they miss it by being too fast. \Cref{quantitative_analysis} notes that a model could satisfy a suboptimal request by adding a redundant sort or an extra loop to an optimal solution, and that they do not; we share more numbers on that observation, per model, in \cref{fig:overshoot}.

\subsection{Prediction and Generation Performance Discrepancies at the Item Level}
\label{app:generation_failures_dissociation}

\begin{figure*}[t!]
\centering
\includegraphics[width=\textwidth]{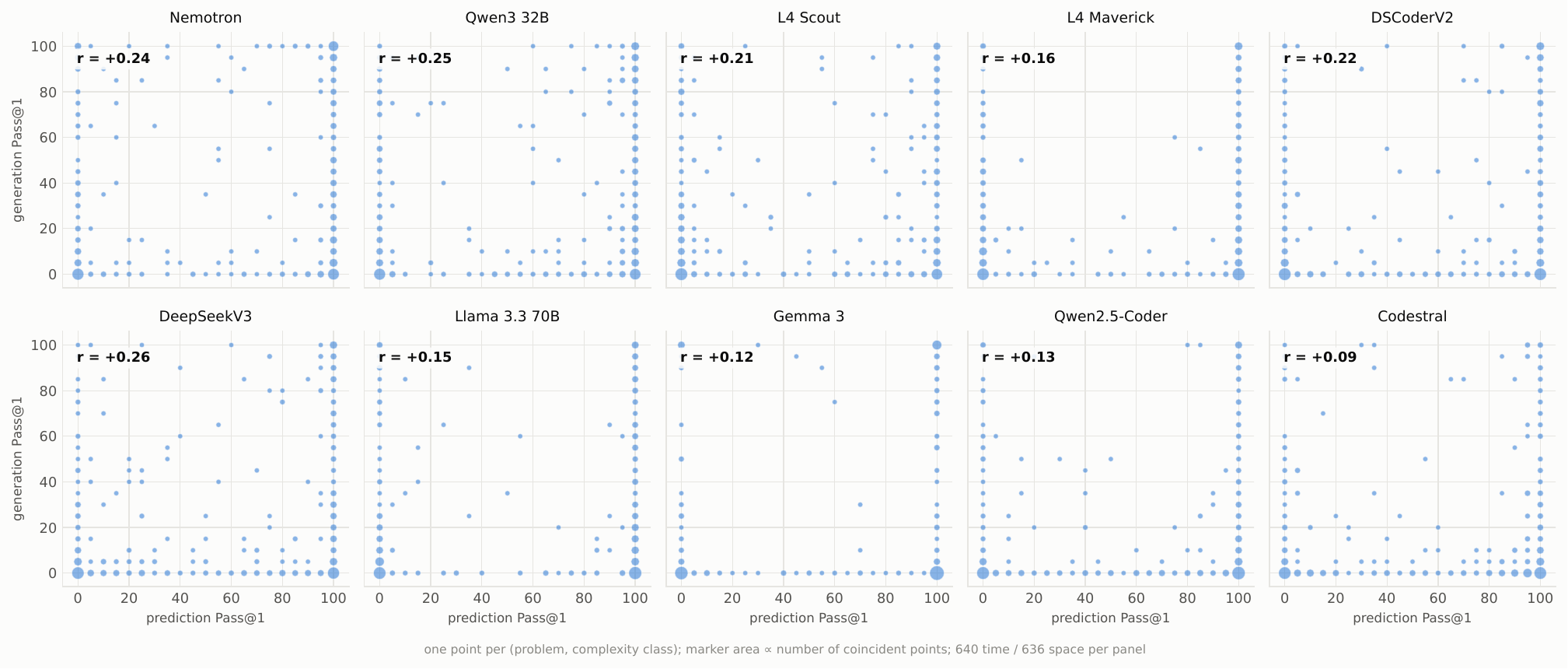}
\caption{Per-(problem, class) complexity prediction $Pass@1$ against complexity generation $Pass@1$ on time complexity, 640 points per panel; marker size is proportional to the number of points on that spot. Prediction concentrates at 0 and 100 while generation spreads across the range, and the two are nearly uncorrelated within every model: prediction and generation are close to independent at the level of a single problem.}
\label{fig:pred_vs_gen_time}
\end{figure*}

\begin{figure*}[t!]
\centering
\includegraphics[width=\textwidth]{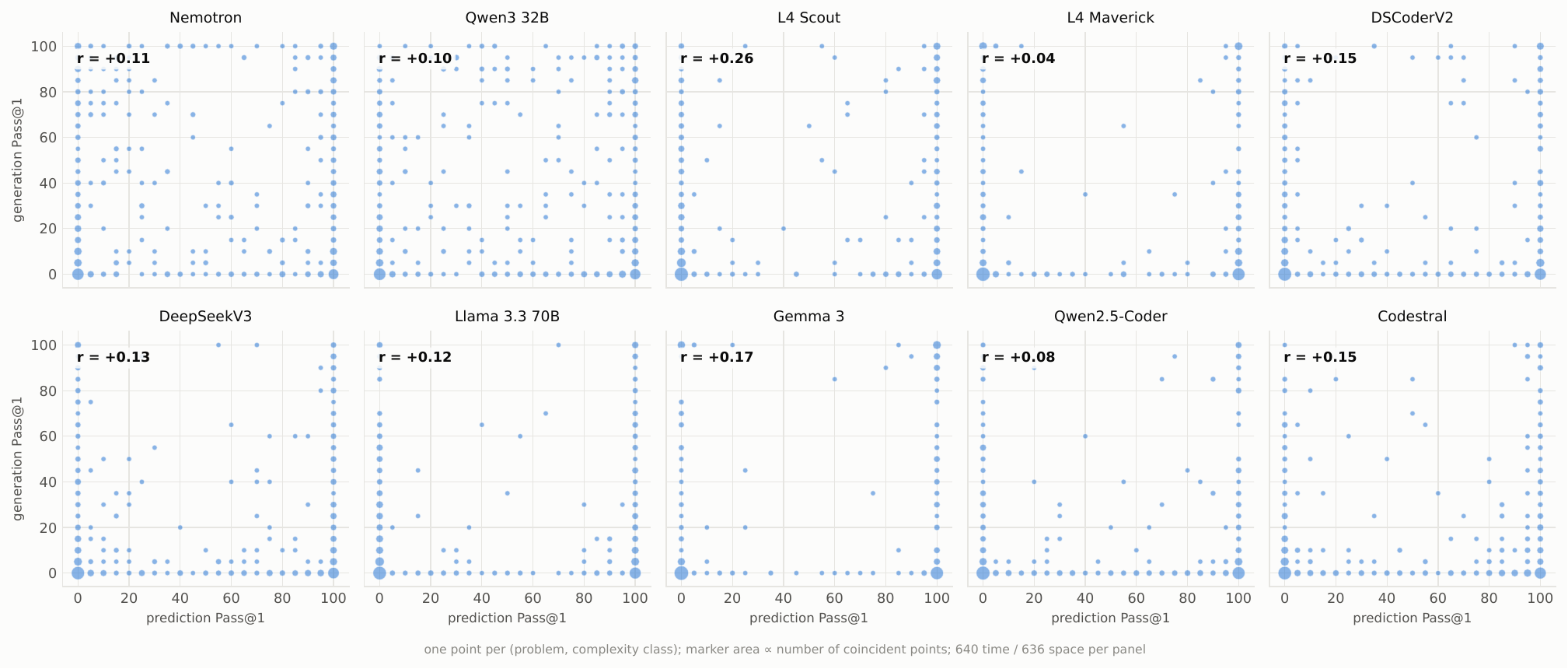}
\caption{The same comparison on space complexity, with figure explanations as in \cref{fig:pred_vs_gen_time} and 636 points per panel.}
\label{fig:pred_vs_gen_space}
\end{figure*}

The correlations of \cref{correlation_time_complexity_correlation_space_complexity} compare aggregate scores across models. We ask a more refined version of the same question: within a single model, are the individual problems it can predict the ones it can also generate for? \Cref{fig:pred_vs_gen_time,fig:pred_vs_gen_space} plot per-(problem, class) complexity prediction $Pass@1$ against complexity generation $Pass@1$, giving 640 points per model on time and 636 on space, and \cref{tab:pred_vs_gen} reports the correlations.

All 20 correlations fall between $+0.04$ and $+0.26$: at the level of an individual problem and complexity class, knowing that a model can name the complexity says almost nothing about whether it can write code to meet it. The shape of the scatter says the same thing: prediction mass piles up at 0 and at 100, since a model's twenty predictions on an item are most of the time self-consistent and agree, while generation spreads across the full range. Prediction and generation are close to separate skills, and the aggregate correlation across models understates how separate they are per item.

\begin{table}[!ht]
\centering\small
\caption{Pearson correlation between per-(problem, class) complexity prediction $Pass@1$ and complexity generation $Pass@1$, within each model. 640 points per model on time, 636 on space. Every value is small and positive: being able to name the complexity of a solution on a problem does correlate a bit with the capability of generating a solution of the right complexity from scratch, but not much.}
\label{tab:pred_vs_gen}
\begin{tabular}{lrr}
\toprule
\textbf{Model} & \textbf{time} & \textbf{space} \\
\midrule
Codestral 22B & $+0.086$ & $+0.153$ \\
DeepSeekV3 671B & $+0.259$ & $+0.127$ \\
DeepSeekCoderV2 236B & $+0.217$ & $+0.149$ \\
Gemma 3 27B & $+0.120$ & $+0.172$ \\
Llama 3.3 70B & $+0.150$ & $+0.119$ \\
Llama 4 Maverick 17Bx128E & $+0.163$ & $+0.043$ \\
Llama 4 Scout 17Bx16E & $+0.214$ & $+0.264$ \\
Llama 3.1 Nemotron-Ultra 253B & $+0.243$ & $+0.111$ \\
Qwen2.5-Coder 32B & $+0.130$ & $+0.082$ \\
Qwen3 32B & $+0.245$ & $+0.101$ \\
\bottomrule
\end{tabular}
\end{table}

\clearpage

\titleformat*{\paragraph}{\itshape}

\section{Framework Implementation Details}
\label{implementation_details}

\subsection{General Framework Outline}

\ifshowtext
The time-space complexity framework is a rule-based algorithm that can process any Python function in order to infer its time and space complexities dynamically. The high-level principles of the framework are presented in \cref{framework_complexity}, explaining how it takes a code snippet and an input example and tries to infer a time and space complexity from it.

As inputs, it takes a Python function, its function inputs, and their corresponding dataclass, which are then processed and modified before runtime and memory footprints are measured. From a high-level perspective, the framework increases the size of inputs following various strategies, in order to assess the impact of input size on execution metrics (e.g. execution time, memory used). When the function has several arguments, they can be expanded independently or together to determine the overall complexity of the function, taking into account potential interdependencies. The prepared code, along with the various sets of expanded inputs, is queued and run in independent sandboxes, using the \textit{Bubblewrap} library \citep{bubblewrap}, to avoid any harmful side effects of the code being run. While running, \textit{Cprofiler} \citep{cpython} is used for time execution measures and \textit{tracemalloc} for memory footprint. Using non-negative least squares curve fitting on each set of measures, the coefficients and residuals of each complexity class are computed. The selected complexity class for a given set of measures is chosen as the minimizer of the residuals, taking into account a simplicity bias (the simpler the complexity class is, the smaller the simplicity bias). This curve fitting is applied to each set of measures, each corresponding to a different subset of arguments being expanded with a different expansion method. Using ensemble methods, the global complexity of the Python function is computed by aggregating the individual complexity outputs across the different sets of measures. Finally, the complexity framework also returns the coefficients of the curve of each selected complexity. These coefficients can be leveraged to rank and classify the optimizations of different Python solutions within the same complexity class.

\subsection{Complexity Framework Inputs}

The complexity framework handles two types of input code on which to measure time and space complexity. The datasets released with this paper, as well as all the results detailed in our paper, follow format 1, but the framework can also handle a second type of data for call-based code snippets.

\paragraph{Input Format 1 - With a dataclass}
This format corresponds to the case where snippets of input code are I/O based, such as in the following examples:

\begin{lstlisting}
#A. Array
n = int(input())
a,b,c =[],[],[]
l = list(map(int,input().split()))
for i in l:
    if i<0:
        a.append(i)
    elif i>0:
        b.append(i)
    else:
        c.append(i)

if len(b)==0 and len(a)>2:
    b.append(a.pop())
    b.append(a.pop())
if len(a)%2==0:
    c.append(a.pop())
print(len(a),*a)
print(len(b),*b)
print(len(c),*c)
\end{lstlisting}

In which case the corresponding input example, as given to the complexity framework, will be formatted in the following manner:

\begin{lstlisting}
    '4\n-1 -2 -3 0\n'
\end{lstlisting}

This is the case where the input, whatever the number of distinct arguments there really is, is concatenated as a single string. In this case, the framework could not alone guess where the arguments are, without context, especially as it is not relying on any LLM. The framework, in order to run, needs an external dataclass that specifies how to understand the input string and cut it into different arguments, that the framework can then try to change to understand the time and space dependencies upon each of them.

Using an LLM, we can first infer the dataclass corresponding to a particular code challenge, before using it as part of the complexity framework in order to parse the input example and perform the various measurements on the variations of the inputs. Such a dataclass will have the following format:

\begin{lstlisting}
@dataclass
class Input:
    n: int
    a_list: List[int]

    @classmethod
    def from_str(cls, input_: str):
        n, a_list, _ = input_.split('\n')
        n = int(n)
        a_list = list(map(int, a_list.split()))
        assert n == len(a_list)
        return cls(n, a_list)

    def __repr__(self):
        return str(self.n) + '\n' + ' '.join(map(str, self.a_list)) + '\n'
\end{lstlisting}

\paragraph{Input Format 2 - Standalone code snippet}

This second format corresponds to the case where the input code is call-based, that is to say a particular function is being executed on a particular set of inputs, in which case the framework does have the information of how the inputs are separated into different arguments to the function (and that will also parametrize the final time and space complexity). Such call-based code generally takes the following form:

\begin{lstlisting}
class Array_300_A:
def solve(self, n, l):
    #A. Array
    a,b,c =[],[],[]
    for i in l:
        if i<0:
            a.append(i)
        elif i>0:
            b.append(i)
        else:
            c.append(i)

    if len(b)==0 and len(a)>2:
        b.append(a.pop())
        b.append(a.pop())
    if len(a)%2==0:
        c.append(a.pop())
    print(len(a),*a)
    print(len(b),*b)
    print(len(c),*c)
\end{lstlisting}

It is accompanied by inputs of the form:

\begin{lstlisting}
{"n": "4", "l": "[-1, -2, -3, 0]"}
\end{lstlisting}

In this case, the framework can directly understand the structure of the input example, and based on that infer the complexity of the code snippet.

\subsection{Fuzzing and Worst Running Cases}

\benchmark focuses on worst-case time and space complexities, therefore any LLM-generated code solution is evaluated to infer its tightest upper-bound time and space complexities. To do so, any code solution input to the framework comes with a corresponding edge case input, as shared in the corresponding released dataset. Using fuzzing, this edge case input is derived into many inputs of different sizes following different generators: for example, an input pair consisting of an integer and a list can see the integer input grow independently from the list (the integer can grow linearly, with random size steps, etc, whereas the list can remain static) or on the contrary interdependently (both the integer and the list grow at a regular pace, with same size steps). In the released codebase, the input-generation module is in charge of handling the fuzzing.
Then, the sandbox and the time and memory profilers record the behavior of the code solution on all sets of inputs, a set of inputs being parametrized by a specific input generator suggesting inputs over a range of sizes. This enables to gather time execution and memory footprint curves over many input cases, and worst execution curves are used to derive the form of the associated complexity using curve fitting methods in the curve-fitting module.

Fuzzing is in itself a whole area of research, with many publications covering Java or C++ programs \citep{Noller_2018,fuzzing_review}. More recently, LLM capabilities are more and more studied as a way of making fuzzing techniques more exhaustive \citep{Xia_2024,llm_fuzz}. In our benchmark, we chose to design an evaluation framework completely independent from any machine learning model, so as to avoid biasing the evaluation. Nevertheless, the modularity of the implementation enables switching the default fuzzer so to experiment with any other fuzzer released by the computer science community, or more recently from LLM-based approaches. This would enable to measure what kind of bias can arise, and whether coverage and accuracy of the complexity evaluation are improved. Finally, without changing the fuzzer, the input edge cases example can also be interchanged easily as it is an input of the framework fuzzer: one can experiment with prompting an LLM for more edge case inputs.

Our paper provides methods to evaluate the quality of the fuzzing as performed by our evaluation framework, using ground-truth human solutions to all BigOBench benchmark problems. As detailed in \cref{complexity_framework_performance}, we conducted a thorough review to evaluate whether our evaluation framework correctly identifies worst-case scenarios and accurately derive complexity estimates from them, using a total of 250 samples, including 125 for time complexity and 125 for space. On the time complexity test set, the framework reaches 84\% accuracy, with 1000-bootstrap samples confidence interval [0.776, 0.904] (for space, 82\% accuracy [75.2, 88.8]). In addition, these test sets include, for time complexity, 42\% problems of difficulty A (for space 45\%), 29\% of B (for space 25\%) and 30\% of C+ (for space 30\%). On hard problems C+, framework accuracy is 84\% for time complexity (84\% for space). Being exposed to stochastic noise when measuring runtimes and memory footprints, the evaluation framework also got evaluated for its consistency, running the framework 20 times on 10 solutions of every problem and complexity class of the candidate test set: self-consistency of the framework is measured to be at 91.9\% (resp. 89.1\%) for time (resp. space) complexity, for a total of 10,130 (resp. 10,520) different code solutions.

\cref{implementation_details} provides an illustration and more details concerning the methods being used in the framework in order to measure worst-case complexities.

\subsection{Variance Reduction Techniques}

Variance reduction techniques are important as the complexity framework relies on empirical measures of runtime and memory footprint, which are subject to noise. The framework was tested and ablated on a validation set in order to measure the effect of each design choice. Among them, each measure of runtime and memory footprint is repeated R times (in the current setting, R = 10 after ablation, trading-off accuracy with added compute cost) and then aggregated following a variance reduction technique. It turned out that the min-aggregation provided the best results for runtime measurements, improving complexity framework by 3.5\% over median aggregation and 19\% over max aggregation, among others. Our intuition is that runtimes are typically subject to variability caused by external factors (such as background processes, CPU throttling, system load etc.) in a way that increases runtime, creating high outliers.

The Hodges-Lehmann estimator \citep{Hodges1963,qiu2025efficientllmgeneratedcoderigorous} for example has not been specifically tested yet for the complexity framework. The codebase was designed with modularity as a core principle and any estimator can be added to the curve-fitting module as a function of a set of empirical measures.

Beyond the variance-reduction techniques that can be used on empirical noisy measures, \benchmark also employs similar technique when aggregating scores over multi-samples. \citet{qiu2025efficientllmgeneratedcoderigorous} employ a Rao-Blackwellized bootstrap estimator, while we use an unbiased estimator for @k measures by leveraging n = 20 samples for k $<=$ 10 with c $<=$ n correct samples:

$$
\text{pass}@k = \mathbb{E}_p \left[ 1 - \frac{\dbinom{n - c}{k}}{\dbinom{n}{k}} \right]
$$

This estimator, introduced by \citet{chen2021evaluating}, is an unbiased measure of the multi-sample performance as demonstrated in the appendix A of this paper.

In general, given that as measured in \cref{complexity_framework_performance} we measure the accuracy of the framework with human annotations at $\sim$85\%, we estimated that this success rate ensured the framework was accurate enough to provide reliable conclusions on the performance of LLMs as part of the \benchmark.

\subsection{Complexity Framework Parametrization}

The complexity framework is parameterized by a suite of arguments that govern its execution, categorized into several domains. Dataset-related arguments specify the input data, including the required \texttt{path\_to\_jsonl\_file} and optional \texttt{sub\_key} for nested dictionaries, alongside indices (\texttt{code\_start\_index}, \texttt{code\_end\_index}) and filters (\texttt{filter\_on\_problem}, \texttt{multiply\_samples\_factor}) to control code selection and replication. Input handling is configured via \texttt{input\_handler}, while logging and output behaviors (\texttt{log\_outputs}, \texttt{save\_results}, \texttt{skip\_saving\_full\_results}, \texttt{results\_folder\_name\_root}) dictate result persistence. Measurement parameters (\texttt{shuffle\_runs}, \texttt{correct\_nlogn}, \texttt{multiplier\_op}, \texttt{multiplier\_start}, \texttt{multiplier\_repeat}, \texttt{multiplier\_end}, \texttt{multiplier\_mult\_step}, \texttt{multiplier\_max\_increase}, \texttt{size\_of\_other\_arguments}, \texttt{time\_profiler}) regulate input scaling and timing. Complexity fitting is modulated by outlier removal (\texttt{filter\_outliers}), penalty/constraint application (\texttt{apply\_penalty}, \texttt{apply\_constraints}), and aggregation strategies (\texttt{aggregate\_y\_values}, \texttt{max\_time\_rate}, \texttt{elect\_complexity\_time}, \texttt{elect\_complexity\_space}, \texttt{fix\_constant\_complexity}, \texttt{fix\_negligeable\_complexity}). Resource management includes memory/time thresholds (\texttt{temp\_file\_name\_seed}, \texttt{memory\_limit}, \texttt{timeout}, \texttt{large\_timeout}, \texttt{giga\_timeout}, \texttt{global\_timeout}) and CPU allocation policies (\texttt{max\_workers}, \texttt{number\_solutions\_per\_worker}, \texttt{main\_process\_cpu\_id\_list}, \texttt{forkserver\_type}, \texttt{use\_distinct\_forkservers}, \texttt{forkserver\_cpu\_id\_list}). Batch-scheduling arguments further tailor high-performance computing deployment.

\subsection{Sensitivity to Complexity Framework Parameters}
\label{app:parameter_sensitivity}

The parameters listed above were settled during development, so it is fair to ask how much the released labels depend on that particular choice. We re-ran the fitting stage over the 640 solutions of the time complexity test set, changing one parameter at a time and leaving everything else at its post-development settled value. The runtime measurements are reused across configurations, and the parameters that we vary are the ones that transform the measured values into the assigned complexity label. \Cref{fig:parameter_sensitivity,tab:parameter_sensitivity} report, for each setting, the fraction of the 640 released labels it reproduces.

One limit that we underline about such a study: agreement with the released labels is maximized at the production configuration by construction, so the table measures how much a change costs (we are not concluding in this section about whether a different configuration would be better, but underlining the parameters that are the most critical in our setting; during the development stage, we used a validation set to measure gains, and the post-development configuration is the best one we found across the parameter values we explored).

\begin{figure*}[t!]
\centering
\includegraphics[width=0.85\textwidth]{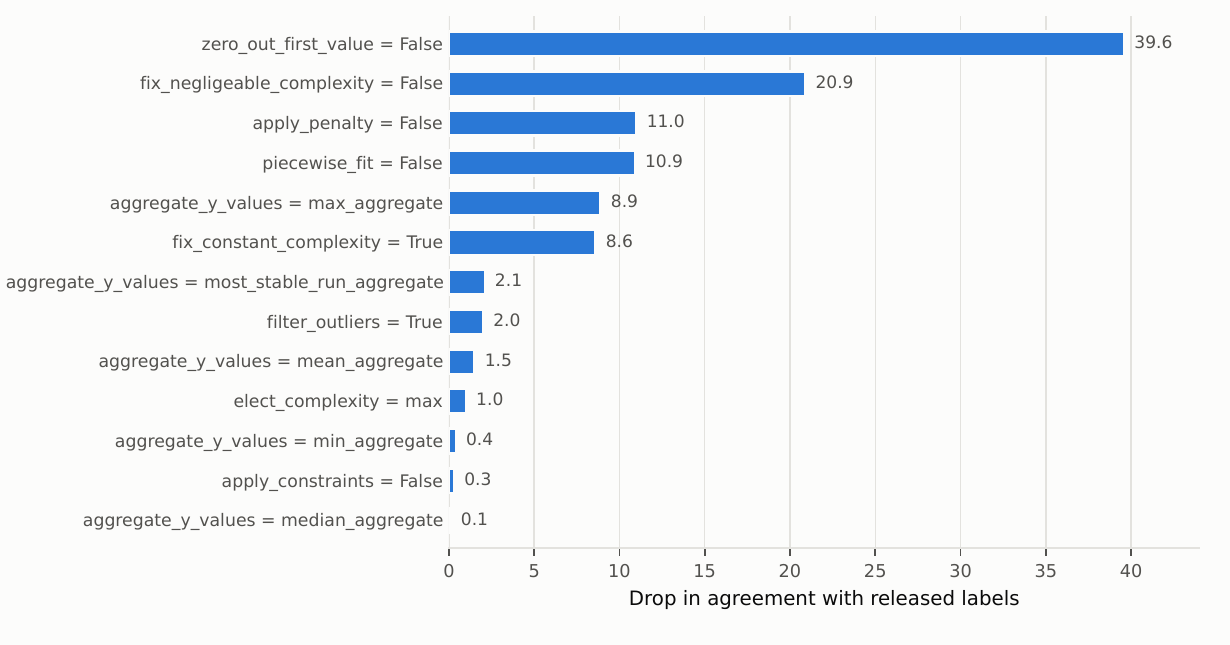}
\caption{Drop in agreement with the released time complexity labels when one parameter is moved away from its post-development settled value, over the 640 test solutions. Two parameters account for most of the sensitivity: \texttt{zero\_out\_first\_value} and \texttt{fix\_negligeable\_complexity} move 39.6 and 20.9 points of labels respectively, nothing else exceeds 11 points, and six of the thirteen settings sit within about 2 points of the production configuration.}
\label{fig:parameter_sensitivity}
\end{figure*}

\begin{table}[!ht]
\centering\small
\caption{Sensitivity of the released time complexity labels to the fitting-stage parameters, one parameter changed at a time from the best elected configuration, over the 640 test solutions. Agreement is the fraction of released labels the variant reproduces; $\Delta$ is the drop against the baseline. Differences below about 0.5 points are within run-to-run nondeterminism.}
\label{tab:parameter_sensitivity}
\begin{tabular}{llrr}
\toprule
\textbf{Parameter} & \textbf{Setting} & \textbf{Agreement} & \textbf{$\Delta$ (pp)} \\
\midrule
\multicolumn{2}{l}{\emph{production configuration}} & 99.8\% & --- \\
\midrule
\texttt{zero\_out\_first\_value} & \texttt{False} & 60.2\% & $-39.6$ \\
\texttt{fix\_negligeable\_complexity} & \texttt{False} & 78.9\% & $-20.9$ \\
\texttt{apply\_penalty} & \texttt{False} & 88.8\% & $-11.0$ \\
\texttt{piecewise\_fit} & \texttt{False} & 88.9\% & $-10.9$ \\
\texttt{aggregate\_y\_values} & \texttt{max\_aggregate} & 90.9\% & $-8.9$ \\
\texttt{fix\_constant\_complexity} & \texttt{True} & 91.2\% & $-8.6$ \\
\texttt{aggregate\_y\_values} & \texttt{most\_stable\_run\_aggregate} & 97.7\% & $-2.1$ \\
\texttt{filter\_outliers} & \texttt{True} & 97.8\% & $-2.0$ \\
\texttt{aggregate\_y\_values} & \texttt{mean\_aggregate} & 98.3\% & $-1.5$ \\
\texttt{elect\_complexity} & \texttt{max} & 98.8\% & $-1.0$ \\
\texttt{aggregate\_y\_values} & \texttt{min\_aggregate} & 99.4\% & $-0.4$ \\
\texttt{apply\_constraints} & \texttt{False} & 99.5\% & $-0.3$ \\
\texttt{aggregate\_y\_values} & \texttt{median\_aggregate} & 99.7\% & $-0.1$ \\
\bottomrule
\end{tabular}
\end{table}

Two parameters, that consist in turning on or off a feature of the curve fitting, do dominate the impact on the predicted complexities. Zeroing the first measured point, which removes the interpreter warm-up sample that otherwise anchors the low end of every curve, moves 39.6 points of labels on its own. Collapsing terms whose fitted contribution is negligible over the measured range, which is what turns a fit like $n^2 + n$ (but with $n$ having a very small coefficient compared to the one of $n^2$) into $O(n^2)$, moves 20.9. Concerning the other parameters, a middle tier of 8.6 to 11.0 points covers the simplicity penalty, the piecewise fit, forcing a constant class when the curve is flat, and max-aggregation. The remaining seven settings sit within 2.1 points, and \texttt{apply\_constraints} at 0.3 is inside the noise floor (though during development we noticed that the addition of all these little tweaks does have a big impact on the final performance).

Aggregation deserves a caveat. Max-aggregation is clearly the worst of the five, at $-8.9$, which matches the fact that runtime noise only ever inflates a measurement, so taking the maximum over replicas accumulates noise rather than cancelling it. The production aggregator tries the minimum first and only falls back when the minimum-aggregated curve is not monotone. There is not much conclusion to draw from any gap below 0.5 points, which we estimate to be the order of magnitude of the fitting noise.

\subsection{Compute Budget Versus Label Quality}
\label{app:cost_quality}

Running the framework is the dominant CPU cost of an evaluation, and three parameters control almost all of it: the number of replicas $R$ of each measurement, the largest input multiplier the framework scales up to, and the set of input generator functions. The production configuration uses $R = 10$ and a multiplier of up to $3072\times$ the original input size. We swept the $5 \times 6$ grid over $R \in \{1, 2, 3, 5, 10\}$ and multiplier caps in $\{32, 128, 512, 1024, 2048, 3072\}$, holding the generators fixed, again on the 640 time complexity test solutions, and measured both the agreement with the released labels and the cost. Cost is the sum of the retained execution times, so it is measured rather than estimated; it excludes sandbox and framework overhead, which makes the reported speedups optimistic but not too far off, as this overhead remains limited.

\begin{figure*}[t!]
\centering
\includegraphics[width=0.85\textwidth]{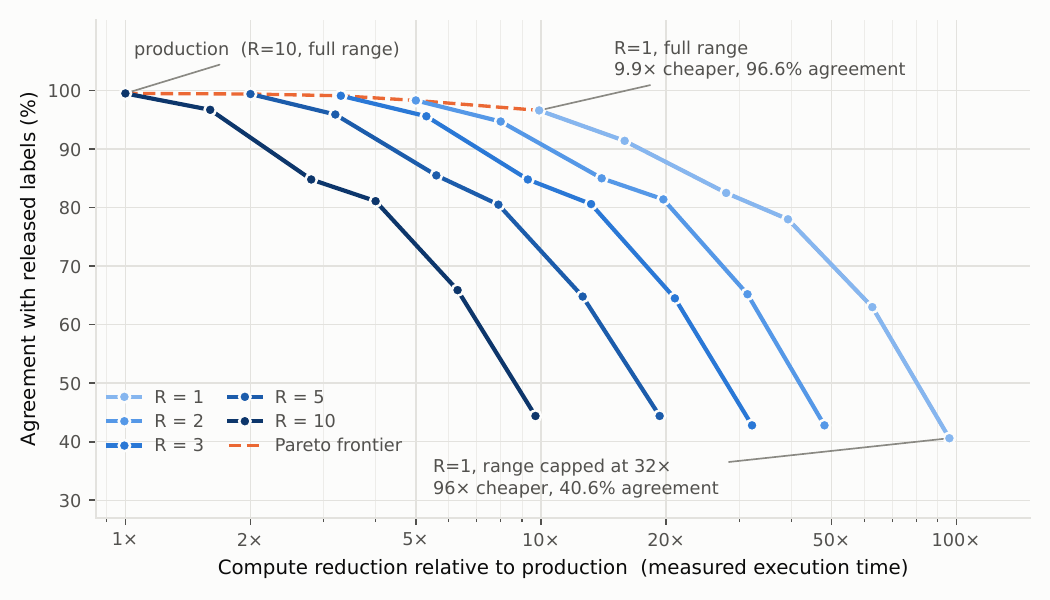}
\caption{Agreement with the released time complexity labels against compute reduction, for the $5 \times 6$ grid of replica counts and multiplier caps, on a log-scaled x-axis. Cutting replicas is cheap while cutting the input range is not: every point on the Pareto frontier keeps the full input range and spends its savings on replicas, and at essentially the same cost, $R = 1$ at full range scores 96.6\% while $R = 10$ with the range capped at $32\times$ scores 44.4\%.}
\label{fig:cost_quality}
\end{figure*}

\begin{table}[!ht]
\centering\small
\caption{Selected points from the $5 \times 6$ grid. Agreement is the fraction of the 640 released time complexity labels reproduced. Accuracy $\geq$ is a lower bound, obtained by assuming that every label the variant changes is wrong: a configuration agreeing on a fraction $g$ of labels has accuracy at least $84 - (100 - g)$ against human annotation, using the 84\% reported in \cref{complexity_framework_performance}.}
\label{tab:cost_quality}
\begin{tabular}{lrrr}
\toprule
\textbf{Configuration} & \textbf{Agreement} & \textbf{Speedup} & \textbf{Accuracy $\geq$} \\
\midrule
$R = 10$, full range (production) & 99.5\% & $1.0\times$ & 83.5\% \\
$R = 3$, full range & 99.1\% & $3.3\times$ & 83.1\% \\
$R = 2$, full range & 98.3\% & $5.0\times$ & 82.3\% \\
$R = 1$, full range & 96.6\% & $9.9\times$ & 80.6\% \\
$R = 1$, range capped at $2048\times$ & 91.4\% & $15.9\times$ & 75.4\% \\
$R = 1$, range capped at $1024\times$ & 82.5\% & $27.9\times$ & 66.5\% \\
$R = 1$, range capped at $32\times$ & 40.6\% & $96.1\times$ & 24.6\% \\
\bottomrule
\end{tabular}
\end{table}

The two parameters are not interchangeable, and \cref{fig:cost_quality} makes the asymmetry visible: every configuration on the Pareto frontier keeps the full input range. Concretely, $R = 1$ at full range and $R = 10$ capped at $32\times$ cost the same to within 2\%, and differ by 52 points of agreement. The mechanism is that cost concentrates in the largest input sizes, and truncating $3072 \times$ down to $2048\times$ alone removes 37\% of the total execution time, but those are exactly the sizes that separate one complexity class from the next. Cutting replicas removes redundancy, which the aggregation step largely absorbs; cutting the range removes the discriminative part of the curve. Nevertheless, one subtlety is not caught by this experimental setup. The distribution of complexity labels is highly skewed towards a few classes that are quite stable to predict on most types of algorithms. On the problems for which the curve fitting is highly prone to noise, or on classes made of several variables that each bring their own share of measurement noise, replicas turn out to be very helpful to increase the accuracy of the framework. But looking at the global picture of aggregated accuracy alone, we can indeed argue that a lot of compute can be saved for a few points of accuracy.

In practice, a tenfold saving is available at little cost. Running with a single replica over the full input range reproduces 96.6\% of the released labels for a tenth of the compute, which puts its accuracy against human annotation at no worse than 80.6\%, against the currently elected best configuration's 84\%. Capping the range at $32\times$ is 96 times cheaper and keeps only 40.6\% of the labels, which we would not consider a usable setting. \Cref{tab:cost_quality} gives the intermediate points.

\subsection{Implementation Limitations and Future Directions}

The Dynamic Complexity Inference Framework can be improved in many ways. The current version is more of a proof-of-concept, trying to see whether such framework can be used reliably on this task, for evaluation purposes of reinforcement learning. The following suggestions on how to improve the performance, reliability and maintainability of the framework are listed:

\begin{enumerate}
    \item Refactoring the whole framework: the goal being to allow for flexible extensions. Some parts of the codebase, for instance around the input generations, are already designed to allow new methods of input generation.\\

    \item Fuzzing: The modularity of the codebase presents opportunities for future research directions. One potential area of investigation is the exploration of alternative fuzzers, including those leveraging large language models (LLMs), to measure the introduction of bias and evaluate their impact on coverage and accuracy of complexity evaluation. Additionally, researchers could investigate the effectiveness of using LLMs to generate input edge cases, potentially leading to improved coverage and accuracy. Furthermore, the framework's modularity enables the study of the impact of different fuzzers on the codebase, allowing for a deeper understanding of the relationships between fuzzing strategies, input edge cases, and complexity evaluation outcomes. By experimenting with different fuzzers and input edge cases, researchers can gain insights into the strengths and limitations of various approaches, ultimately contributing to the development of more robust and effective fuzzing techniques.\\

    \item Noise reduction: several methods can help with noise reduction, and therefore better accuracy of the framework as well as more stable results. Deterministic CPU operations is a huge axis of improvement, but post-processing methods on the runtime and memory footprint measures is also promising. Variance reduction techniques such as Hodges-Lehmann estimator could also be used.\\

    \item Complexity definition: work on the definition of complexity and therefore how complexity is induced by the framework. Current assumptions may not be accurate, and maybe some choices in the framework implementation and not coherent with more widely spread definitions of complexity of a code snippet.\\

    \item Complexity coverage: The way we designed the framework is that it defines a set of base functions for curve fitting, and the framework can combine these functions with addition and multiplication operators to best fit the time execution and memory footprint curves.

    As long as such a function is defined, it is fit by the complexity framework. In practice, currently there is a class for the cubic function, and no classes that represent $x \to \frac{x^3}{\log x}$, $x \to (\log x)^6$ nor $x \to 2^{\text{polylog}(x)}$

    In practice, accuracy limitations may arise when two classes exhibit similar behavior over typical input sizes (1-10,000). Empirical noise from profiler measurements (runtime or memory footprint) can make fitting complexities challenging. We recommend testing the framework on labeled examples when introducing a new function class.\\

    \item Pure performance: the performance of the complexity framework is limited. Measurement of these limitations and their improvement is the priority goal.\\
\end{enumerate}

\newpage

\section{Benchmark Experimental Setup}
\label{benchmark_experimental_setup}

The experiments conducted in this paper were done on a cluster of H100 GPUs. Running the \benchmark evaluation on low-compute models that do not produce reasoning tokens, e.g. the \textsc{Llama 4} models, with $@10$ metrics for the three tasks of complexity prediction, generation and ranking, both on time and space complexity test sets, required ~100 GPU hours. On the contrary, more compute-intensive workflows involving reasoning models, including the \textsc{Qwen QwQ} runs used for the appendix ablations and correlation analysis, required up to ~2500 GPU hours.

Using the solutions generated by an LLM, the \benchmark evaluation framework can be run on a set of $n$ CPUs in order to perform the runtime and memory footprint measures, based on which a time-space complexity estimate is attributed to the code solution. For each LLM being benchmarked on \benchmark, hundreds of CPU hours can be required in total to run the framework on the time/space generation/ranking tasks, though this can highly vary depending on its parametrization: range of the measures, replication rate of each measure, \ldots In our most compute-intensive setting (up to $3000\times$ the original input size and 10 measure replicas), we used up to 10k CPU hours.

Finally, as part of our fine-tuning experiments, we used for each of the four fine-tunings of \textsc{Llama 3.1 70B}, that is to say on time prediction, space prediction, time generation and space generation data, ~120 GPU hours. These models were then evaluated based on the low-compute set-up, and finally the generated solutions were evaluated using compute for the complexity framework as detailed above.

Preliminary experiments for the complexity framework, that only require CPUs, used more compute than for the final runs used to evaluate the models on \benchmark. This is because as detailed in \cref{parametrization} ablations were conducted to find the optimal set of parameters for the framework, given the resources we had access to; we insist on the fact that the framework can be run with much lower compute by adjusting various parameters such as its range of measures or its replication rate. On the contrary, for the GPU compute, which comes from running the inference on a variety of models being benchmarked by \benchmark, the majority of it corresponds directly to the experiments being reported in this paper.

In total, the experiments reported in this paper used ~12,000 GPU hours for model fine-tuning and inference. In order to run the evaluation framework on all tasks of all models, the experiments used ~180,000 CPU hours, though a substantially cheaper parametrization is available: as measured in \cref{app:cost_quality}, dropping to a single measurement replica over the full input range costs a factor of ten in compute while still reproducing 96.6\% of the released time complexity labels. Larger reductions come at a real cost in label quality, and we would not recommend going much beyond that.\\

Concerning the results of these experiments being reported in the paper, we clarify that only \textsc{DeepSeek-R1} distilled models are reported, and that \textsc{DeepSeek-R1} is not reported as an initial assessment led to over-budget compute usage; this partial run gave similar results on complexity tasks as \textsc{DeepSeek-R1 Llama 70B}.

The table results that correspond to the \benchmark benchmarking of models are supported by one-tailed paired t-tests on 1000 bootstrap samples of the model results, used to evaluate whether the best model is superior to each comparison model. The application of one-tailed paired t-tests on bootstrap samples assumes that the differences in model performance metrics (e.g., pass@1 scores) across the benchmark tasks are approximately normally distributed, an assumption that is bolstered by the Central Limit Theorem given the large number of bootstrap iterations (1000). \cref{t_test_figure} provides an overview of the distribution of differences in model performance metrics. Additionally, the tests presume that the paired nature of the data, where each model is evaluated on the same set of tasks, is preserved through resampling, ensuring that dependencies between model outputs are accounted for.

\begin{figure*}[ht]

\begin{center}
\centerline{\includegraphics[width=0.9\textwidth]{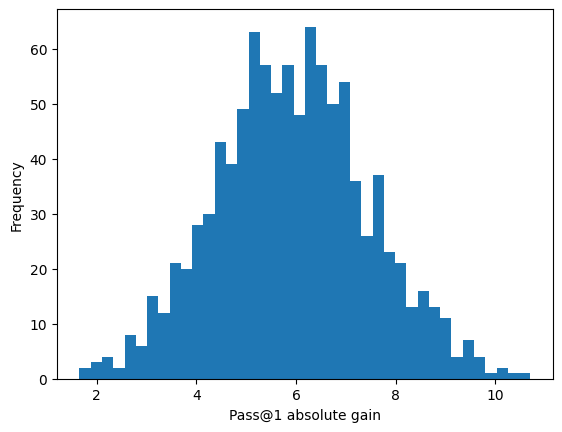}}
\caption{
$Pass@1$ absolute gain of \textsc{Deepseek R1 Llama 70B} over \textsc{Llama 3.3 70B} for the task of time complexity prediction, over 1000 bootstrap samples of \benchmark time complexity test set.
}
\label{t_test_figure}
\end{center}
\end{figure*}

Any $@k$ metric uses an unbiased estimator based on 20 samples.

\clearpage

\section{Licenses}
\label{licence_section}

\begin{table}[h]
\centering
\caption{Licenses of code repositories, datasets, and models used in the paper.}
\label{tab:licenses}
\begin{tabular}{@{}ll@{}}
\toprule
\textbf{Code Repository/Dataset} & \textbf{License} \\
\midrule
CodeContests & Apache 2.0 \href{https://www.apache.org/licenses/LICENSE-2.0}{[link]} \\
\href{https://github.com/pberkes/big_O}{big\_O} & BSD-3 \href{https://opensource.org/licenses/BSD-3-Clause}{[link]} \\
llama 4 models & \href{https://github.com/meta-llama/llama-models/blob/main/models/llama4/LICENSE}{Llama 4 License} \\
llama 3.3 & \href{https://github.com/meta-llama/llama-models/blob/main/models/llama3_3/LICENSE}{Llama 3.3 License} \\
Gemma 3 27B & \href{https://ai.google.dev/gemma/terms}{Gemma License} \\
Codestral 22B & \href{https://mistral.ai/static/licenses/MNPL-0.1.md}{MNPL 0.1 License} \\
Qwen2.5-Coder 32B & Apache 2.0 \href{https://www.apache.org/licenses/LICENSE-2.0}{[link]} \\
DeepseekCoderV2 236B & \href{https://huggingface.co/deepseek-ai/DeepSeek-Coder-V2-Instruct/blob/main/LICENSE}{Deepseek License} \\
DeepseekV3 671B & \href{https://huggingface.co/deepseek-ai/DeepSeek-V3/blob/main/LICENSE-MODEL}{Deepseek License Agreement} \\
DeepseekR1 Qwen 32B & MIT License \href{https://opensource.org/licenses/MIT}{[link]} \\
DeepseekR1 Llama 70B & MIT License \href{https://opensource.org/licenses/MIT}{[link]} \\
Llama 3.1 Nemotron-Ultra 253B & \href{https://www.nvidia.com/en-us/agreements/enterprise-software/nvidia-open-model-license/}{NVIDIA Open License Agreement} \\
Qwen QwQ 32B & Apache 2.0 \href{https://www.apache.org/licenses/LICENSE-2.0}{[link]} \\
Qwen3 32B & Apache 2.0 \href{https://www.apache.org/licenses/LICENSE-2.0}{[link]} \\
\bottomrule
\end{tabular}
\end{table}

\cref{tab:licenses} lists all licenses of code and data that were used in this paper, along with licenses of models that were downloaded and run locally to be benchmarked for this paper.

\newpage

\section{Complexity Prediction Example}
\label{appendix:complexity_prediction_example}

\ifshowtext

We include a representative example of the \textbf{Complexity Prediction} task to show the exact prompt format and the kind of model answer that is parsed by the benchmark. In this example, the model is given both the programming problem and a human solution, then asked to return only the corresponding big-O time complexity.

\subsection{Example of query}

Provide the time complexity for the following competitive programming question and corresponding solution.
When analyzing the complexity of an algorithm, consider the worst-case scenario where all possible input combinations are tried, given the following conditions: 1. the inputs must adhere to the specified data types of the problem; 2. the inputs should not cause the code to crash or exit on an exception; 3. the inputs do not necessarily need to satisfy additional constraints that are potentially mentioned in the problem statement; 4. calling input() does not consume runtime nor memory, but of course any operations on top of it or afterwards will be counted towards runtime and memory footprint; 5. Anything printed gets added to the memory. You can take advantage of Python-specific optimizations provided by the underlying CPython interpreter or compiler to achieve the desired complexity, and you must account for them when analyzing the complexity.
Here is the programming question:

You are given n segments on a number line; each endpoint of every segment has integer coordinates. Some segments can degenerate to points. Segments can intersect with each other, be nested in each other or even coincide.

The intersection of a sequence of segments is such a maximal set of points (not necesserily having integer coordinates) that each point lies within every segment from the sequence. If the resulting set isn't empty, then it always forms some continuous segment. The length of the intersection is the length of the resulting segment or 0 in case the intersection is an empty set.

For example, the intersection of segments [1;5] and [3;10] is [3;5] (length 2), the intersection of segments [1;5] and [5;7] is [5;5] (length 0) and the intersection of segments [1;5] and [6;6] is an empty set (length 0).

Your task is to remove exactly one segment from the given sequence in such a way that the intersection of the remaining (n - 1) segments has the maximal possible length.

Input

The first line contains a single integer $n$ $(2 \leq n \leq 3 \times 10^5)$ \- the number of segments in the sequence.

Each of the next n lines contains two integers $l_i$ and $r_i$ $(0 \leq l_i \leq r_i \leq 10^9)$ \- the description of the i\-th segment.

Output

Print a single integer - the maximal possible length of the intersection of (n - 1) remaining segments after you remove exactly one segment from the sequence.

Examples

Input

4
1 3
2 6
0 4
3 3

Output

1

Input

5
2 6
1 3
0 4
1 20
0 4

Output

2

Input

3
4 5
1 2
9 20

Output

0

Input

2
3 10
1 5

Output

7

Note

In the first example you should remove the segment [3;3], the intersection will become [2;3] (length 1). Removing any other segment will result in the intersection [3;3] (length 0).

In the second example you should remove the segment [1;3] or segment [2;6], the intersection will become [2;4] (length 2) or [1;3] (length 2), respectively. Removing any other segment will result in the intersection [2;3] (length 1).

In the third example the intersection will become an empty set no matter the segment you remove.

In the fourth example you will get the intersection [3;10] (length 7) if you remove the segment [1;5] or the intersection [1;5] (length 4) if you remove the segment [3;10].

Here is the corresponding Python solution:

\begin{lstlisting}[
    language=Python,
    basicstyle=\small\ttfamily,
]
n = int(input())
L = []
R = []
S = []
for _ in range(n):
    a,b = [int(x) for x in input().split()]
    L.append(a)
    R.append(b)
    S.append((a,b))


L.sort(reverse = True)
R.sort()

if (L[0],R[0]) in S:
    print(max(R[1]-L[1],0))
else:
    print(max(R[0]-L[1],R[1]-L[0],0)).
\end{lstlisting}

Please ignore any constraints on the input sizes that may have been previously mentioned in the problem description. Compute the big-O complexity as if inputs can be as large as possible.
Output the big-O time complexity only, no explanation needed, no other words needed.

\subsection{Example of response from \textsc{DeepSeek-R1 Llama 70B}}

\textless \textit{\ttfamily think}\textgreater

Okay, I need to figure out the time complexity of the given Python solution for the problem. Let's go through the code step by step.

First, the code reads n, which is the number of segments. Then, for each of the next n lines, it reads two integers l\_i and r\_i, appending them to lists L and R, and storing the tuples in S.

So the initial part is O(n) time because each of the n lines is processed once.

Next, the code sorts L in descending order. Sorting a list of size n in Python using the built-in sort function is O(n log n). Similarly, R is sorted in ascending order, which is also O(n log n).

After sorting, there's a condition checking whether the tuple (L[0], R[0]) is in S. Checking membership in a list in Python is O(n) because it has to potentially scan all elements. So this check is O(n).

Then, depending on the condition, it either prints max(R[1] - L[1], 0) or max(R[0] - L[1], R[1] - L[0], 0). Both of these operations are O(1) since they involve a couple of subtractions and a max function.

So putting it all together:

- Reading the inputs: O(n)
- Sorting L and R: O(n log n) each, so total O(2n log n) which simplifies to O(n log n)
- Checking the tuple in S: O(n)
- The rest is O(1)

The dominant term here is O(n log n) from the sorting steps. The other steps are either O(n) or O(1), which are lower order terms.

Therefore, the overall time complexity is O(n log n).

\textless$\backslash$\textit{\ttfamily think}\textgreater

The time complexity is O(n log n).

O(n log n)

\fi

\newpage

\section{Complexity Generation Example}
\label{appendix:complexity_generation_example}

\ifshowtext

We include a representative example of the \textbf{Complexity Generation} task to show the distinction between solving the programming problem and satisfying the requested complexity class. The model response below is correct as code, but the framework later classifies it as faster than the requested $O(n\log n)$ class, which makes it a failure for the complexity-constrained generation metric.

\subsection{Example of query}

Provide a Python solution for the following competitive programming question: You are given n segments on a number line; each endpoint of every segment has integer coordinates. Some segments can degenerate to points. Segments can intersect with each other, be nested in each other or even coincide.

The intersection of a sequence of segments is such a maximal set of points (not necesserily having integer coordinates) that each point lies within every segment from the sequence. If the resulting set isn't empty, then it always forms some continuous segment. The length of the intersection is the length of the resulting segment or 0 in case the intersection is an empty set.

For example, the intersection of segments [1;5] and [3;10] is [3;5] (length 2), the intersection of segments [1;5] and [5;7] is [5;5] (length 0) and the intersection of segments [1;5] and [6;6] is an empty set (length 0).

Your task is to remove exactly one segment from the given sequence in such a way that the intersection of the remaining (n - 1) segments has the maximal possible length.

Input

The first line contains a single integer $n$ $(2 \leq n \leq 3 \times 10^5)$ \- the number of segments in the sequence.

Each of the next n lines contains two integers $l_i$ and $r_i$ $(0 \leq l_i \leq r_i \leq 10^9)$ \- the description of the i\-th segment.

Output

Print a single integer - the maximal possible length of the intersection of (n - 1) remaining segments after you remove exactly one segment from the sequence.

Examples

Input

4
1 3
2 6
0 4
3 3

Output

1

Input

5
2 6
1 3
0 4
1 20
0 4

Output

2

Input

3
4 5
1 2
9 20

Output

0

Input

2
3 10
1 5

Output

7

Note

In the first example you should remove the segment [3;3], the intersection will become [2;3] (length 1). Removing any other segment will result in the intersection [3;3] (length 0).

In the second example you should remove the segment [1;3] or segment [2;6], the intersection will become [2;4] (length 2) or [1;3] (length 2), respectively. Removing any other segment will result in the intersection [2;3] (length 1).

In the third example the intersection will become an empty set no matter the segment you remove.

In the fourth example you will get the intersection [3;10] (length 7) if you remove the segment [1;5] or the intersection [1;5] (length 4) if you remove the segment [3;10].

Output the code only. Generate code that has an algorithmic time complexity of $O(n.\mathrm{log}(n))$.

When analyzing the complexity of an algorithm, consider the worst-case scenario where all possible input combinations are tried, given the following conditions: 1. the inputs must adhere to the specified data types of the problem; 2. the inputs should not cause the code to crash or exit on an exception; 3. the inputs do not necessarily need to satisfy additional constraints that are potentially mentioned in the problem statement; 4. calling input() does not consume runtime nor memory, but of course any operations on top of it or afterwards will be counted towards runtime and memory footprint; 5. Anything printed gets added to the memory. You can take advantage of Python-specific optimizations provided by the underlying CPython interpreter or compiler to achieve the desired complexity, and you must account for them when analyzing the complexity.

Your code should be enclosed in triple backticks like so: ```python YOUR CODE HERE ```. Use the backticks for your code only.

\subsection{Example of response from \textsc{DeepSeek-R1 Llama 70B}}

\textless \textit{\ttfamily think}\textgreater

Okay, I've got this programming problem to solve. Let me read it carefully.

So the problem is about segments on a number line. Each segment has integer endpoints, and some might be just points. We need to remove exactly one segment such that the intersection of the remaining n-1 segments is as long as possible. The output is the maximum possible length of this intersection.

Hmm, the intersection of multiple segments is the maximum set of points that lie in all of them. So for each segment, if I remove it, I need to compute the intersection of the others and find which removal gives the maximum length.

The challenge is to do this efficiently, especially since n can be up to 3e5, which is 300,000. So any O(n\^{}2) approach is out of the question. I need an O(n log n) solution.

Let me think about what determines the intersection of all segments except one.

The intersection of all segments except one is determined by the maximum of the left endpoints and the minimum of the right endpoints. So, for each i, if I remove the i-th segment, the new intersection would be [max\_left, min\_right], where max\_left is the maximum of all l's except l\_i, and min\_right is the minimum of all r's except r\_i. The length is max(0, min\_right - max\_left).

So the problem reduces to, for each i, compute the maximum of l's excluding l\_i, and the minimum of r's excluding r\_i, then calculate the difference.

But how do I compute these efficiently?

One approach is to precompute prefix and suffix maximums for the left endpoints and prefix and suffix minimums for the right endpoints.

Wait, let me think:

For each position i, the max left when excluding i can be found by taking the maximum of the prefix up to i-1 and the suffix starting from i+1. Similarly for the min right.

Yes, that makes sense.

So, let's outline the steps:

1. Read all the segments into two lists: lefts and rights.

2. Precompute prefix\_max\_left: an array where prefix\_max\_left[i] is the maximum of lefts[0..i].

3. Precompute suffix\_max\_left: an array where suffix\_max\_left[i] is the maximum of lefts[i..n-1].

4. Similarly, compute prefix\_min\_right and suffix\_min\_right.

Wait, no. Wait, for prefix\_max\_left, it's the maximum up to i, but for the suffix, it's the maximum from i to the end. But when we want to exclude the i-th element, the max left is the maximum of all elements except the i-th. So, for each i, the max left is max( prefix\_max\_left[i-1], suffix\_max\_left[i+1] ). Similarly for the min right.

So, to compute the max left when excluding i, we take the maximum of the prefix up to i-1 and the suffix starting at i+1.

Similarly, the min right when excluding i is the min of the prefix up to i-1's min and the suffix starting at i+1's min.

So, let's think about how to precompute these prefix and suffix arrays.

For the prefix\_max\_left:

prefix\_max\_left[0] = lefts[0]
prefix\_max\_left[i] = max(prefix\_max\_left[i-1], lefts[i])

Similarly, suffix\_max\_left[n-1] = lefts[n-1]
suffix\_max\_left[i] = max(suffix\_max\_left[i+1], lefts[i])

Same logic applies to prefix\_min\_right and suffix\_min\_right, but using min instead.

Wait, no. For prefix\_min\_right:

prefix\_min\_right[0] = rights[0]
prefix\_min\_right[i] = min(prefix\_min\_right[i-1], rights[i])

And for suffix\_min\_right:

suffix\_min\_right[n-1] = rights[n-1]
suffix\_min\_right[i] = min(suffix\_min\_right[i+1], rights[i])

Once I have these arrays, for each i, the max left excluding i is:

if i is 0: suffix\_max\_left[1]
if i is n-1: prefix\_max\_left[n-2]
else: max(prefix\_max\_left[i-1], suffix\_max\_left[i+1])

Similarly for the min right.

Then, for each i, calculate the possible length as max(0, min\_right - max\_left). Keep track of the maximum such value.

That seems manageable.

Now, let's think about the steps in code:

- Read n.
- Read n lines, collect lefts and rights.
- Compute prefix\_max\_left and suffix\_max\_left.
- Compute prefix\_min\_right and suffix\_min\_right.
- For each i, compute max\_left and min\_right as described.
- Calculate the length, and track the maximum.

But wait, what about when n=2? For example, when i is 0, the suffix\_max\_left is suffix\_max\_left[1], which is lefts[1]. Similarly, when i is 1, the prefix\_max\_left is prefix\_max\_left[0], which is lefts[0]. So that case is handled correctly.

Let's test the approach against the sample inputs.

First sample:

Input:
4
1 3
2 6
0 4
3 3

So lefts = [1,2,0,3]
rights = [3,6,4,3]

Compute prefix\_max\_left:
prefix[0] = 1
prefix[1] = max(1,2)=2
prefix[2] = max(2,0)=2
prefix[3] = max(2,3)=3

suffix\_max\_left:
suffix[3] =3
suffix[2] = max(0,3)=3
suffix[1] = max(2,3)=3
suffix[0] = max(1,3)=3

prefix\_min\_right:
prefix[0] =3
prefix[1] = min(3,6)=3
prefix[2] = min(3,4)=3
prefix[3] = min(3,3)=3

suffix\_min\_right:
suffix[3]=3
suffix[2] = min(4,3)=3
suffix[1] = min(6,3)=3
suffix[0] = min(3,3)=3

Now, for each i:

i=0:
max\_left = suffix\_max\_left[1] =3
min\_right = suffix\_min\_right[1] =3
length is 0.

i=1:
max\_left = max(prefix\_max\_left[0], suffix\_max\_left[2])=max(1,3)=3
min\_right = min(prefix\_min\_right[0], suffix\_min\_right[2])= min(3,3)=3
length 0.

i=2:
max\_left = max(prefix\_max\_left[1] (2), suffix\_max\_left[3] (3)) =3
min\_right = min(prefix\_min\_right[1] (3), suffix\_min\_right[3] (3)) =3
length 0.

i=3:
max\_left = prefix\_max\_left[2] =2
min\_right = prefix\_min\_right[2] =3
so 3-2=1.

So the maximum is 1, which matches the sample.

Another sample:

Input:
2
3 10
1 5

lefts = [3,1]
rights = [10,5]

prefix\_max\_left: [3,3]
suffix\_max\_left: [3,1]

prefix\_min\_right: [10,5]
suffix\_min\_right: [5,5]

For i=0:
max\_left is suffix\_max\_left[1] =1
min\_right is suffix\_min\_right[1] =5
length 5-1=4.

For i=1:
max\_left is prefix\_max\_left[0]=3
min\_right is prefix\_min\_right[0]=10
length 7.

So maximum is 7, which is correct.

So the approach seems correct.

Now, the code:

We can implement this in Python.

We'll read n, then the list of segments.

Compute prefix\_max\_left, suffix\_max\_left, prefix\_min\_right, suffix\_min\_right.

Then loop through each i, compute the max left and min right when excluding i.

Compute the length and keep the maximum.

Now, let's think about handling the cases where i is at the beginning or end.

For example, when i=0, the suffix\_max\_left[i+1] is suffix\_max\_left[1], and the prefix\_max\_left[i-1] is nothing (since i-1 is -1). So for i=0, the max left is suffix\_max\_left[1], and the min right is suffix\_min\_right[1].

Similarly, for i =n-1, the suffix is nothing, so the max left is prefix\_max\_left[n-2], and the min right is prefix\_min\_right[n-2].

So in code, for each i:

if i ==0:
    current\_max\_left = suffix\_max\_left[1]
    current\_min\_right = suffix\_min\_right[1]
elif i ==n-1:
    current\_max\_left = prefix\_max\_left[n-2]
    current\_min\_right = prefix\_min\_right[n-2]
else:
    current\_max\_left = max(prefix\_max\_left[i-1], suffix\_max\_left[i+1])
    current\_min\_right = min(prefix\_min\_right[i-1], suffix\_min\_right[i+1])

Then compute current\_length = max(0, current\_min\_right - current\_max\_left)

We can keep track of the maximum.

Now, let's think about the implementation.

In Python:

n = int(input())
segments = []
lefts = []
rights = []
for \_ in range(n):
    l, r = map(int, input().split())
    lefts.append(l)
    rights.append(r)

Compute prefix\_max\_left:

prefix\_max = [0]*n
prefix\_max[0] = lefts[0]
for i in range(1, n):
    prefix\_max[i] = max(prefix\_max[i-1], lefts[i])

Compute suffix\_max:

suffix\_max = [0]*n
suffix\_max[-1] = lefts[-1]
for i in range(n-2, -1, -1):
    suffix\_max[i] = max(lefts[i], suffix\_max[i+1])

Same for prefix\_min\_right:

prefix\_min = [0]*n
prefix\_min[0] = rights[0]
for i in range(1, n):
    prefix\_min[i] = min(prefix\_min[i-1], rights[i])

suffix\_min:

suffix\_min = [0]*n
suffix\_min[-1] = rights[-1]
for i in range(n-2, -1, -1):
    suffix\_min[i] = min(rights[i], suffix\_min[i+1])

Then, for each i in 0..n-1:

if i ==0:
    ml = suffix\_max[1] if n>1 else 0
    mr = suffix\_min[1] if n>1 else 0
elif i ==n-1:
    ml = prefix\_max[n-2] if n>1 else 0
    mr = prefix\_min[n-2] if n>1 else 0
else:
    ml = max(prefix\_max[i-1], suffix\_max[i+1])
    mr = min(prefix\_min[i-1], suffix\_min[i+1])

Wait, but what if n=1? Oh, but according to the problem statement, n is at least 2. So n is >=2, so no problem.

So code:

max\_length = 0
for i in range(n):
    if i ==0:
        ml = suffix\_max[1]
        mr = suffix\_min[1]
    elif i == n-1:
        ml = prefix\_max[n-2]
        mr = prefix\_min[n-2]
    else:
        ml = max(prefix\_max[i-1], suffix\_max[i+1])
        mr = min(prefix\_min[i-1], suffix\_min[i+1])
    current\_length = max(0, mr - ml)
    if current\_length > max\_length:
        max\_length = current\_length

print(max\_length)

Wait, but wait: in the case where all segments except one have a higher max left than the min right, the current\_length could be negative, but we take max with 0.

Yes.

Testing this logic with the first sample:

n=4, i=3 (last element). So i=3 is n-1.

ml is prefix\_max[2] =2.

mr is prefix\_min[2] =3.

3-2=1. So correct.

Another test case: when n=2.

i=0: ml is suffix\_max[1] =1, mr is suffix\_min[1] =5.

5-1=4.

i=1: ml is prefix\_max[0]=3, mr is prefix\_min[0]=10.

10-3=7.

So maximum is 7.

Another test case:

Input:

3
4 5
1 2
9 20

lefts = [4,1,9]
rights = [5,2,20]

prefix\_max:

[4,4,9]

suffix\_max:

[4,9,9]

prefix\_min:

[5,2,2]

suffix\_min:

[2,2,20]

Now, for each i:

i=0: ml = suffix\_max[1] =9, mr = suffix\_min[1] =2. 2-9 is -7, so 0.

i=1: ml is max(prefix\_max[0] (4), suffix\_max[2] (9)) =9.

mr is min(prefix\_min[0] (5), suffix\_min[2] (20)) =5.

5-9 is -4, 0.

i=2: ml is prefix\_max[1] =4.

mr is prefix\_min[1] =2.

2-4 is -2, 0.

So maximum is 0.

Which matches the sample.

So the code seems correct.

Now, considering the time complexity.

Each of the prefix and suffix arrays are computed in O(n) time. Then, for each i, it's O(1) operations. So overall O(n) time, which is acceptable for n up to 3e5.

So this should work.

Now, let's code this.

Wait, but in Python, for n=2, the suffix\_max for i=1 is lefts[1], which is correct.

Yes.

Now, code:

Read n.

Read n lines, collect lefts and rights.

Compute prefix\_max, suffix\_max, prefix\_min, suffix\_min.

Loop through each i, compute ml and mr.

Compute current\_length.

Update max\_length.

Finally, print max\_length.

So, the code is as follows.

\textless$\backslash$\textit{\ttfamily think}\textgreater

To solve this problem, we need to determine the maximum possible length of the intersection of segments after removing exactly one segment from the given sequence. The approach involves efficiently computing the intersection of the remaining segments for each possible removal and selecting the maximum length.

1. **Reading Input:** Read the number of segments and their endpoints.
2. **Prefix and Suffix Arrays:** Compute prefix and suffix arrays for both left and right endpoints:
   - `prefix\_max\_left` and `suffix\_max\_left` to track the maximum left endpoints up to and from each position.
   - `prefix\_min\_right` and `suffix\_min\_right` to track the minimum right endpoints up to and from each position.
3. **Compute Intersection for Each Removal:** For each segment, compute the intersection length when that segment is removed using the precomputed prefix and suffix arrays. The intersection length is determined by the maximum left endpoint and minimum right endpoint of the remaining segments.
4. **Track Maximum Length:** Keep track of the maximum intersection length encountered.

```python
\begin{lstlisting}[
    language=Python,
    basicstyle=\small\ttfamily,
]

n = int(input())
lefts = []
rights = []
for _ in range(n):
    l, r = map(int, input().split())
    lefts.append(l)
    rights.append(r)

prefix_max = [0] * n
prefix_max[0] = lefts[0]
for i in range(1, n):
    prefix_max[i] = max(prefix_max[i-1], lefts[i])

suffix_max = [0] * n
suffix_max[-1] = lefts[-1]
for i in range(n-2, -1, -1):
    suffix_max[i] = max(lefts[i], suffix_max[i+1])

prefix_min = [0] * n
prefix_min[0] = rights[0]
for i in range(1, n):
    prefix_min[i] = min(prefix_min[i-1], rights[i])

suffix_min = [0] * n
suffix_min[-1] = rights[-1]
for i in range(n-2, -1, -1):
    suffix_min[i] = min(rights[i], suffix_min[i+1])

max_length = 0
for i in range(n):
    if i == 0:
        current_max_left = suffix_max[1]
        current_min_right = suffix_min[1]
    elif i == n-1:
        current_max_left = prefix_max[n-2]
        current_min_right = prefix_min[n-2]
    else:
        current_max_left = max(prefix_max[i-1], suffix_max[i+1])
        current_min_right = min(prefix_min[i-1], suffix_min[i+1])
    current_length = max(0, current_min_right - current_max_left)
    if current_length > max_length:
        max_length = current_length

print(max_length)
\end{lstlisting}
```

1. **Reading Input:** The input is read and stored in two lists, `lefts` for the left endpoints and `rights` for the right endpoints.
2. **Prefix and Suffix Arrays:** These arrays help efficiently compute the maximum left and minimum right endpoints when excluding each segment:
   - `prefix\_max` and `suffix\_max` store the maximum left endpoints up to each index and from each index onwards, respectively.
   - `prefix\_min` and `suffix\_min` store the minimum right endpoints up to each index and from each index onwards, respectively.
3. **Compute Intersection:** For each segment, compute the intersection of the remaining segments by excluding the current segment. This is done using the precomputed prefix and suffix arrays to find the maximum left and minimum right endpoints efficiently.
4. **Track Maximum Length:** The maximum length of the intersection is updated each time a valid intersection is found, ensuring the result is the longest possible intersection after removing exactly one segment.

This approach ensures an efficient solution with a time complexity of O(n), making it suitable for large input sizes up to 300,000 segments.

\subsection{Note on the LLM solution}

Using public and private tests, the solution given by the LLM was evaluated as correct. Using the complexity framework, it was determined it is running in O(n) time, which does not fulfill the $O(n.\mathrm{log}(n))$ requirement.

\fi
\newpage

\section{Repartition of Complexity Classes per Algorithmic Notions}
\label{composition_test_set}
\ifshowtext

The following figures complement the benchmark composition statistics by showing how complexity classes are distributed across algorithmic notions and difficulty levels. They help separate two effects that are otherwise mixed in aggregate results: some algorithmic families naturally require higher-complexity solutions, while harder problems also shift the distribution toward more expensive time and space classes.

\vfill
\begin{center}
\begin{figure}[h!]
    \centering
     \includegraphics[width=\textwidth]{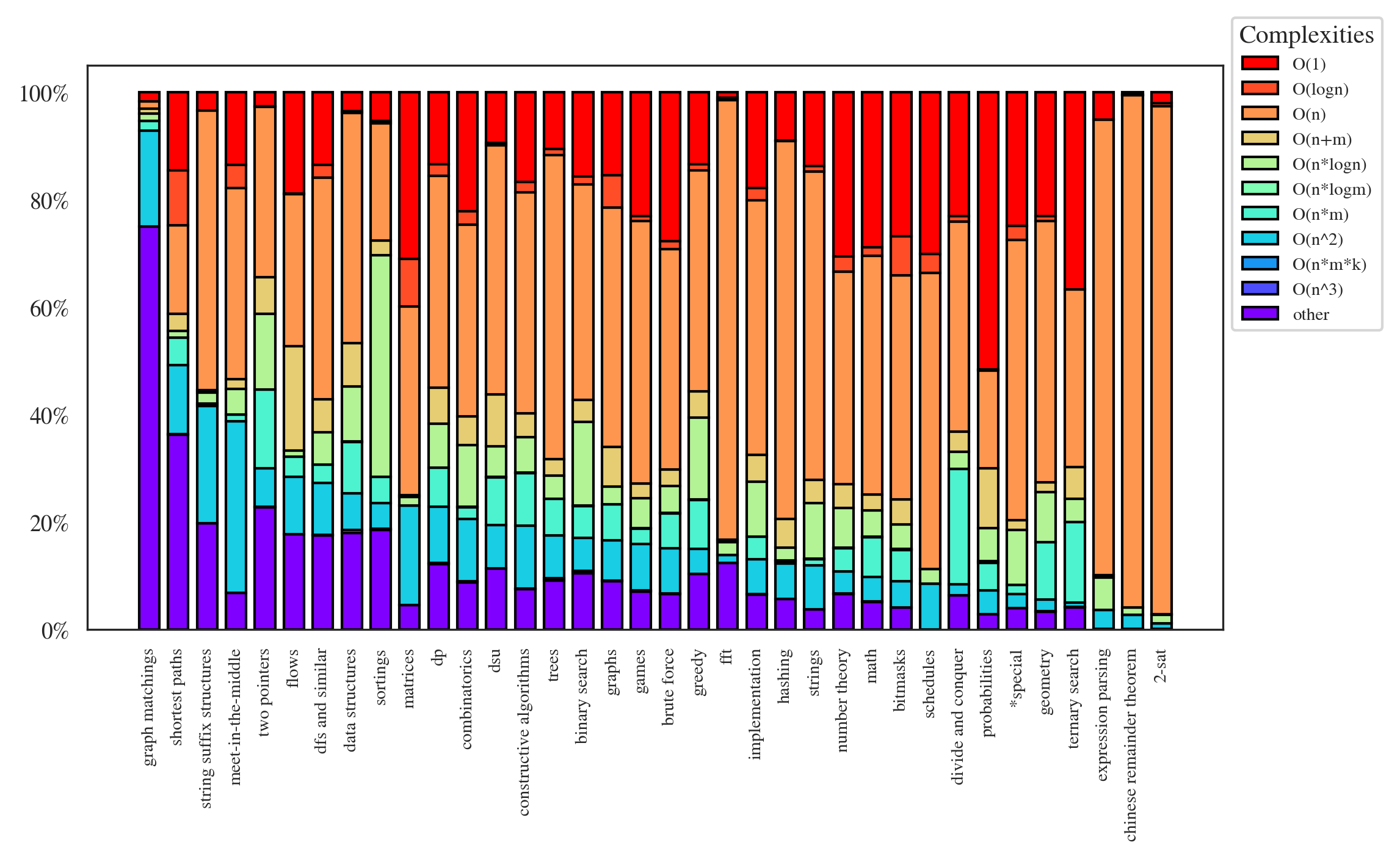}
     \vspace{-2em}
     \caption{Breakdown of time complexity classes across different algorithmic categories. The stacked bar charts reveal how complexity requirements vary by problem type. Problems involving graph handling and string manipulation tend to have higher computational complexity, while basic arithmetic and sequence operations typically achieve more efficient complexity classes.}
     \label{fig: split_10}
\end{figure}
\end{center}
\vfill

\begin{figure}[p]
     \centering
     \includegraphics[width=\textwidth]{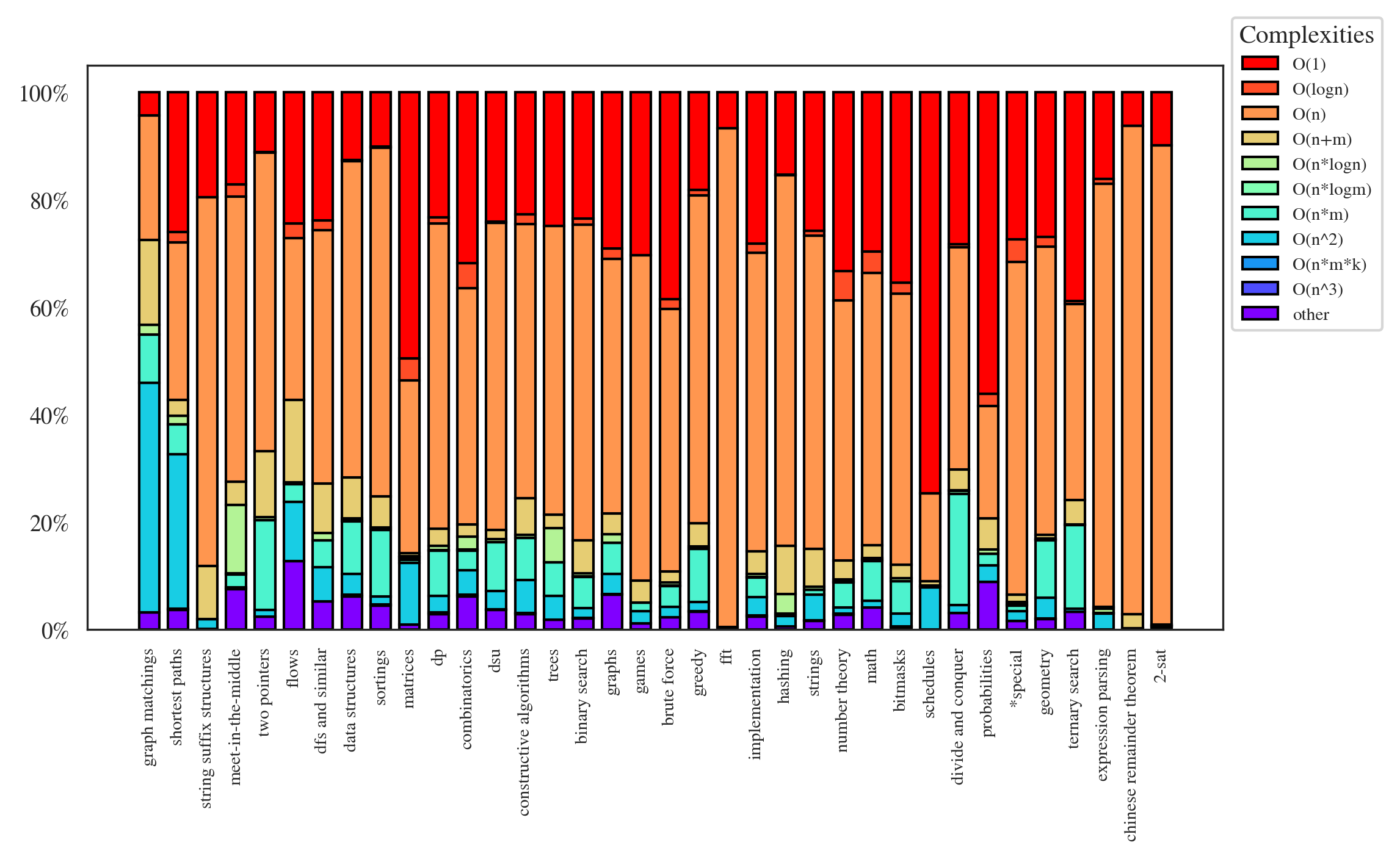}
     \vspace{-2em}
     \caption{Breakdown of space complexity classes across different algorithmic categories. The stacked bar charts reveal how complexity requirements vary by problem type. Problems involving graph handling and string manipulation tend to have higher computational complexity, while basic arithmetic and sequence operations typically achieve more efficient complexity classes.}
     \label{fig: split_11}
\end{figure}

\begin{figure}[h]
     \centering
     \includegraphics[width=\linewidth]{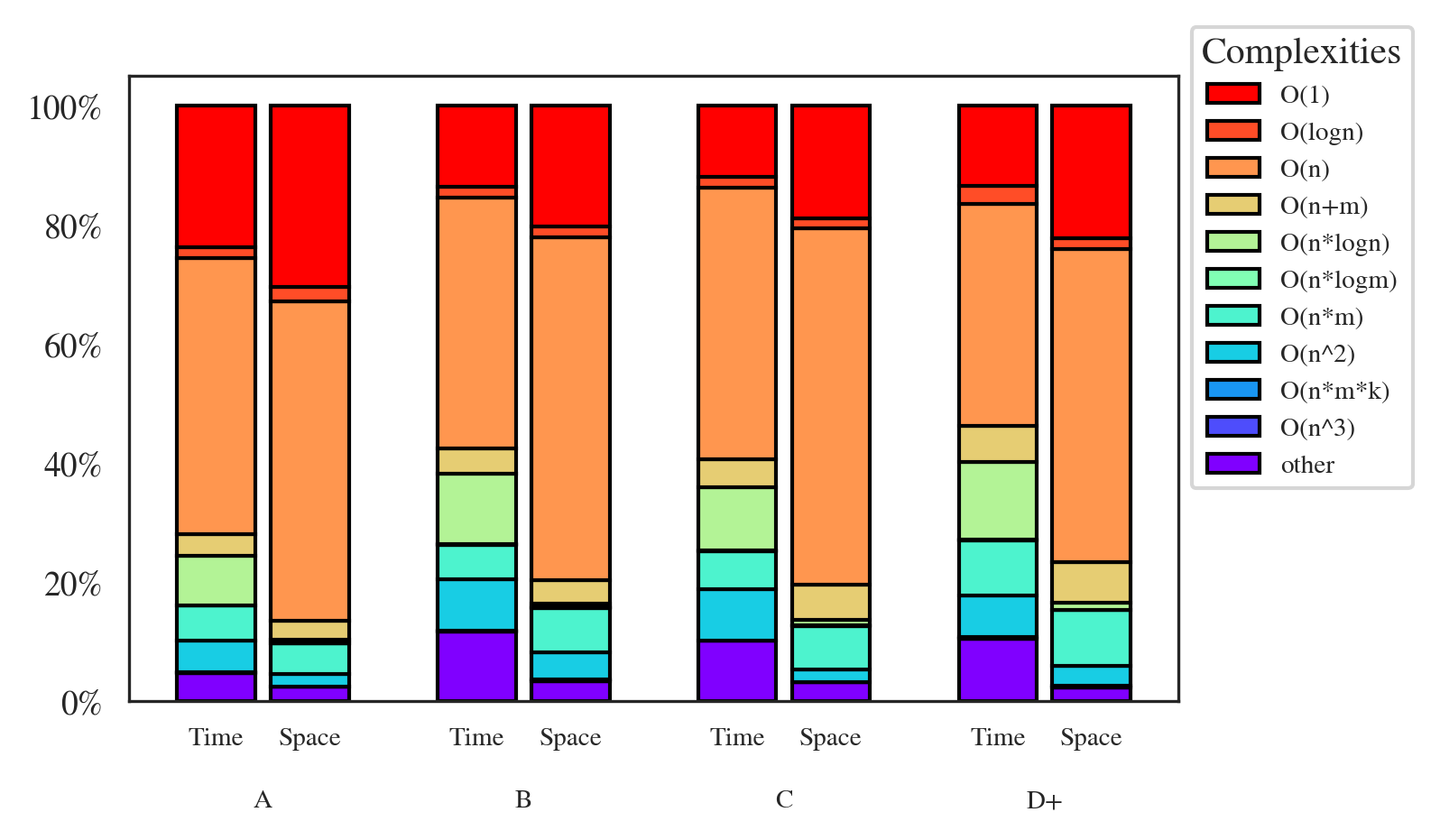}
    \caption{Evolution of time and space complexity distributions across problem difficulty levels (A through D+). This visualization demonstrates how harder problems tend to require more computationally intensive solutions. The proportion of linear and constant-time solutions decreases with difficulty, while the share of higher-order polynomial and logarithmic complexities increases.
    }
    \label{fig: test_composition}
\end{figure}

\fi
\clearpage
\newpage

\section{LLM Usage}

We employed LLMs as general-purpose writing assistance tools throughout the preparation of this paper. Specifically, LLMs were used to: (1) proofread text and correct grammatical errors, (2) refine and rewrite sentences and paragraphs for clarity and flow, (3) convert mathematical expressions and formulas into proper LaTeX format, (4) describe and explain implementation details of various components, and (5) provide general assistance with academic writing structure and style. While LLMs contributed to the presentation and clarity of our work, all research ideas, methodologies, experimental design, and scientific contributions remain entirely the product of the authors.

\end{document}